# Transformative AGI by 2043 is <1% likely

By Ari Allyn-Feuer and Ted Sanders

## Abstract

This paper is a submission to the Open Philanthropy AI Worldviews Contest. In it, we estimate the likelihood of *transformative* artificial general intelligence (AGI) by 2043 and find it to be <1%.

Specifically, we argue:

- **The bar is high:** AGI as defined by the contest—something like AI that can perform nearly all valuable tasks at human cost or less—which we will call *transformative* AGI is a *much* higher bar than merely massive progress in AI, or even the unambiguous attainment of expensive superhuman AGI or cheap but uneven AGI.
- **Many steps are needed:** The probability of transformative AGI by 2043 can be decomposed as the joint probability of a number of necessary steps, which we group into categories of software, hardware, and sociopolitical factors.
- **No step is guaranteed:** For each step, we estimate a probability of success by 2043, conditional on prior steps being achieved. Many steps are quite constrained by the short timeline, and our estimates range from 16% to 95%.
- **Therefore, the odds are low:** Multiplying the cascading conditional probabilities together, we estimate that transformative AGI by 2043 is **0.4% likely**. Reaching >10% seems to require probabilities that feel unreasonably high, and even 3% seems unlikely.

Thoughtfully applying the cascading conditional probability approach to this question yields lower probability values than is often supposed. This framework helps enumerate the many future scenarios where humanity makes partial but incomplete progress toward transformative AGI.

# Executive summary

For AGI to do most human work for <$25/hr by 2043, many things must happen:

| **Event** | **Forecast** by 2043 or TAGI, ***conditional*** on prior steps[1] | **Summary of our reasoning** |
|---|---|---|
| We invent algorithms for transformative AGI | 60% | Hard to say. No one knows how to do it, but progress is promising and effort is accelerating. |
| We invent a way for AGIs to learn faster than humans | 40% | Hard to say. Many skills take humans years or decades to learn, are learned from sequential feedback, not demonstration. Rapid learning will require new simulation or parallelization techniques. |
| AGI inference costs drop below $25/hr (per human equivalent) | 16% | This step is challenging.<br><br>Software: Our best anchor for how much compute an AGI needs is the human brain, which we estimate to perform 1e20–1e21 FLOPS. It may be that AGI needs much less (fewer design constraints than brains), or much more (early versions are always crummy, plus silicon matrix multipliers have their own constraints). We are very uncertain and therefore guess it's 80% likely that AGI will be within ±4 orders of magnitude of the brain, with 20% reserved for the tails.<br><br>Hardware: In addition, we estimate that today's computer hardware is ~5 orders of magnitude less cost efficient and energy efficient than brains. Based on long-run trends, and the lack of any real successor to transistors, we expect hardware to improve by no more than ~2 orders of magnitude within 15 years, with most gains coming from chip design. We are pessimistic that AI will rapidly accelerate progress, as energy efficiency in particular is up against well-known laws of semiconductor physics.<br><br>We combine probability distributions over software needs and hardware progress to arrive at a final estimate of 16%. |
| We invent and scale cheap, quality robots | 60% | Robots are much worse than human bodies, and basic tasks remain unsolved after decades of effort. But the trend is positive, and conditional on AGI, it will likely accelerate. |

[1] There might be a few spots where it makes more sense to condition the probabilities in a different order than top to bottom. E.g., what are the odds we scale semiconductor production given that we avoid all derailments rather than vice versa. We wanted to keep the positive and negative factors grouped, and do not dedicate time to discuss different orderings. You may, if you wish. It makes more of a difference on our 2100 estimates than our 2043 estimates.

| | | |
|---|---|---|
| We massively scale production of chips and power | 46% | If AGI needs are small (say, ~100k silicon wafers), manufacturing will be easy. If AGI needs are large (say, ~1B silicon wafers), it will be nearly impossible to make the gargantuan investments required, especially if the ROI is uncertain.<br><br>Our uncertainties rank: (1) compute & power needs, (2) acceleration by AI, and (3) numbers of years to build. |
| We avoid derailment by human regulation | 70% | Conditional on AGI, we expect the odds of regulation to rise massively. Imagine a world with computers that can think and talk, and whose costs are falling rapidly. People's jobs and futures will feel unsafe. There may be massive social anxiety and unrest. There will also be massive national security implications. We expect governments to act. |
| We avoid derailment by AI-caused delay | 90% | Furthermore, superintelligent but expensive AGI may itself warn us to slow progress, to forestall potential catastrophe that would befall both us and it. |
| We avoid derailment from wars (e.g., China invades Taiwan) | 70% | China has stated plainly it intends to reunify (invade) Taiwan. A majority of its population supports invasion. Its military is preparing to be ready for invasion. Even if an invasion doesn't spark war, the sanctions applied in retaliation will shut down most of the world's AI chip production.<br><br>Conditional on AGI, we expect the odds of war to rise substantially. Imagine you're Russia and the USA invents AGI. Soon, your export market may be eradicated. Your military power may be eradicated. Your standing in the world may be eradicated. Your economic vitality may be eradicated. Your odds of lashing out rise. |
| We avoid derailment from pandemics | 90% | Conditional on AGI and increasingly accessible biotechnology, the risk of natural pandemics will fall while the risk of engineered pandemics will rise. |
| We avoid derailment from severe depressions | 95% | If depressions occur once every 200 years (~10% chance per 20 years), and a depression has a 50% chance of delaying AGI beyond 2043 via chilled investment, there may be a 5% chance that this occurs by 2043. |
| **Joint odds** | **0.4%** | |

If you think our estimates are pessimistic, feel free to substitute your own here. You'll find it difficult to arrive at odds above 10%.

Of course, the difficulty is by construction. Any framework that multiplies ten probabilities together is almost fated to produce low odds.

So a good skeptic must ask: Is our framework fair?

There are two possible errors to beware of:

- Did we neglect possible parallel paths to transformative AGI?
- Did we hew toward unconditional probabilities rather than fully conditional probabilities?

We believe we are innocent of both sins.

Regarding failing to model parallel disjunctive paths:

- We have chosen generic steps that don't make rigid assumptions about the particular algorithms, requirements, or timelines of AGI technology
- One opinionated claim we do make is that transformative AGI by 2043 will almost certainly be run on semiconductor transistors powered by electricity and built in capital-intensive fabs, and we spend many pages justifying this belief

Regarding failing to really grapple with conditional probabilities:

- Our conditional probabilities are, in some cases, quite different from our unconditional probabilities. In particular, we assume that a world on track to transformative AGI will…
    - Construct semiconductor fabs and power plants at a far faster pace than today (our unconditional probability is *substantially* lower)
    - Have invented very cheap and efficient chips by today's standards (our unconditional probability is *substantially* lower)
    - Have *higher* risks of disruption by regulation
    - Have *higher* risks of disruption by war
    - Have *lower* risks of disruption by natural pandemic
    - Have *higher* risks of disruption by engineered pandemic

Therefore, for the reasons above—namely, that transformative AGI is a very high bar (far higher than "mere" AGI) and many uncertain events must jointly occur—we are persuaded that the likelihood of transformative AGI by 2043 is <1%, a much lower number than we otherwise intuit. We nonetheless anticipate stunning advancements in AI over the next 20 years, and forecast substantially higher likelihoods of transformative AGI beyond 2043.

# Introduction

The discourse around AI, and particularly AGI, is tumultuous right now. Thrust into popular thought by the flashy, captivating achievements of ChatGPT and scientific AI like AlphaFold, the subject has gained the attention of the world. More people are awakening to the notion that the future of AI development, its general nature, timing, and specific details, will mean a lot, maybe everything, to humans at large and humanity as a whole. Eliezer Yudkowsky, a prominent AI theorist, is fond of saying that the development of AI is the only important thing about this era of history. If existential AI-related risk to humanity is realized, he will be right.

As a result of all this, and of the human tendency to think in sweeping categorical terms, the debate over AGI timelines often takes the form of dueling, orthogonal categorical pronouncements between people with hopeful, despondent, ecstatic, and dismissive casts of mind. Some people extrapolate recent rapid developments in some areas and argue that transformative AGI is imminent, even to the point of neglecting future-oriented activities like saving and childbearing, or advocating violence to restrain imminent AGI development. Meanwhile, others dismiss recent accomplishments, and maintain that AGI is a remote and dubious cause, or even a philosophical impossibility. Others lazily eschew prediction and analysis entirely. And these groups of people, in addition to disagreeing, by and large do not even know how to talk to each other.

This essay is an exercise in quantification, an attempt to create a common framework and language for thinking about one particular AI timeline question, and use that framework to construct and defend a quantitative estimate. This question is one of the topics of your worldview contest: the probability of transformative AGI developing by 2043. Because it is precise and time-bound, this question is particularly amenable to being analyzed quantitatively, relative to nebulous questions like “can AI really be creative” and “is GPT-4 the path to AGI” and “in light of imminent AGI, should I bother eating broccoli?”

Our framework is based on a cascade of conditional probabilities, of software, hardware, and sociopolitical events that must happen, sequentially (either conceptually sequentially or temporally sequentially), in order for AGI to transform the world economy by 2043. They are approximately seven:

1. We must discover fundamental algorithmic improvements enabling computers to perform brainlike functions.
2. We must quickly figure out how to train computers to do human tasks without sequential reinforcement learning.
3. Efficiency of AGI computation must achieve that of humans (e.g., <$25/hr).
4. We must invent and scale cheap and capable robot bodies.
5. We must quickly scale up semiconductor manufacturing and electrical generation.
6. We, and any AGI partners, must not choose to slow our progress toward transformative AGI as we see it developing.
7. We must not allow progress toward transformative AGI to be derailed by geopolitical competition, pandemics, or severe economic depressions.

For each of the major events, we will explain what it is and why it's necessary, and derive and defend a specific numerical estimate, our opinion, of its probability contingent on the preceding events. Because the subject is broad and details matter, constructing and evaluating this framework will take us through a lot of subjects, and we thank you in advance for your attention and patience.

Although we will defend specific numerical estimates of these individual conditional probabilities, we hope that this decomposition of the problem can be of value to people who disagree with us about some or many of the individual estimates. And perhaps even that this kind of thinking can be a bridge, or a piece of a bridge, between people with very different ways of thinking about AI.

In the end, the specific numerical estimate on which we landed will surprise you, as it did us. We estimate that the chance of transformative AGI by 2043 is a small fractional percentage, and moreover, that the general trend, wherein the probability of this event is much smaller than the contest judges and many experts think, is robust to responsible disagreement about the individual parameters. This puts us in the position of arguing that the judges of the contest are wrong, and that we ourselves were wrong until recently. This may be a poor way to win an essay contest, but we will pursue it nonetheless, because we are convinced. We hope to convince you as well.

# About the authors

This essay is jointly authored by Ari Allyn-Feuer and Ted Sanders. Below, we share our areas of expertise and track records of forecasting. Of course, credentials are no guarantee of accuracy. We share them not to appeal to our authority (plenty of experts are wrong), but to suggest that if it sounds like we've said something obviously wrong, it may merit a second look (or at least a compassionate understanding that not every argument can be explicitly addressed in an essay trying not to become a book).

## Ari Allyn-Feuer

**Areas of expertise**

I am a decent expert in the complexity of biology and using computers to understand biology.

- I earned a Ph.D. in Bioinformatics at the University of Michigan, where I spent years using ML methods to model the relationships between the genome, epigenome, and cellular and organismal functions. At graduation I had offers to work in the AI departments of three large pharmaceutical and biotechnology companies, plus a biological software company.
- I have spent the last five years as an AI Engineer, later Product Manager, now Director of AI Product, in the AI department of GSK, an industry-leading AI group which uses cutting edge methods and hardware (including Cerebras units and work with quantum computing), is connected with leading academics in AI and the epigenome, and is particularly engaged in reinforcement learning research.

**Evidence of calibration**

While I don't have Ted's explicit formal credentials as a forecaster, I've issued some pretty important public correctives of then-dominant narratives:

- I said in print on January 24, 2020 that due to its observed properties, the then-unnamed novel coronavirus spreading in Wuhan, China, had a significant chance of promptly going pandemic and killing tens of millions of humans. It subsequently did.
- I said in print in June 2020 that it was an odds-on favorite for mRNA and adenovirus COVID-19 vaccines to prove highly effective and be deployed at scale in late 2020. They subsequently did and were.
- I said in print in 2013 when the Hyperloop proposal was released that the technical approach of air bearings in overland vacuum tubes on scavenged rights of way wouldn't work. Subsequently, despite having insisted they would work and spent millions of dollars on them, every Hyperloop company abandoned all three of these elements, and development of Hyperloops has largely ceased.
- I said in print in 2016 that Level 4 self-driving cars would not be commercialized or near commercialization by 2021 due to the long tail of unusual situations, when several major car companies said they would. They subsequently were not.
- I used my entire net worth and borrowing capacity to buy an abandoned mansion in 2011, and sold it seven years later for five times the price.

Luck played a role in each of these predictions, and I have also made other predictions that didn't pan out as well, but I hope my record reflects my decent calibration and genuine open-mindedness.

## Ted Sanders

**Areas of expertise**

I am a decent expert in semiconductor technology and AI technology.

- I earned a PhD in Applied Physics from Stanford, where I spent years researching semiconductor physics and the potential of new technologies to beat the 60 mV/dec limit of today's silicon transistor (e.g., magnetic computing, quantum computing, photonic computing, reversible computing, negative capacitance transistors, and other ideas). These years of research inform our perspective on the likelihood of hardware progress over the next 20 years.
- After graduation, I had the opportunity to work at Intel R&D on next-gen computer chips, but instead, worked as a management consultant in the semiconductor industry and advised semiconductor CEOs on R&D prioritization and supply chain strategy. These years of work inform our perspective on the difficulty of rapidly scaling semiconductor production.
- Today, I work on AGI technology as a research engineer at OpenAI, a company aiming to develop transformative AGI. This work informs our perspective on software progress needed for AGI. (Disclaimer: nothing in this essay reflects OpenAI's beliefs or its non-public information.)

**Evidence of calibration**

I have a track record of success in forecasting competitions:

- Top prize in SciCast technology forecasting tournament (15 out of ~10,000[2], ~$2,500 winnings)
- Top Hypermind US NGDP forecaster in 2014 (1 out of ~1,000)
- 1st place Stanford CME250 AI/ML Prediction Competition[3] (1 of 73)
- 2nd place 'Let's invent tomorrow' Private Banking prediction market (2 out of ~100)
- 2nd place DAGGRE Workshop competition (2 out of ~50)
- 3rd place LG Display Futurecasting Tournament (3 out of 100+)
- 4th Place SciCast conditional forecasting contest
- 9th place DAGGRE Geopolitical Forecasting Competition
- 30th place Replication Markets (~$1,000 winnings)
- Winner of ~$4200 in the 2022 Hybrid Persuasion-Forecasting Tournament on existential risks (told ranking was "quite well")

Each finish resulted from luck alongside skill, but in aggregate I hope my record reflects my decent calibration and genuine open-mindedness.

# Our understanding of your thinking

Generously, you've shared the beliefs of the panel of judges:

- You believe it is unlikely that AGI will be developed by Jan 1, 2043 ("*Panelist credences… range from ~10% to ~45%*")
- By AGI, you do not mean merely a computer as smart as human—you mean "*something like AI that can quickly and affordably be trained to perform **nearly all** economically and strategically valuable tasks at **roughly human cost or less***."

## Where we agree & disagree

We agree with you far more than we disagree:

- We agree AGI is possible.
- We agree AGI is likely (in fact, we are so bullish on AI that we are investing our careers in it).
- We agree AGI is likely to have an enormous impact on humanity's trajectory.
- We agree AGI poses a potential existential threat to humanity, and is worth worrying about today.
- We agree that predicting uninvented technology is hard, and that it's prudent to avoid high confidence in any scenario.[4]
- We agree AGI is unlikely (<50%) to be capable of nearly all human work at human cost or less by 2043.

Where we *disagree* is quite narrow: although we agree directionally that AGI is unlikely to be capable of nearly all human work at human cost or less by 2043, we are more confident than you. The rest of this essay focuses on the evidence and arguments that have persuaded us to raise our confidence.

---

[2] To be fair, most of the ~10,000 signups did not remain active.
[3] Not technically a forecasting competition. It was a Kaggle-style competition where the class had to train AI models to predict data collected in the past. Still, hopefully a small but useful datapoint in assessing my credibility to speak about AI.
[4] https://www.gwern.net/Forking-Path

# The high bar of transformative AGI

You define AGI as follows:

> *By "AGI" we mean something like "AI that can quickly and affordably be trained to perform nearly all economically and strategically valuable tasks at roughly human cost or less." AGI is a notoriously thorny concept to define precisely. What we're actually interested in is the potential existential threat posed by advanced AI systems. To that end, we welcome submissions that are oriented around related concepts, such as transformative AI, human-level AI, or PASTA.*

We understand this to mean that by 2043 the systems in question must already exist and be deployed at scale, as opposed to technology which merely leads suggestively in that direction.

This interpretation is bolstered by your explanation that your main motivation is the consideration of existential risk. Existential risk scenarios for AGI usually fall into two categories. Firstly, those based on the one-time performance of exceptional tasks generally beyond human capability, decapitating strikes against humanity with nanorobots or viruses or the like, enabled by superintelligence. Secondly, those based on the performance of many already-possible tasks at large scale (rise-of-the-robots and labor supplantation and wireheading/hedonic scenarios). For the first set of scenarios, the performance of currently existing and contemplated economically useful tasks, and the precise cost thereof, are irrelevant; designing life-eradicating nanobots is not an economically useful task, and since it need be done only once, cost is irrelevant, as is the versatility to perform a wide variety of other tasks.

We interpret that you are worried about the second set of scenarios, where AI algorithms exert existential risk onto human civilization by performing economically useful tasks in bulk, under whatever aegis. This means that deployment and scale, on the stated timeline, are necessary for the satisfaction of the hypothetical.

Because the term "AGI" by itself could be used to refer to AIs that can perform fewer tasks, or do so more slowly or with more difficulty or at smaller scale or at greater expense, **we will use the term "transformative AGI" to refer to AGI that meets your criteria**.

Under this stringent definition, many amazing things would ***not*** individually count as transformative AGI. E.g.,

- Self-driving cars and trucks and tractors safer than humans, which automate all driving jobs and put us on a clear path to making human driving obsolete.
- Robot bodies good enough to automate many routine tasks, and enable e.g. e-commerce fulfillment warehouses or retail with nearly no humans.
- Language models as good at reading and writing as humans, which power amazing feats of automated writing and text transformation, expert human-level translation of all text including poetry, and fluent chatbot interfaces for all major websites and businesses.
- Speech-to-text and text-to-speech models potent enough to make the above language models function as voice interfaces for all computers and websites and businesses.

- Conversational, agentic, continuous AGIs just as intelligent as humans, which can converse and perform a wide variety of tasks, and advise government and business leaders, and deploy to all space missions, and become fixtures on television, and write best-selling books and well-cited scientific papers, but require outsized computational resources to train and run.
- AGI-driven robot bodies able to perform most human labor tasks, though they are too expensive to substitute for most low-wage human labor.
- A marked trend in the cost and scale of AGI and robot bodies, indicating that they will eventually be cheap and capable enough to transform the world, though they have not already done so prior to 2043.

Our argument should not be taken as a prediction that transformative AGI as you define it is unlikely to develop later in the century, or that amazing developments in AI, and AGI, are unlikely to have taken place by January 1, 2043. In fact, our framework suggests strongly that it's likely that important steps in the direction of AGI, and potentially non-transformative AGI, will have been attained by 2043. In addition, below we use the same framework to argue that transformative AGI is rather likely by 2100.

That is, this is not an argument for general-purpose pessimism about AGI or transformative AGI, it is an argument that this specific, very high level of development is very unlikely to take place by that specific, and rather soon, date.

In addition, we will comment on how we have chosen to think about your cost criterion. As a way to operationalize your phrasing "roughly human cost or less," we are using a shorthand of $25/hour, including inference, robot bodies where necessary, and the amortized cost of training, to perform human-performable tasks. We choose a deliberately high figure, above the 90th percentile of human labor income now[5], for two reasons. One is that we do not want to be lawyering the rules with a primarily cost-based argument, but rather focusing on the fundamentals of what is possible and what is likely to be invented. AGIs performing nearly all tasks at even the 90th percentile cost of humans would be remarkable and transformative, and we want to reflect that. In addition, twenty years of global economic growth, especially in the optimistic timelines where AGI does emerge, might mean that $25/hour in today's money will be a much more common value for human laborers than it is today.

[5] https://howrichami.givingwhatwecan.org/how-rich-am-i?income=100000&countryCode=USA&household%5Badults%5D=2&household%5Bchildren%5D=3

# Prelude: How we think about forecasting

History abounds with experts who confidently claimed that a new, wild sounding technology was impossible (e.g., flight, spaceflight, etc.), and then were quickly refuted by its subsequent development.

Two excellent essays on the difficulties of technology forecasting are Arthur C. Clarke's Hazards of Prophecy[6] and Gwern's Technology Forecasting: The Garden of Forking Paths[7]. Both contemplate the reasoning mistakes that led otherwise intelligent experts to be rashly overconfident when proclaiming the impossibility of new technologies. Clarke classifies errors into Failures of Nerve and Failures of Imagination; Gwern similarly blames motivated stopping and lack of imagination.

The key idea is this: proving a technology *won't* happen is harder than it might seem. In fact, it is woefully insufficient to prove that a particular assumed configuration of a technology will fail; to actually prove a technology won't happen requires proving that *all* possible configurations of that technology will fail, *including those you have not yet imagined*. It's a terrible error to assume that no path exists just because you cannot personally imagine one.

One might conclude the antidote is to counterweight forecasts toward successful technological progress. But, of course, the same arguments can be mirrored. History *also* abounds with otherwise intelligent experts who confidently claimed a new wild sounding technology was arriving imminently (e.g., commercial hydrogen cars or anti-aging therapies or drone delivery or flying cars or moon bases or cures for cancer), and then were quickly proven wrong by subsequent challenges in development. Such failed predictions have been seen repeatedly in the AI field specifically, around self driving cars (in 1958), self driving cars (in 1995), self driving cars (in 2016), medical diagnosis and prognosis, and neurological simulation. Perhaps these experts similarly failed to imagine all the possible ways that technological development could be stymied within their forecast's time horizon.

As an analogy, you can picture technological development like attempting to traverse a dense, foggy forest laced with connected paths occasionally blocked by giant immovable boulders. Even if you are an expert who has traversed forest paths for decades and seen every path blocked by boulders, that is no guarantee that an unblocked path will never be found. Only a single path needs to connect the two sides of the forest for the forest to be traversable.

---

[6] https://book.cc.irregulaire.com/IT_cybernetics_speculation/profiles-of-the-future.pdf
Notably, Clarke's observations in Hazards of Prophecy don't necessarily translate into accurate forecasts. His book lists a table of future inventions extrapolated from past progress that he warns not to take too seriously. By 2020, we still lacked nuclear rockets, planetary landings, planetary colonization, interstellar probes, fusion power, wireless energy, sea mining, weather control, cyborgs, time & perception enhancement, and we don't seem to be close to space mining or bioengineering of intelligent animals. His predictions of personal radios, translating machines, and narrow/sub-human AI were decently accurate, and maybe even efficient electric storage, depending on how you measure it (though 2 of the 4 were way behind schedule). Incidentally, even with his, in hindsight, blatantly optimistic bias, his date of superhuman AI was way out in 2080.

[7] Technology Forecasting: The Garden of Forking Paths · Gwern.net

But conversely, even if you find a promising path that looks free of boulders and points in the right direction, it's no guarantee that the forest is traversable, nor that it will be successfully traversed within a specific time limit (e.g., 20 years).

Given the difficulty of forecasting both unseen paths and unseen obstacles, we find it prudent to approach technology forecasting with a flexible mindset, guided by two key principles:

- Avoid anchoring on any single implementation or approach
    - Continuing the analogy above, this means forecasting exploration of the forest without anchoring on a single path, whose blockage might lead us to wrongly proclaim the forest as impassible
    - This prevents us from making mistakes like a New York Times editorial in 1920 that infamously asserted that spaceflight was impossible because nothing can push against a vacuum. The editorial correctly understood Newton's laws, but failed to imagine expelled propellant. (The New York Times eventually printed a retraction as Armstrong, Aldrin, and Collins were on their way to the moon in 1969.)
- Avoid extreme confidence (unless there are good arguments otherwise; e.g., laws of physics)
    - Continuing the analogy above, this means acknowledging our lack of visibility into foggy unexplored paths, and therefore not making strong claims about the forest's traversability
    - This prevents us from making mistakes like Elon Musk, who since 2015 has claimed repeatedly that full self-driving is just a year or two away.

# What's needed for transformative AGI?

We adopt a framework of cascading conditional probabilities: if a number of subsidiary events are necessary, and cumulatively sufficient, for a larger event, the probability of the larger event may be expressed as the product of the probabilities of the subsidiary events, each conditional on preceding events. This framework has the advantages of being relatively easy to implement, well founded in theory, and functioning as a corrective to intuition in many cases, wherein the cumulative impact of many requirements may be underestimated by intuition.

The events we identify may be grouped into three major categories: software, hardware, and sociopolitical. Software events include the invention of fundamental algorithms for AGI which are currently missing, the invention of training methods for AGI that do not depend on real-time active learning, and the efficiency of AGI algorithms at performing humanlike tasks with less computation than the human brain. Hardware events include a suitable fall in the cost of computation (whether accomplished by progress in silicon computation or a new computing substrate), the scaling up of semiconductor manufacturing, and the invention and scaling of cheap, capable robot bodies. Sociopolitical events include AGI progress not being interrupted by human or AGI decisions, geopolitical competition, pandemics, or economic collapse.

In the section that follows, we explain each factor, including why we believe it's necessary, and our probability estimates. We summarize them in tabular form first:

| Event | **Forecast**<br>by 2043 or TAGI, ***conditional*** on prior steps |
|---|---|
| We invent algorithms for transformative AGI | 60% |
| We invent a way for AGIs to learn faster than humans | 40% |
| AGI inference costs drop below $25/hr (per human equivalent) | 16% |
| We invent and scale cheap, quality robots | 60% |
| We massively scale production of chips and power | 46% |
| We avoid derailment by human regulation | 70% |
| We avoid derailment by AI-caused delay | 90% |
| We avoid derailment from wars (e.g., China invades Taiwan) | 70% |
| We avoid derailment from pandemics | 90% |
| We avoid derailment from severe depressions | 95% |
| Joint odds of transformative AGI | **0.4%** |

*Source data & calculations:* **Transformative AGI by 2043: what will it take?**

# Explanations, estimates, and justifications of the elements:

## Element 1: We discover fundamental algorithmic improvements enabling computers to perform brainlike functions

### What it is and why it's necessary:

Progress in AI over the past decade has been astounding. Neural networks running on GPUs (or TPUs) have equalled or surpassed humans on difficult tasks, such as:

- Some aspects of image recognition
- Some aspects of image synthesis (e.g., GANs, DALL-E)
- Go (AlphaGo, AlphaZero) and Starcraft (AlphaStar)
- Some aspects of protein structure prediction (AlphaFold)
- Some aspects of text processing and synthesis (GPTs)

What's especially impressive about these achievements is that:

- These problems were considered difficult due to their high dimensionality
- They were solved at high performance by relatively small teams

Given this rapid rate of progress, and the accelerating investment in this space, it seems plausible that algorithms could soon advance to the point that they can produce AI able to understand and learn from the world as well as humans.

However, the mere plausibility of such advancement is no guarantee that it will occur. At present, we have only observed rapid progress on a limited number of narrow tasks and we have no clear roadmap to develop AGI. Whether progress continues at its past rate, and whether that rate is sufficient for AGI by 2043, and whether resultant AGIs approach the efficiencies of human brains, are entirely open questions.

To modestly inform those questions, we review rates of progress on four still-unsolved tasks for AI:

- Walking
- Being a worm
- Driving
- Radiology

**Walking**

Walking is not a hard task for biological neural networks. Ants have only 250,000 neurons in their entire central nervous system, and they are able not only to fuse visual, tactile, and olfactory data to walk at many body lengths per second over unprepared terrain, but also to remember pathways and navigate.

And although walking is the kind of task that should be very amenable to reinforcement learning, as with human babies learning to walk, biological neural networks often don't even require this. Ants are able to walk with very little training, getting up and going as soon as they hatch from a larval state. And yet humans still consider getting our robot bodies to walk a very hard problem, with a lot of academic and industrial work focusing on the basics. Leaders like Boston Dynamics are not using deep neural networks at all, but explicit algorithms grounded in control theory. The efforts to walk using neural networks, like Tesla's Optimus, are struggling to shuffle slowly across a room, and even they are not using monolithic neural networks, but also large amounts of explicit task-specific engineering. A recent DeepMind paper on walking[8], which was published during the drafting of this essay uses a monolithic neural network (we believe from reading their paper), and does exhibit state of the art performance for the specific robot platform it uses. However, that performance is modest (routinely tripping while walking unencumbered on flat surfaces, walking much slower than comparably sized and powered animal bodies), not state of the art for walking in general (Boston Dynamics robots do parkour routines), and in any event, was accomplished with large amounts of pretraining in a high-fidelity simulation.

Overall, progress is promising, but despite long investment we have still so far failed to solve tasks efficiently accomplished by tiny insect brains.

**Neurological simulation**

Walking is beyond us, but how about crawling/wriggling? The OpenWorm project has been working for over a decade to attempt to make computational mapping between the perceptions and actions of the C. Elegans nematode worm, a tiny organism whose cells have all been mapped, including 95 muscle cells (individually controlled) and a lilliputian 302 neurons whose connection graph has been mapped. They also have large labeled training sets wherein the worm's entire sensory state has been measured, and its subsequent actions are known. This should be easy, or at least possible. Even accounting for the considerable computational complexity of individual neurons, the total complexity of the worm's nervous system should be well within our capabilities.

As a result, a decade ago the AI community was bullish[9], with project leaders projecting success by 2014, or at latest 2020. They were bullish to such an extent that the popular press was using the impending success of the OpenWorm simulation as the rhetorical tip of the spear driving us toward the singularity[10]. And yet, after a decade of work, we can't get simulated worms to crawl or eat.

We highlight this example not because we believe AGI will come from brain simulations, but because it illustrates how we are still struggling to replicate the functions and outputs of even the tiniest brains.

Something's missing on the algorithm side.

**Self-driving**

Let's now turn to self-driving AI.

---

[8] [2304.13653] Learning Agile Soccer Skills for a Bipedal Robot with Deep Reinforcement Learning
[9] Whole Brain Emulation: No Progress on C. elegans After 10 Years - LessWrong
[10] Is This Virtual Worm the First Sign of the Singularity? - The Atlantic

**Compared to AGI, self-driving AI is child's play.** Navigating unfamiliar terrain while avoiding collisions is easy enough that baby animals and tiny insects can do it—and they aren't transformative AGIs. And driving, specifically, is a task which a variety of animals have been trained to do at (at least) rudimentary examples, such as apes[11], rats[12], and goldfish[13].

**Creating reliable self-driving AI has been consistently more difficult than expected.** Getting AI to do things *reliably* is much harder than just getting AI to do things. For a flavor of the difficulty, look at how often early unreliable tech demos have fooled forecasters into thinking a reliable solution is mere years away:

| Grade | Predicted year of self-driving | Year of prediction | Years until self-driving | Source |
|---|---|---|---|---|
| Overoptimistic | 1960 | 1940 | 20 | Magic Motorways |
| Overoptimistic | 1976 | 1956 | 20 | GM |
| Overoptimistic | 1975 | 1960 | 15 | NY Times |
| Overoptimistic | 1985 | 1965 | 20 | Herbert Simon (Nobel & Turing winner) |
| Overoptimistic | 1993 | 1973 | 20 | 25% of AI experts surveyed |
| Unknown | 2025 | 2010 | 15 | US Secretary of Transportation |
| Overoptimistic | 2017 | 2012 | 5 | Google founder |
| Overoptimistic | 2022 | 2012 | 10 | Intel CTO |
| Overoptimistic | 2020 | 2013 | 7 | Nissan CEO + EVP |
| Overoptimistic | 2017 | 2014 | 3 | Audi |
| Overoptimistic | 2017–2020 | 2014 | 3–6 | Waymo director |
| Unknown | 2024 | 2014 | 10 | Jaguar & Land Rover |
| Overoptimistic | 2017 | 2015 | 2 | Tesla CEO |
| Overoptimistic | 2020 | 2015 | 5 | Ford CEO |
| Unknown | 2030 | 2015 | 15 | Uber CEO |
| Overoptimistic | 2019 | 2016 | 3 | Baidu Chief Scientist |
| Overoptimistic | 2019 | 2016 | 3 | Volkswagen |
| Overoptimistic | 2019 | 2016 | 3 | Delphi and MobilEye |
| Overoptimistic | 2020 | 2016 | 4 | Stanford AI100 Standing Committee |
| Overoptimistic | 2021 | 2016 | 5 | Ford CEO |
| Overoptimistic | 2021 | 2016 | 5 | BMW CEO |
| Overoptimistic | 2021 | 2016 | 5 | Lyft CEO |

[11] Orangutan Driving Golf Cart
[12] Rats love driving tiny cars, even when they don't get treats | Ars Technica
[13] Israeli scientists have trained goldfish to drive, in a scene out of a Dr. Seuss book

| | | | | |
|---|---|---|---|---|
| Overoptimistic | 2019 | 2017 | 2 | Tesla CEO |
| Overoptimistic | 2020 | 2017 | 3 | Audi |
| Overoptimistic | 2021 | 2017 | 4 | NVIDIA CEO |
| Overoptimistic | 2019 | 2018 | 1 | Tesla CEO |
| Overoptimistic | 2020 | 2019 | 1 | Tesla CEO |
| Overoptimistic | 2021 | 2020 | 1 | Tesla CEO |
| Overoptimistic | 2022 | 2021 | 1 | Tesla CEO |
| Overoptimistic | 2022 | 2022 | 0 | Tesla CEO |
| Unknown | 2023 | 2023 | 0 | Tesla CEO |

*Note 1: Many of these predictions are "by year X" rather than "in year X."*
*Note 2: We take 'widespread' or 'available for sale' to be implicit in some of these predictions. If you do not, then testing and pilot taxi programs may qualify a subset of these predictions as true.*

**Investors are increasingly pessimistic about the odds of success.** Even with a trillion[14] dollar market dangling as the prize for anyone who can crack self-driving, and the current wild pace of AI progress, investors are pessimistic about whether we can soon teach AIs to drive cars reliably.

- Uber pulled out of self-driving in 2020[15]
- Lyft pulled out of self-driving in 2021[16]
- Argo AI shut down in 2022[17]
- Comma.ai's founder called self-driving "a scam"[18] and left his company in 2022[19]
- Self-driving startup valuations have declined 81% in the past two years ending Oct 10, 2022[20]
- Cruise is valued at only $21B[21] after $15B invested[22]
- Even the market leader Waymo is struggling enough that Alphabet has turned to outside funding, at a valuation of only $30B[23] (Alphabet had $125B cash on hand in late 2022, which is plenty to fund Waymo if it expected a return[24])

---

[14] Roughly, we estimate the global self-driving market is $2T: ~1B passenger cars * 1 hr/day + ~50M commercial vehicles * 5 hrs/day all at $5/hr
[15] Uber Gives Up on the Self-Driving Dream | WIRED
[16] Lyft is getting out of the self-driving business | Ars Technica
[17] Ford, VW-backed Argo AI is shutting down | TechCrunch
[18] George Hotz is on a hacker crusade against the 'scam' of self-driving cars - The Verge
[19] https://geohot.github.io//blog/jekyll/update/2022/10/29/the-heroes-journey.html
[20] https://www.forbes.com/sites/johnkoetsier/2022/10/17/self-driving-startups-have-lost-40-billion-in-stock-market-valuation-in-2-years/?sh=6eb147b63337
[21] What Pawn Stores Are Saying About the Economy | Barron's
[22] Cruise - Funding, Financials, Valuation & Investors
[23] Waymo raises $2.5 billion in second external funding round - The Verge
[24] Alphabet Cash on Hand 2010-2023 | GOOGL | MacroTrends

No one is better positioned to predict the success of self-driving AI than self-driving companies and their investors, who have spent ~$100B[25] over the past decade as they've researched better self-driving AI technology. Yet, with all of this information, many of them have decided to give up entirely.

Despite magic demos and promising disengagement trends, improving the general intelligence of AI to handle long-tail scenarios has been expensive, slow, and hard, even with modern hardware and modern machine learning. We have been trying to teach algorithms to drive for far longer than we teach teenagers, and still, we have not succeeded. It's not obvious we'll have human-cost-and-quality AI drivers by 2043.

One lesson to take from this experience is that early progress in a field should not be taken as strong evidence that all its difficulties will be surmounted. Artificial intelligence is brittle in a way human brains are not, and often it takes a surprising amount of work to handle the long tail of edge cases.

Part of the reason that self-driving has taken so long compared to chess or Go is that its reinforcement learning loop is expensive and slow. What makes chess and Go so amenable to AI training is that learning loops are incredibly cheap, as (1) taking actions is very cheap (as games are computerized) and (2) calculating the reward is very cheap (as a simple function can calculate the winner). An AI can be horribly sample inefficient relative to a human, but by sheer volume of games (AlphaZero played 44M games in 9 hours[26]), it still quickly surpasses humanity. With self-driving, such cheap training is not an option. Acquiring training data costs billions of dollars and years of time, as companies staff up large fleet operations. This is partly why AIs cannot solve self-driving in 9 hours, as they did chess. Instead of 9 hours, Waymo has been at it for 14 years, without yet achieving human-level performance, let alone human-level cost. A big reason is that the active training loop is expensive.[27]

It's important to note that these issues, a long tail of unusual situations and much more difficult reinforcement loops than tasks that have recently fallen to NNs, are both true of many important tasks.

### Radiology

In 2016, machine learning luminary Geoffrey Hinton proclaimed:

> "*Let me start by just saying a few things that seem obvious. I think if you work as a radiologist, you're like the coyote that's already over the edge of the cliff, but hasn't yet looked down, so doesn't realize there's no ground underneath him. People should stop training radiologists now. It's just completely obvious that within 5 years deep learning is going to do better than radiologists, because it's going to get a lot more experience. It might be 10 years, but we've got plenty of radiologists already.*"[28]

---

[25] Even After $100 Billion, Self-Driving Cars Are Going Nowhere - Bloomberg
[26] Do AlphaZero/MuZero learn faster in terms of number of games played than humans? - Artificial Intelligence Stack Exchange
[27] Every large self-driving effort relies on high quantities of cheap simulation to augment its expensive real-world data. Unfortunately simulation is not a complete solution, as the simulations are only as good as their models of reality, and good models of reality is a big part of what is trying to be learned.
[28] Geoff Hinton: On Radiology

This was a “completely obvious” prediction from a world-renowned expert about a narrow AI task that already had significant research traction, on a timeline of only 5–10 years.

How much of it came to pass?

Very little.

7 years later, AI tools are now used regularly by radiologists, and ~200 algorithms have been approved by the FDA.[29] But so far, these tools appear to be complements to human labor, not substitutes. In 2022, the RSNA reported a global radiologist *shortage*, caused by training pipelines not keeping pace with the elderly population.[30] ACR job postings have been rising, not falling:[31]

Figure 1.1
**New Radiology Job Postings**
**ACR Job Board, By Quarter**

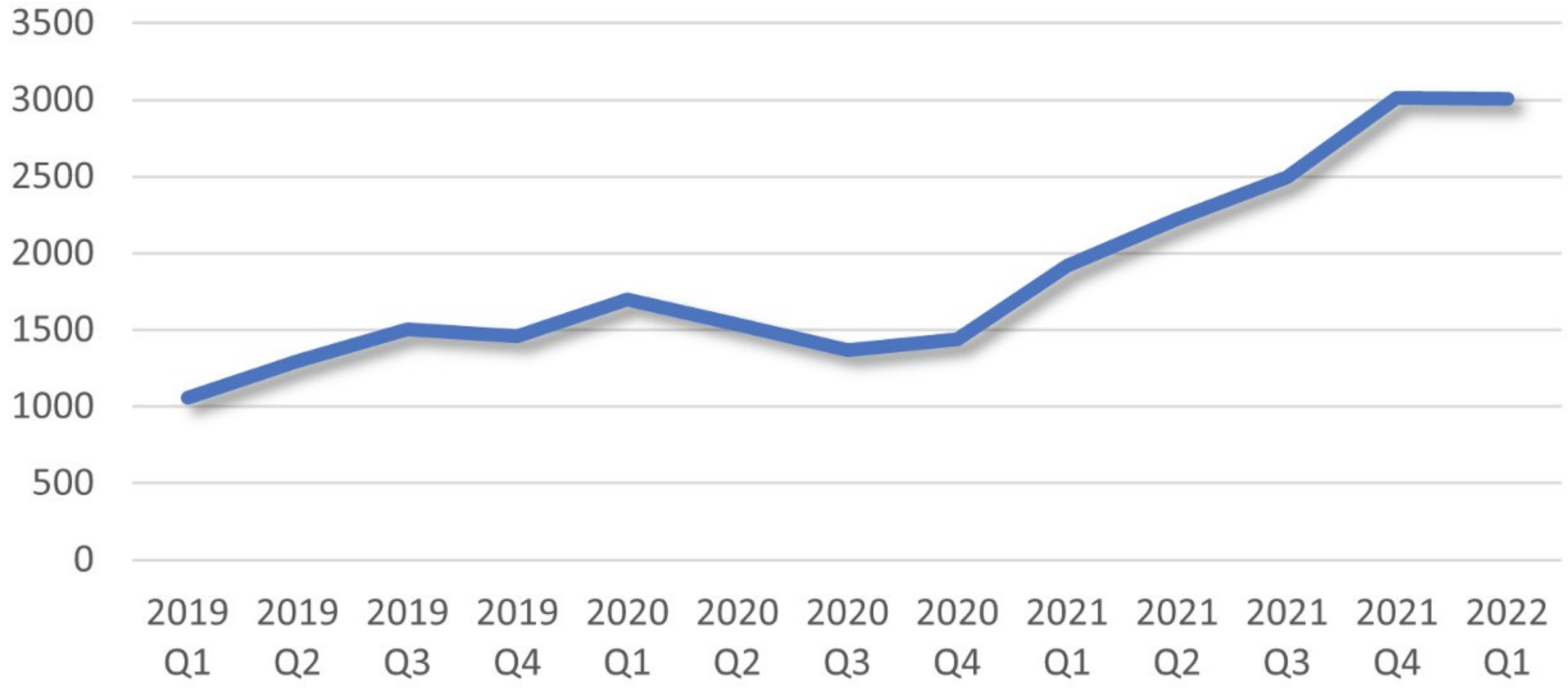

Source: vRad analysis of ACR job board data, 2019-2022

Clearly radiology is still a career, at least for now.

What did Hinton and so many others get wrong?

---

[29] [This radiologist is helping doctors see through the hype to an AI future - The Reporter | UAB]
[30] [Radiology Facing a Global Shortage]
[31] https://twitter.com/erikbryn/status/1662564885589561344?s=20

Kevin Fischer, who obtained one of the first approvals in machine learning in radiology, gives the following diagnosis:[32]

> *Here's a bunch of stuff you wouldn't know about radiologists unless you built an AI company WITH them instead of opining about their job disappearing from an ivory tower.*
>
> *(1) Radiologists are NOT performing 2d pattern recognition - they have a 3d world model of the brain and its physical dynamics in their head. The motion and behavior of their brain to various traumas informs their prediction of hemorrhage determination.*
>
> *(2) Radiologists have a whole host of grounded models to make determinations, and actually, one of the most important first order determination they make is whether there is anything notably wrong with a brain structure that "feels" off. As a result, classifiers aren't actually performing the same task even as radiologists.*
>
> *(3) Radiologists, because they have a grounded brain model, only need to see a single example of a rare and obscure condition to both remember it and identify it in the future. This long tail of rare conditions to avoid missing is a large part of their training, and no one has any clue how to make a model that acts similar in this way.*
>
> *(4) There's so many ways to make Radiologist lives easier instead of just replacing them, it doesn't even make sense to try. I interviewed and hired 25 radiologists, whose primary and chief complaint was that they had to reboot their computers several times a day.*
>
> *(5) A large part of the radiologist job is communicating their findings with physicians, so if you are thinking about automating them away you also need to understand the complex interactions between them and different clinics, which often are unique.*
>
> *(6) Every hospital is a snowflake, data is held under lock and key, so your algorithm might not work in a bunch of hospitals. Worse, the imagenet datasets have such wildly different feature sets they don't do much for pretraining for you.*
>
> *(7) Have you ever tried to make anything in healthcare? The entire system is optimized to avoid introducing any harm to patients - explaining the ramifications of that would take an entire book, but suffice to say even if you had an algorithm that could automate away radiologists I don't even know if you could create a viable adoption strategy in the US regulatory environment.*
>
> *(8) The reality is that for every application, the amount of specific and UNKNOWABLE domain knowledge is immense.*

---

32
https://twitter.com/kevinafischer/status/1662853371118641154?s=46&t=4rpMhoHJ70JGj3o1SOGEyw&fbclid=IwAR3seSIM_yKaBoSXWVwdrgPa4HKyTayZAKMQyl_KVvbMc_lse2m20dchAio

*LONG STORY SHORT: thinkers have a pattern where they are so divorced from implementation details that applications seem trivial, when in reality, the small details are exactly where value accrues.*

In general, reality has a surprising amount of detail.[33] If you lack knowledge of those details, it's easy to accidentally be biased into thinking automation will be quick. This effect likely explains why if you ask AI researchers to predict when AGI can do AI research, they give longer timelines than when you ask them to predict when AGI can do *all human tasks*! (Incidentally, if taken seriously, their forecasts are a negative sign for the likelihood of scenarios relying on autocatalyzing AGI improvement.)

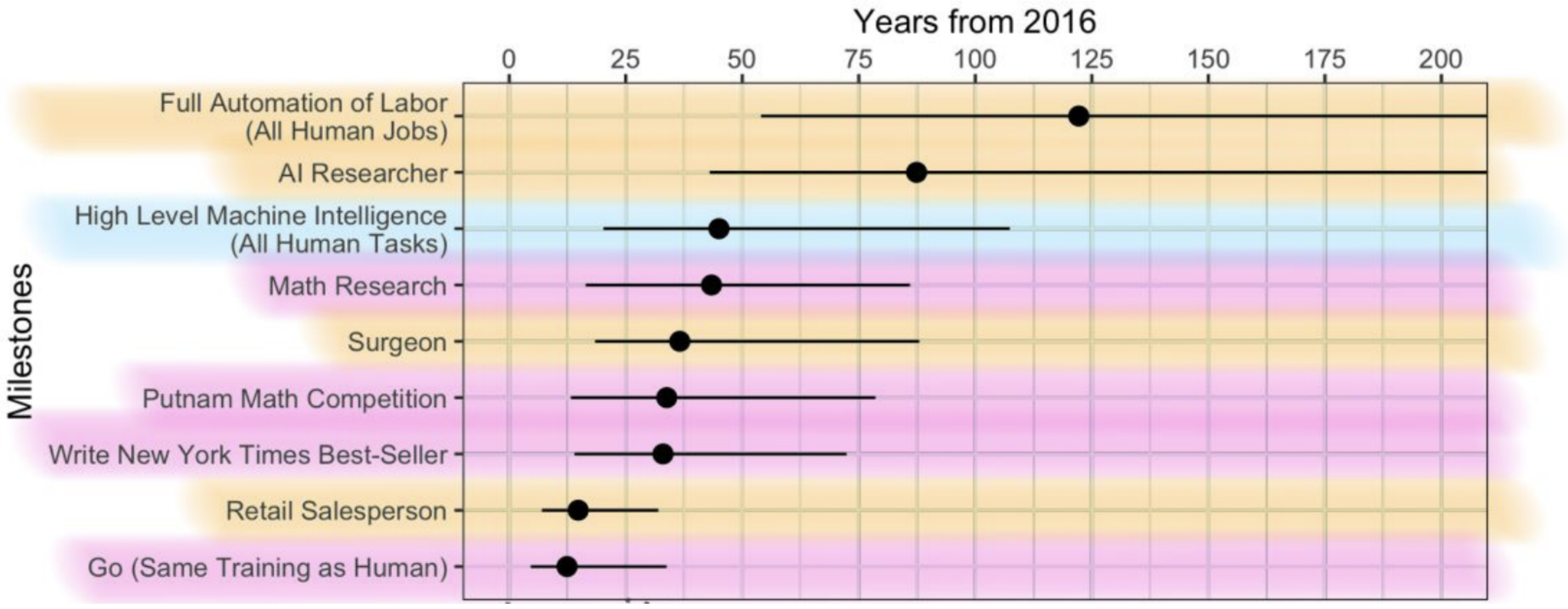

*Source: AI Impacts' 2016 Expert Survey on Progress in AI*[34]

Not only do machine learning practitioners need to be wary of neglecting all the fine details of human work, they also need to be wary of assuming that benchmark results will translate into real-world results. A nice Nature article reviews why impressive radiology benchmark performance had trouble translating into real-world success, and warns of many biases:[35]

- Training sets biased toward easy-to-collect data
- Noisy improvements that revert to the mean when deployed
- Leakage of information between train and test set
  - this is a more difficult problem than often supposed; e.g., if an individual is scanned multiple times and train and test are randomized by image, the same patient can appear in both sets, albeit with different scans; very often not every correlate is logged which makes perfect decontamination impossible
- Overfitting by observer / garden of forking paths (i.e., no preregistration of hyperparameter search)
- Publication bias
- Metrics not aligned with patient value

---

[33] Reality has a surprising amount of detail

[34] Some survey results! – AI Impacts. Note that later surveys have increased optimism about AGI. We share this older survey as the later ones didn't appear to compare AI researcher vs all human tasks.

[35] Machine learning for medical imaging: methodological failures and recommendations for the future | npj Digital Medicine

- A lack of statistical significance

So although progress by deep neural networks on benchmarks looked promising, it ended up taking years longer than expected for those algorithms to succeed, as all sorts of biases caused real-world performance, even when successful, to be more fragile and less reliable than the benchmarks indicated. And the success of these algorithms have so far been limited to productivity-enhancing tools, not seismic shifts to the labor market.

**What does this mean for AGI timelines?**

Summarizing four active areas of AI R&D:

- In walking, Boston Dynamics has been working for thirty years to achieve pretty good performance with extensive task-specific engineering of explicit algorithms, while monolithic NNs have only recently become able to achieve anything, and only with the aid of high fidelity simulations
- In neurological simulation, we cannot make worms crawl or eat even with explicit maps of their entire nervous systems and large training data, despite recently believing we would be able to do this
- In self-driving, it’s been 16 years since the promising DARPA 2007 Urban Challenge inspired Waymo and others to invest billions in self-driving AI; and yet there’s been no impact to the taxi job market
- In radiology, it’s been 11 years since the promising results of AlexNet on the ImageNet 2012 Challenge; and yet there’s been little impact to the radiologist job market

These tasks should be far, far easier than anything approaching AGI, let alone transformative AGI. And yet none bore fruit in their first decades of often heavy investment.

And while worms are not a large market, the other three tasks are large markets and two of them even have abundant training data. Yet we’ve comprehensively failed to make AI walkers, AI drivers, or AI radiologists despite massive effort.

This must be taken as a bearish signal. If it’s taking this long for common jobs like driver or radiologist, just imagine how much harder it will be to train an AI to perform jobs like summer camp counselor, for which there is little training data.

Generative Pretrained Transformers, for all their success and intelligence, do not satisfy this challenge. They are capable of impressive feats of recall, creativity, and what looks like understanding, but remain woefully unreliable. For example, a nonprofit organization which built a chatbot to counsel eating disorder patients shut it down within days of launch because it was advising patients to undertake behaviors that would exacerbate their anorexia[36], a behavior the organization said it had not seen in thousands of test conversations but which emerged instantly in real world use. GPT-4’s capabilities are inspirational, but the lack of solutions thus far to its unreliability (from hallucinations to instruction following to alignment) could caution one from ascribing near-certain probability to the proposition that these problems will be shortly solved.[37]

---

[36] https://www.theguardian.com/technology/2023/may/31/eating-disorder-hotline-union-ai-chatbot-harm

Something fundamental is missing on the algorithm side.

For transformative AGI to be achieved by 2043, it will require something new, something we don't have. A type of AI development that is truly *general* in character, to handle in a graceful way the edge cases that have stymied narrow AIs, attain traction on problems that current NNs do not get traction on, and thereby permit the development of algorithms for many diverse tasks without ideal conditions like giant datasets of task-specific information, high fidelity simulations for active learning, or billions of dollars of task-specific engineering effort over decades. If those kinds of things are still required, then we won't get transformative AGI by 2043. This is what we refer to as the fundamental algorithm problem of AGI.

Whatever this will be, it will definitely involve massive progress in AI sample efficiency, massive efforts to gather training feedback, and massive parallelization across jobs and industries. This is not impossible, but our key point is that **we do not have it now, and it is not guaranteed to emerge on convenient timelines.**

Another key bearish signal on the emergence of this missing algorithm matter is that it's not even clear precisely what it is, or even that it's a single major thing. What unites our inability to simulate worms, and our inability to make robot bodies walk reliably, and our trouble handling the long tail of driving situations, and our inability to generalize radiology models, and our inability to make GPTs reliable, and all the other tasks we "should" be able to do, but can't? Is it network architectures, optimizers, training loop or dataset construction methods, or higher-level concepts that sit outside the network-training-inference paradigm of ML development? What, precisely, should people be working on in order to materialize this dark matter of AI, and can we really be sure they will succeed within the relevant timeframe?

## Our estimate and justification:

We estimate optimistically that there is a 60% chance that all the fundamental algorithmic improvements needed for AGI will be developed on a suitable timeline. The development of both GANs and the Transformer within the first decade of GPU computing is an optimistic note, and there are still twenty years left to work on new algorithms and assemble unprecedented quantities of compute. However, decades of failure in a variety of important and well-explored tasks, and the lack of even a clear problem statement, are still vexing indications of what's missing, and make it hard for us to forecast anything resembling 80% or higher.

---

[37] Obviously, we expect continued rapid progress in these fields and would not be surprised by self-driving achieving commercial success this decade, GPT-X looking brilliant next to GPT-4, or commercial medical imaging models achieving breakthrough successes. One of the authors regularly takes self-driving Cruises, which, despite being presently subhuman in terms of performance and cost and generalization ability, work marvelously compared to self-driving vehicles from years prior.

# Element 2: We quickly figure out how to train computers to do human tasks without sequential reinforcement learning

## What it is and why it's necessary:

Transformative AGI by 2043 depends critically on the development of non-sequential reinforcement learning training methods with no real human analogue.

Imagine trying to raise a human child from infancy to adulthood strapped to a chair, without letting them move or interact with the environment or other people, just playing them sensory input. Is there some combination of movies, sounds, tastes and massage which could enable a child raised this way to step out of their chair, walk away, speak, write, and go off to college? If so, finding it would be hard, because this is not how humans learn, at all.

Humans learn almost every important thing by doing sequential, real-time, reinforcement learning with a variety of tasks, and in particular versatile transfer learning from task to task. It begins with learning to see and hear, walk and talk, read and write, and through more complex tasks, over a period of (approximately) thirty years, in the case of complex, economically valuable cognitive and skilled labor tasks. This is also the way that many of our most significant AI achievements in recent years have been training themselves, accelerated by the use of accurate simulations (or precisely specified symbolic tasks). This includes AlphaGo, AlphaZero, AlphaStar, and DeepMind's recent paper on walking[38].

This kind of learning is great, because it's so easy to construct training sets; they construct themselves. Once we have artificial human-style intelligences, particularly in robot bodies, we could start didactically training them for every task we could imagine, and in particular, making the process efficient by checkpointing at versatile states that are amenable to subsequent training in a lot of different tasks. Make an AI high school graduate, and you can send it to college to study nearly any subject, in parallel if you want. Start with an AI with a bachelors' degree in Chemistry and you can send it to graduate school to learn any subfield of chemistry, in parallel if you want. Start with an AI with a Ph.D. in medicinal chemistry, and you can set hundreds of them loose in a pharmaceutical company to acclimatize to their new roles.

But this kind of learning makes it hard to accelerate learning beyond the inherent timeline of the active learning loop, because it's sequential. Run computers faster or put more systems together in parallel, and you may be able to learn faster from each new piece of information as it comes in, but the speed of the training loop is controlled by the sequential learning processes, not the computation. An AI chef can't bake a hundred souffles one at a time, refining its technique after each one, any faster than it takes a human to bake a hundred souffles one by one.[39] Nor are we likely to make massive improvement in the amount of active learning it takes; humans have been under tremendous evolutionary pressure to make development faster to arrive at the state we're in, and still take many times longer to arrive at adulthood than almost any other animal. And humans are already much better at low-information learning than existing AI algorithms.

---

[38] [2304.13653] Learning Agile Soccer Skills for a Bipedal Robot with Deep Reinforcement Learning
[39] An AI will be able to parallelize some tasks, however. E.g., a blind grid search over ingredient ratios is amenable to parallelization, whereas gradient-descent-type experiments that depend on the results of prior experiments will be less amenable. We envision many skills as requiring trees or graphs of subskills. Many branches can be explored in parallel, but the branches that are long and sequential will be difficult to speed up.

This kind of learning, the only type for which we have an existence proof, could absolutely power an AGI revolution. But not in 2043. Even if we developed AI systems that were just as capable of learning and generalizing to arbitrary tasks as humans, and well in advance of 2043, we would miss a 2043 deadline to perform a wide variety of complex human tasks, because the way we train humans to perform these tasks involves 20–30 years of sequential reinforcement learning. There's not even enough time before 2043 to do this starting now, but because of the dependency of AGI on progress in semiconductors, discussed below, the hardware needed for human scale models wouldn't emerge until only shortly prior to 2043. We'd invent AGI-enabling algorithms and hardware by 2043, but not have AGI outcomes, the chemists and lawyers and so on, until 2073.

To have transformative AGI by 2043, we'd have to take the sequential real-world training loop out of the training process. This would be a novel, very significant step over and above the development of learning agents with human level capabilities. It would mean synthetically constructing self-supervised training sets and/or simulated reinforcement learning loops for all the tasks we wanted these AIs to perform, that were capable of being trained on quickly, and the algorithms that would train them into a useful state without much active learning. Quite a challenge!

Of course, this is by no means impossible. We envision several ways that it might be done. For example:

- We could build realistic simulations of the tasks that AGIs would need to perform, and let them perform reinforcement learning in those simulations. This was the approach taken by the recent DeepMind paper on walking. This would be tough because of the wide range of tasks involved and the difficulty of simulating them.
- We could build huge corpuses of real world data on tasks and actions, the analogue of text and image pretraining corpuses used by many of our models, and hope that pretraining on them could yield a "collegiate" AGI which could be trained on many advanced tasks quickly. This might involve putting cameras in many places to watch humans do things, or even putting biometric sensors on humans to follow their actions. It's hard to know whether this would work or whether humans would even want to do this.
- We could do real world reinforcement learning in parallel, with many agents acting from and updating a shared model. This would obviously suffer somewhat in learning rate per unit total experience because many parallel steps would occur before updating happened (analogous to making 100 souffles from one bad recipe or technique before updating, instead of updating each time), but could still offer speedups relative to serial active learning, depending on the nature of

---

This is especially true in cases where the training is bottlenecked by human supervision; e.g., a manager explaining the policy for restocking shelves. Listening to the manager's explanation and asking follow up questions is not something that can be parallelized. Humans are able to parallelize work by teaching different people to take different jobs and acquire different specializations. An AGI should have no problem doing this as well. But for humans it takes many years to train a specialized workforce, a luxury we may lack if, say, the seeds of transformative AGI are planted in 2035 and we need all work automatable/automated by 2043.

the task. It’s hard to know how much, and there may be long tails of skills that require more sequential learning than others.

And perhaps there are other approaches we can’t think of now. But one of these approaches, or a combination of them, or something else, would have to be riotously successful to enable AGIs to train quickly for all the subtle tasks involved in a modern economy, and permit robots to step into the shoes of employees in any given job, prior to 2043.

Whatever form this accelerated learning paradigm took, it would have to be accomplished quickly, while in the knowledge that AGI would arrive in a generation even without it. Even if it were possible to find a way to train AGIs without sequential reinforcement learning, is that the approach humans will actually take, when AGIs are available and can be trained to do anything in a human-like, pedagogical fashion with sequential reinforcement learning, and doing so without it is still hard and uncertain?

## Our estimate and justification:

We think this is actually unlikely to happen. Recent successes in agent-based tasks like games have shown the superiority of even de-novo sequential reinforcement learning (self play) over even the largest and highest quality corpuses of games in the absence of sequential reinforcement learning, and we think this is likely to translate. Even if the algorithm side would work, the idea of constructing a corpus of experience at open-ended, not-well-defined tasks of the kinds that constitute many economically useful tasks, and the wide variety and incredible number of such tasks, quickly, seems remote. We estimate the probability of this happening at 40%, and this may be an overestimate.

# Element 3: Efficiency of AGI computation achieves that of humans (e.g., <$25/hr)

Or, restated: Software and hardware efficiencies combine to surpass current computation cost efficiency, and/or the efficiency of the human brain, by at least five orders of magnitude.

## What it is:

AGIs will be software which runs on hardware; very complex software which will require powerful hardware to run. Even if all the algorithmic and data problems are solved, and AGI then becomes possible, in order to be transformative AGI, the substrate to train human scale AGIs for many tasks, and perform many billions of hours of inference, must be available and cheap. It must be possible to do these things with cost-effective quantities of computational resources.

## Why it's necessary:

Several experts have recently generated estimates suggesting that the computational intensity of AGI training and inference tasks is surprisingly modest and could easily be accomplished soon. We will discuss three such estimates, coming from Ajeya Cotra, Joseph Carlsmith, and Cerebras. Their overall approaches are sound, and somewhat convergent, and there is much good to say about them. However, we will explain why we think they are making massive underestimates of intensity (many orders of magnitude). As a result, we estimate that at the cost-efficiency of today's silicon, and the computational intensity of the human brain, a human-level AGI, if available today, might cost about $1 million per hour to run. This means that about five orders of magnitude of improvement would be needed to make AGI economical enough for transformative AGI to emerge.

Like these experts, we are *not* arguing that AGI will look like human brains or work by emulating human brains. AGI may be as alien to humans as airplanes are to birds.[40] But to answer the question of how much compute a human-brain-level AGI will need, there is no better starting point than estimating the compute performed by a human brain. And of course, in the same way that airplanes have quite different requirements and tradeoffs than birds, early AGIs may end up needing far less (or far more) compute than a human brain performs. Nevertheless, human brains are the best starting point we have, so that's where we will start.

### Review of Cotra

In a lengthy whitepaper[41], Cotra estimates the number of floating-point operations needed to train a human-or-superhuman AGI model. She uses several conceptual anchors ranging from the lifespan of a human (the "lifetime anchor"), through neural network training with estimates of training corpus size derived from network size plus a time horizon (the "neural network anchors"), through to the total computation performed by all animals during evolutionary history (the “evolutionary anchor”) to estimate the amount of human-equivalent computation that training an AGI model might require. Due to the massive dispersion between one human lifespan and the total lifespan of all animals in history, this results in a spectacular range of approximately 1e24 to 1e50 floating-point operations.

---

[40] Planes are still decades away from displacing most bird jobs - Alexey Guzey
[41] 2020 Draft Report on Biological Anchors

However, these estimates are all based on a common, relatively narrowly-specified estimate of the computational intensity of the human brain. Cotra notes that the brain has about 1e15 synapses, each of which fires at, on average, a frequency of about once per second. She assumes the computational activity of each synaptic firing is one floating-point operations, and thereby estimates the human brain is performing about 1e15 FLOPS of computation, and derives all her other estimates from this estimate, although she uses a tenfold fudge factor to arrive at 1e16, a number she uses in several calculations.

**Review of Carlsmith**

In a lengthy whitepaper[42], Carlsmith attempts to estimate the number of FLOPS of computation being performed in a human brain. He uses a number of different methods, which he terms “mechanistic” (based on the number of neurons and estimates of computation performed per neuron), “functional” (based on the known complexity of computational tasks being performed by certain known brain regions), “limit” (based on the inherent energetic cost of bit erasure, and the energy dissipated by the human brain), and “communication” (based on the approximate amount of communication of information through the brain, analogized to memory bandwidth in a silicon computer).

Carlsmith finds that these four methods converge on a broad range of approximately 1e15 FLOPS to 1e21 FLOPS of computation occurring in the human brain, i.e. from approximately the value adopted by Cotra to about six orders of magnitude higher. However, due to the scope of his inquiry, Carlsmith does not make estimates of a computational cost to train a brain-equivalent model.

**Review of Cerebras**

In a whitepaper about very large scaling of ML model training[43], Cerebras estimates the amount of computation needed to train human scale models. They define “human-scale” models by anatomic analogy; a human scale model has as many “neurons” as the human brain has neurons, and as many parameters as the human brain has synapses. With their latest scaling results, they suggest that if 1000 CS-2 wafer-scale engines were to engage in distributed training with a weight sparsity of 10%, they could train a transformer language model of this scale, on a corpus about the size of the one used to train GPT-3, in about a year. That is, they estimate that this human-scale model could be trained with about 1000 wafer-years of computation. The CS-2 has a capacity of about 5e15 16-bit floating point operations per second, so this is an estimate that the human scale model might take about 8e26 floating-point operations to train, in total. To put it another way, since the CS-2 costs about $1.6 million per year to lease[44], this could imply that this scale of model could, now, be trained for about $1.6 billion.

---

[42] How Much Computational Power Does It Take to Match the Human Brain? - Open Philanthropy
[43] Scaling Up and Out: Training Massive Models on Cerebras Systems using Weight Streaming
[44] Cerebras Brings Its Wafer-Scale Engine AI System to the Cloud

**Why we think Cotra, Carlsmith and Cerebras are too optimistic**

We think all three of these models have problems which lead them to dramatic underestimates (with the exception of the upper estimates in Cotra). There are several points where they make unrealistically low assumptions, not all of which affect each estimate.

- Each of the three estimates the computational activity of the human brain by analogizing real neurons and synapses to our computational "neurons" and single parameters. An artificial neural network parameter is a single number, and a "neuron" is a single nonlinear activation function with a number of scalar weights. By contrast, a synapse is not a single number but a complex biological assembly containing hundreds of proteins, which comes in many different varieties containing about two thousand variable proteins, which would be unnecessary if its function were as simple as a single scalar[45]. And individual real neurons perform complex combinatorial activity which single computational "neurons" cannot. A recent attempt by Beniaguev et al to estimate the computational complexity of a biological neuron used neural networks to predict in-vitro data on the signal activity of a pyramidal neuron (the most common kind in the human brain) and found that it took a neural network with about 1000 computational "neurons" and hundreds of thousands of parameters, trained on a modern GPU for several days, to replicate its function[46]. Beniaguev et al postdates Cotra's whitepaper, so she could not have taken it into account. But it is cited by Carlsmith, resulting in an estimate at the very top of Carlsmith's range, several orders of magnitude higher than his other "mechanistic" estimates. Because neurons are much more complex than the simple assumptions made by all three groups, the scale of artificial neural network model which is actually equivalent in computational complexity to the human brain is probably several orders of magnitude higher than the estimates of Cotra and Cerebras, and close to the upper limit of Carlsmith's range.

- Carlsmith and Cotra both assume that the computation performed by synapses can be estimated in relation to the number of times they fire, i.e. approximately one time per second. However, neurons do not perform computation only when they fire; they perform computation in relation to the decision about whether to fire. Reckoning only by the number of firings is like estimating file sizes by counting the 1s, or estimating the computational needs of a classifying model only by the number of positive examples it will classify. Since a neuron is capable of firing approximately 100 times per second, both Carlsmith and Cotra are underestimating the computation performed in the brain by about two orders of magnitude on this account alone.

- Cerebras assumes that a human-scale model, trained to perform human-scale tasks, could be trained with a training set about the same size as the corpus that went into GPT-3 or other models that exist today. GPT-3 was trained on 300 billion tokens, each token equal to about 16 bits, for about 5e12 bits of training data[47]. Cotra, commendably, does not do this. Her Neural Network anchors make assumptions about corpus size based on scaling laws from network complexity, plus the time horizon of an example involved in training, which are actually, we believe, much too severe, as we will discuss below. Carlsmith does not make estimates of corpus size, because it is out of the scope of his inquiry.

---

[45] Grant 2018 Synapse molecular complexity and the plasticity behaviour problem - PMC
[46] Beniaguev et al 2021 Single cortical neurons as deep artificial neural networks - ScienceDirect
[47] Brown et al 2020, [2005.14165] Language Models are Few-Shot Learners

- Cerebras assumes that (at least) tenfold sparsity is free. It starts with a parameter count based on the human brain, and the human brain contains 100% of the synapses that it does contain, not 10%. If this much sparsity were free, the human brain would already have configured itself this way.

In general, we find that all three analyses have made generous assumptions, granting themselves "fudge factors" where multiple orders of magnitude disappear in an instant, producing numbers which are dramatically too low. Cotra and Carlsmith both conclude that the computation performed by the human brain either is (Cotra) or may be (Carlsmith) as low as the level (1e15 FLOPS) of current GPU hardware like the NVIDIA H100[48], an astonishing claim. We assert these values are much too low.

On one point, however, we actually think that one of the three analyses, Cotra, is far too severe. While Cotra estimates a corpus of training data, divided into examples, based on network complexity, a real AGI is unlikely to be trained on a corpus of discrete examples, like supervised networks today, and more likely to be trained on large, continuous blocks of data which are not divided into discrete units. The actual human brain, for example, is trained on only one "example," of lifelong length, but which involves sequential reinforcement learning feedback.

Essentially, we think that the "lifetime anchor," Cotra's lowest, is likely to be correct, since human brains actually do train on this amount of data and perform human tasks. While it is far from certain whether we will invent AGI algorithms that can do this, we theoretically could, since humans do do this. And while it's true, as Cotra points out, that human brains are pre-conditioned by the training of organisms from which they descended, the mechanism by which this pre-conditioning is transmitted to them, the human genome, is of relatively low size. It comprises only billions of bits, with significant repetition and low-information content, an implausibly low number to usefully contain the results of many orders of magnitude of computation more than a human brain actually performs. And in addition, humans, as we train AGIs, will be able to transmit information to them if we think it will be effective, provided we have it, including information about genome biology.

For a quick, top-down estimate of the amount of data a human brain trains on, we may consider the number of nerve fiber inputs received by the human brain from its sensory neurons. Human nerves contain on the order of 2400 fibers per square millimeter[49], while the area of the human spinal cord is about 137 square millimeters[50], so the human spinal cord may contain about 300,000 nerve fibers. In addition to the spinal cord, the optic nerve alone is estimated to contain about one million fibers per eye[51]. This still leaves the auditory nerves, olfactory nerves, and others. So, one million fibers is a conservative estimate. With neurons capable of firing about 100 times per second, a nerve fiber may be considered to transmit approximately 100 bits per second, and the amount of data collected and processed by the human brain could be approximated at 1e8 bits per second, which for a training period 16 hours a day over 30 years (the amount of time needed to train the human brain for many complex, economically valuable functions) is about 5e16 bits. This is about 10,000 times larger than the training set assumed in the Cerebras scaling whitepaper, but much smaller than the assumptions in the “Neural Network” and “Evolutionary” anchors of Cotra.

---

[48] NVIDIA H100 Tensor Core GPU
[49] Liu et al 2015, The diameters and number of nerve fibers in spinal nerve roots - PMC
[50] Frostell et al 2016, A Review of the Segmental Diameter of the Healthy Human Spinal Cord
[51] Jonas et al 1992, Human optic nerve fiber count and optic disc size

Updating all of these assumptions, we could make the following estimates:

Starting from Cotra and Carlsmith, we could say that the human brain is performing not 1e15 to 1e16 FLOPS, but approximately 1e20 to 1e21. We gain about three orders of magnitude from Cotra and Carlsmith’s underestimate of the complexity of neurons, and about two from their underestimate of the frequency with which neurons calculate whether to fire. This puts us within the upper end of the range of Carlsmith’s estimate.

Then, adopting the “lifetime” anchor of Cotra, and training for about thirty subjective years, we arrive at about 1e30 floating-point operations, very approximately, to train a human-scale AGI, assuming the fundamental algorithmic problems had been addressed.

Or, starting from Cerebras, we could say that:

- The brain’s computational complexity may be approximately 1000 times larger than Cerebras estimates, due to a real neuron being about 1000 times as complex as a “neuron.”
- The training set used by the human brain over the course of its development may be about 10,000 times larger than the training set example assumed by Cerebras.
- Cerebras’s estimate of training speed is tenfold too fast, due to their sparsity estimate.

So, we may estimate Cerebras was too optimistic by about eight orders of magnitude, and a true human-scale model may take about 8e34 floating-point operations to train. This number is several orders of magnitude higher than the estimate derived from Cotra and Carlsmith, possibly because Cerebras is reckoning in terms of the real speed to train models on actual hardware with all kinds of bottlenecks, not merely floating-point operations estimates. We thus consider the Cerebras-based estimates more likely to be closer to correct.

Note that these estimates, even at the upper end, are not pushing upward on the ranges estimated by Cotra or Carlsmith; we sit at approximately the top end of Carlsmith’s human brain FLOPS estimate, and actually toward the lower-middle area of Cotra’s estimate of the amount of computation needed to train an AGI. And in addition, we do not challenge Cerebras’s estimate that a thousand CS-2s could train a 1e14-parameter transformer on 300 billion tokens in a year. We merely argue that the lower ends of the Cotra and Carlsmith estimates are too low and must be precluded, and that Cerebras is underestimating the network size and training data volume which are truly equivalent, in any sense, to the human brain.

So, we might believe that a human-equivalent AGI might take between 1e30 and 1e35 floating-point operations to train, and perhaps 1e20 to 1e21 FLOPS to run in real time. If these numbers are even close to correct, what would this mean for the time and cost of training AGI?

Cerebras leases the CS-2 for about $1.6 million per year, and it can perform about 5e15 FLOPS. Thus, with 100% utilization, a CS-2 performs about 1e17 floating-point operations per dollar. This would suggest that today, an AGI might cost in the range of $10 trillion to $100 quadrillion to train, and in the range of $3 million to $30 million per hour to run once trained, if we knew how to do these things. Put in terms of wafers, this would be about 6e6 to 6e10 wafer-years to train an AGI, and about 2e4 to 2e5 wafers to run one in real time.

The economics look a bit better on NVIDIA's H100, recently released. The SXM version delivers ~4e15 FLOPS in the best possible case (low precision, sparse tensors, and full compute utilization), and in reality might optimistically average more like ~1e15 FLOPS of useful compute. As of May 2023, the lowest cloud price for an H100 SXM is $4.76 per hour on CoreWeave (range: $4.76[52]–$6.33[53] per hour). This works out to 8e17 floating-point operations per dollar, an ~8x improvement over the CS-2. Economies of scale and margin compression might be able to drop CoreWeave's price by a factor of ~2, but not past a floor of ~$1/hr, our minimum estimate of the cost for an H100 SXM alone (GPU cost + energy cost, amortized over 5 years). Because we expect AGI operations to be extremely large scale and very efficient, we'll assume an optimistic figure of $2.50 per hour at scale.

With this figure:

- An AGI needing 1e30 to 1e35 floating-point operations to train would cost about $700 billion to $70 quadrillion to train today.
- An AGI needing 1e20 to 1e21 FLOPS to run would cost about $250,000 to $2.5 million per hour to run today.

(Obviously, we're putting aside the sheer impossibility and market distortions of such gargantuan orders.)

**How much improvement in unit economics would it take to make transformative AGI?**

To bring these kinds of costs down to an all-in cost of about $25 per hour of AGI labor, we will need to gain *at least* five or six orders of magnitude in the unit cost of computation and/or the efficiency of our AGI models relative to the amount of computation performed in the human brain, over the next twenty years. Reduced by five orders of magnitude, the above figures would be:

- AGI training costing $7 million to $700 billion
- AGI inference costing $2.50 to $25 per hour

This is approaching success. The lower estimates are consistent with a $25/hour figure for AGI cost, as an inference cost of, say, ~$8/hour could hypothetically leave room for training and robot bodies to likewise cost ~$8/hour each. And at even at the top end estimate of training cost, which is

---

[52] CoreWeave: [CoreWeave GPU Cloud Pricing]()
[53] NVIDIA: [NVIDIA Launches DGX Cloud, Giving Every Enterprise Instant Access to AI Supercomputer From a Browser]()

approximately 30 Manhattan projects[54] or 3 Apollo moon programs[55], this could allow for AGI training to cost about ~$10 per hour of subsequent inference, if a single trained model can be used for tens of billions of hours of inference, and/or if training is branched so that primary from-scratch training is a rare event. And it is plausible to use one model for billions of hours of inference; even midsized professions involve hundreds of millions of hours of labor per year, so, e.g., an AGI pharmacist model, even if it required de novo training, could amortize.

But this kind of success would require, approximately, five orders of magnitude of progress. Three isn't enough. Four isn't enough. Even five is borderline. Transformative AGI can happen by 2043 only if we gain at least five orders of magnitude in cost of computation per unit AGI.

Along with the cost of computation as such, we must consider the question of how much computation, relative to a human brain, would be required to achieve AGI. It could be less than the human brain actually performs, because we get away with smaller models, or because we get away with less training data. Or, in theory, it could be more.

This means the five orders of magnitude drop in cost can come from two places:

1. A drop in the amount of computation per dollar, which necessarily must include both cost of building high performing computer chips, and the cost of energy used per unit computation
2. The amount of computation used per human equivalent task performance.

We will discuss both dimensions below, and conclude that the most plausible path involves improvement by approximately 1–2 orders of magnitude in the cost of computation, and about 3–4 orders of magnitude in computational requirements.

Or, alternately, a completely different substrate for computing could emerge, which could void these constraints. It could be quantum, optical, or biological, or something else. But whatever it would be, it would have to offer about five orders of magnitude improvement over current silicon computation, or more, and attain global scale, in order to be a suitable substrate for transformative AGI.

---

[54] ~$24B in 2021 dollars: [Manhattan Project - Wikipedia]
[55] ~$260B in 2020 dollars: [How much did the Apollo program cost? | The Planetary Society]

**AGI inference costs (US$) as a function of compute needs and compute costs:**

| | | **computation per dollar** | | | | |
|---|---|---|---|---|---|---|
| **computation needed to run one human-equivalent AGI** | | <=1x of 2023 | 10x of 2023 | 100x of 2023 | 1,000x of 2023 | >=10,000x of 2023 |
| human brain equivalents | FLOPS | <=4E+14 FLOPS/$/hr | 4E+15 FLOPS/$/hr | 4E+16 FLOPS/$/hr | 4E+17 FLOPS/$/hr | >=4E+18 FLOPS/$/hr |
| <=1/100,000 | <=1E+16 | ? | <=3E+00 | <=3E-01 | <=3E-02 | <=3E-01 |
| 1/10,000–1/1,000 | 1E+17 | >=3E+02 | 3E+01 | 3E+00 | 3E-01 | <=3E-02 |
| 1/1,000–1/100 | 1E+18 | >=3E+03 | 3E+02 | 3E+01 | 3E+00 | <=3E-01 |
| 1/100–1/10 | 1E+19 | >=3E+04 | 3E+03 | 3E+02 | 3E+01 | <=3E+00 |
| 1/10–1 | 1E+20 | >=3E+05 | 3E+04 | 3E+03 | 3E+02 | <=3E+01 |
| 1–10 | 1E+21 | >=3E+06 | 3E+05 | 3E+04 | 3E+03 | <=3E+02 |
| 10–100 | 1E+22 | >=3E+07 | 3E+06 | 3E+05 | 3E+04 | <=3E+03 |
| 100–1,000 | 1E+23 | >=3E+08 | 3E+07 | 3E+06 | 3E+05 | <=3E+04 |
| 1,000–10,000 | 1E+24 | >=3E+09 | 3E+08 | 3E+07 | 3E+06 | <=3E+05 |
| >=100,000 | >=1E+25 | >=3E+12 | >=3E+09 | >=3E+08 | >=3E+07 | ? |

| | |
|---|---|
| (green) | below human cost |
| (gray) | above human cost |

*Source data & calculations: Transformative AGI by 2043: what will it take?*

## Our estimates and justification:

**On costs of computation:**

Today's top AI chips compute roughly ~4e14 FLOPS per dollar per hour. How might this improve by 2043?

We estimate today's rate of AI hardware progress by examining NVIDIA's market-leading AI chips. Over the past ~3 years, from the A100 to the H100, NVIDIA delivered a ~2.1x improvement in FLOPS/$.[56] This improvement rate of ~27%/yr is in line with the longer trend of GPU improvement from 2006 to 2021, estimated by Epoch AI to be ~26%/yr (but more realistically ~20%/yr if you exclude the unsustained rapid progress prior to 2008).[57]

---

[56] Relative to the V100, the H100 provides ~9x the FLOPS and launched at around ~1.5x the inflation-adjusted price. The exact dates and price numbers are a little squishy.

[57] Trends in GPU price-performance

If this pace sustains over the next fifteen years (leaving us five years to scale production of the resultant hardware), it suggests a further ~38x improvement in hardware cost-effectiveness. And while there are many reasons to expect progress to decelerate (as we detail later), even taken at face value ~38x is not even 2 of the 5+ orders of magnitude needed to achieve human-level cost-effectiveness.

What's worse, chips are not getting better like they used to.

Looking again at the H100 vs A100, despite ~3 years of technological advancement, transistor density only went up 50%, and the price per transistor actually rose![58] This is a far cry from the original Moore's Law, when densities doubled and transistor prices halved every 1–2 years.

| | **DGX-1 (V100)** | | **DGX A100** | | **DGX H100** |
|---|---|---|---|---|---|
| Launch year | ~2017 | - | ~2020 | - | ~2023 |
| Process node | TSMC N12 | - | TSMC N7 | - | TSMC N4 |
| Launch price (2023 dollars) | $183k | 1.2x | $226k | 1.5x | $349k |
| Watts (max) | 3.5 kW | 1.9x | 6.5 kW | 1.7x | 11 kW |
| GPU transistors | 169 B | 2.6x | 432 B | 1.5x | 640 B |
| Transistors/$ | 0.9M / $ | 2.1x | 1.9M / $ | 1.0x | 1.8M / $ |
| Transistors/W | 48M / W | 1.4x | 67M / W | 0.9x | 57M / W |
| FLOPS (*mixed/FP16) | 1.0E+15 | 5.0x | 5.0E+15 | 3.2x | 1.6E+16 |
| FLOPS/$ | 5.5E+09 | 4.1x | 2.2E+10 | 2.1x | 4.6E+10 |
| FLOPS/W | 2.9E+11 | 2.7x | 7.7E+11 | 1.8x | 1.4E+12 |
| FLOPS/transistor | 5.9E+03 | 2.0x | 1.2E+04 | 2.2x | 2.5E+04 |
| MHz (boost) | 1582 | 0.9x | 1410 | 1.4x | 1980 |
| Memory | 16 GB | 2.5x | 40 GB | 2.0x | 80 GB |

*Note 1: To reflect real-world conditions, we report system prices, not GPU prices. The trends are quite similar with GPU-only prices.*
*Note 2: The price figures are imperfect estimators of cost, as they may be influenced by demand, discounting, competition, different system sales partners, etc.; plus, costs will fall as production lines mature.*
*Note 3: NVIDIA marketing materials list the DGX H100 as having a max power of 10.2 kW in some documents and 11.3 kW in others. We use 11.3 kW.*

[58] In fairness, price is not a great metric here because setting up an A100 manufacturing line has a relatively high fixed cost and low variable cost, which means A100s will be manufactured in any scenario, which implies they will be priced at whatever is needed to make them competitive with H100s. So in general we shouldn't expect large performance-per-price deltas between multiple models. An alternative measurement could be the degree to which A100 prices drop when the H100 is released, but that framing has its own flaws and downside. No method is perfect.

*Note 4: These trends are not unique to NVIDIA. Google's TPU, another top AI chip, has nearly identical performance characteristics: a TPU v4 is essentially half of an A100, with roughly equal energy and transistor efficiency.*

Not only is progress well behind the original Moore's Law, you can see it decelerated from the V100 to A100 to H100.

In fact, this trend of distinctly sub-Moore scaling in computation, has been general since approximately the 28 nm node with respect to cost per gate, and since about the 14 nm node in density, i.e. for several nodes and many years. The change in trend is not ephemeral.

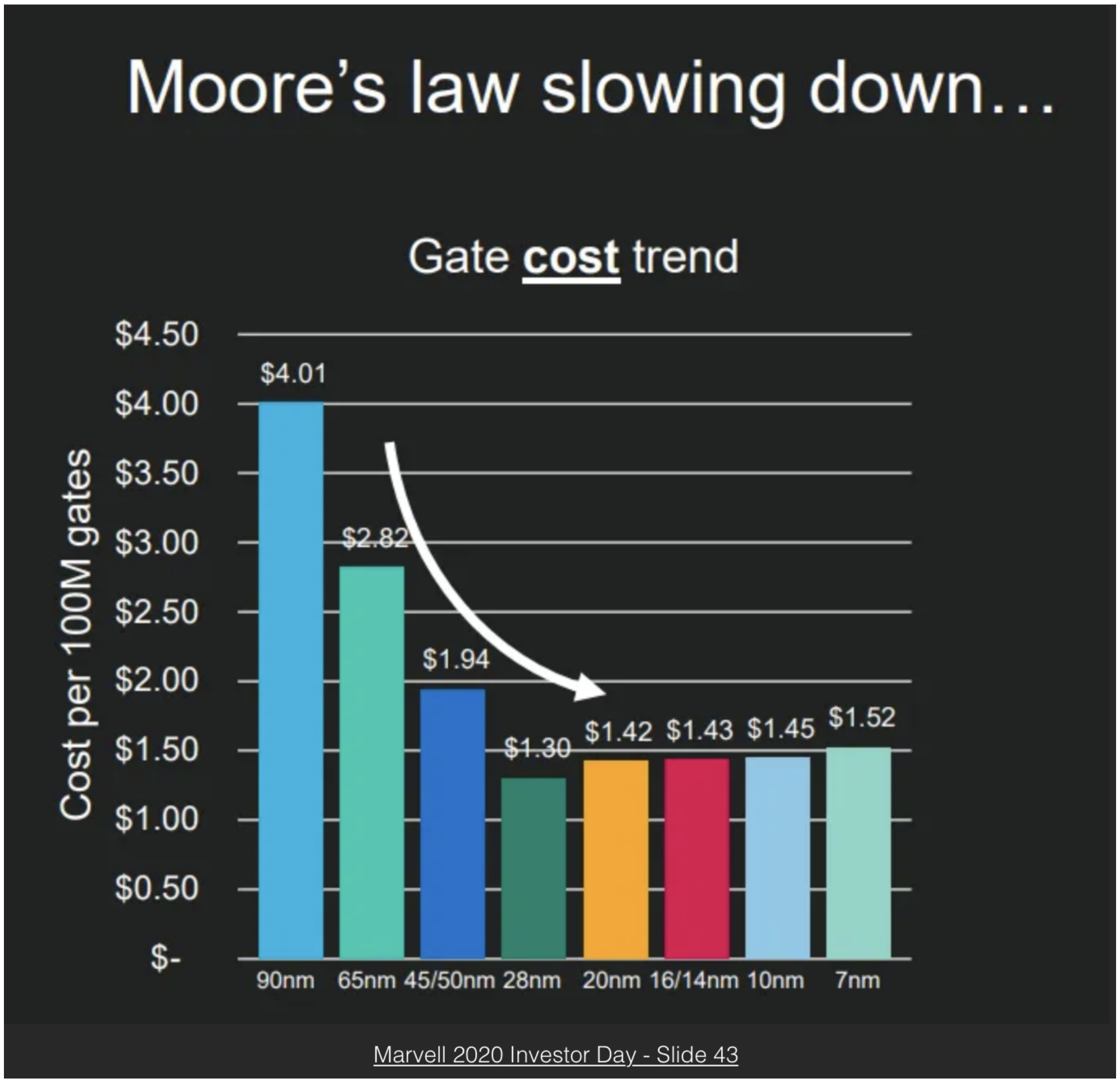

*Taken from Doug O'Laughlin's excellent Fabricated Knowledge Substack article [The Rising Tide of Semiconductor Cost: It Isn't Transitory](#).*

The industry consensus is that the pace of chip progress will continue to be distinctly sub-Moore. Executives of the world's largest semiconductor companies, who have every incentive to be publicly overoptimistic about the future, say plainly that Moore's Law is decelerating:

- NVIDIA's CEO says:
  - "*Moore's law is now 2x every 5 years instead of 10x every 5 years…The semiconductor industry is near the limit…Nobody actually denies it at the physics level. Dennard scaling ended close to 10 years ago. And you could see the curves flattened. Everybody's seen the curves flatten, I'm not the only person….over 10 years now, the disparity between Moore's law is 100 times versus four times, and in 15 years, it's 1,000 times versus eight."* (2022)[59]
  - "*The method of using brute force transistors and the advances of Moore's law has largely ran its course*" (2022)[60]
  - NVIDIA is the world's primary supplier of AI compute, and this is the person who directs its plans and investments.
- AMD's CTO says:
  - *"And you've all heard many times Moore's Law is slowing down. Moore's Law is dead. What does that mean? It's not that there's not going to be exciting new transistor technologies…. we're going to keep improving the transistor technology, but they're more expensive."* (2022)[61]
  - AMD is the #2 GPU maker behind NVIDIA, and this is the person who directs its technical plans and investments.
- ASML's CTO says:
  - *"For years, I've been suspecting that high-NA will be the last NA, and this belief hasn't changed."* (2022)[62]
    - Translation: after ASML releases their next generation lithography tool, the game-changing improvements in wavelength and numerical aperture will likely be exhausted, with only incremental optimizations to follow in the coming years.
  - ASML is the single company that makes the key tool used in chip manufacturing, and this is the person who directs its technical plans.

(In transparency, these quotations are cherry-picked—you can also find quotations by Intel's CEO and others who spin things more optimistically. But we feel the cherry-picking here is fairly mild—the quotations really do reflect broad consensus sentiment, even if the spin varies on the margins.)

Leading semiconductor companies also forecast slowing improvements to their investors and customers:

- ASML forecasts decelerating density & energy efficiency improvements from 2023–2030:[63]

[59] Nvidia CEO Jensen Huang: 'The semiconductor industry is near the limit'
[60] Intel says Moore's Law is still alive and well. Nvidia says it's ended.
[61] AMD: Moore's Law ain't dead, it's just a whole lot more expensive | PC Gamer
[62] Hyper-NA after high-NA? ASML CTO Van den Brink isn't convinced – Bits&Chips
[63] Investor Day 2021 – ASML | Supplying the semiconductor industry

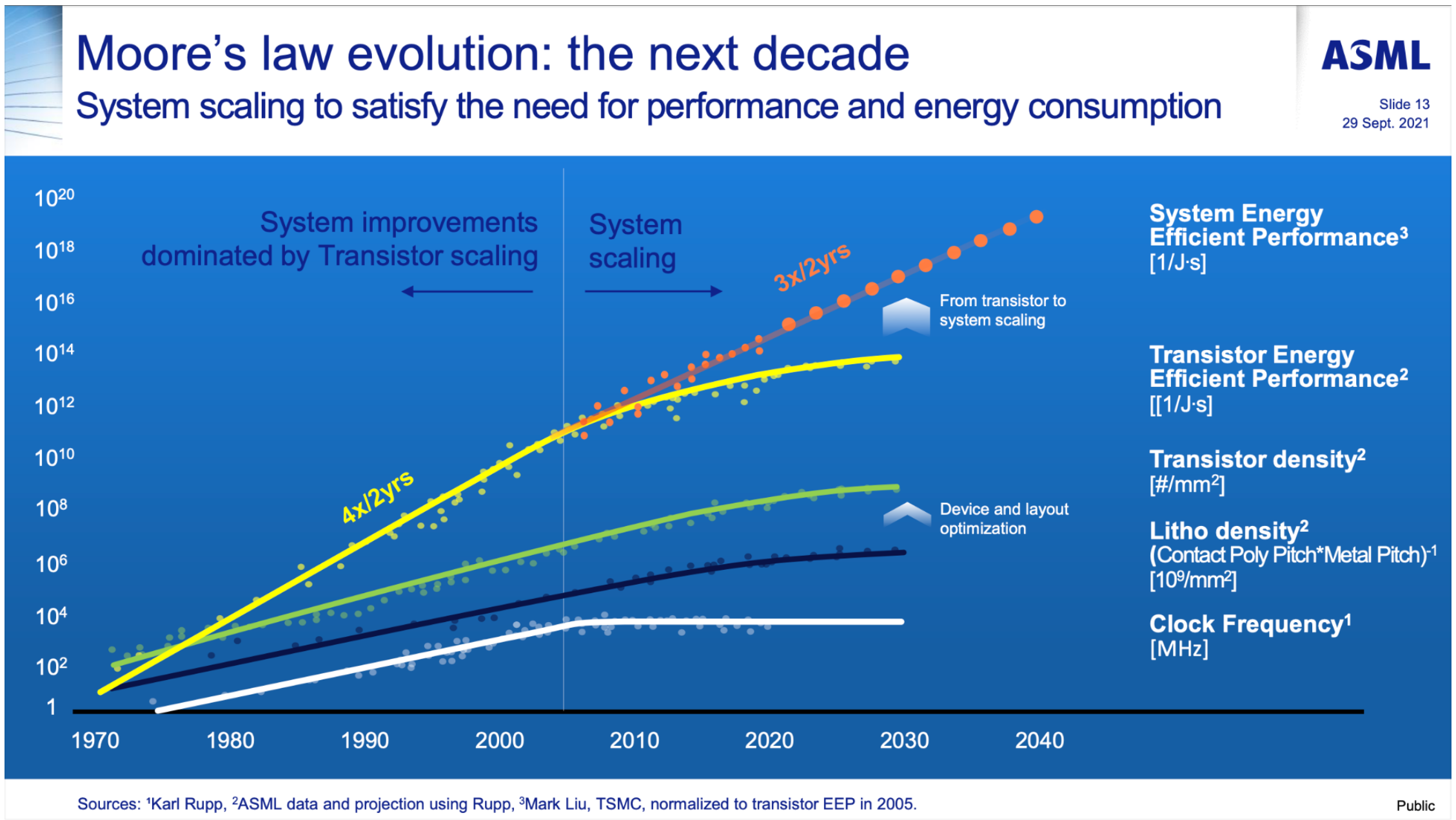

- TSMC, the world's leading chipmaker and the supplier of all major AI chips today, forecasts ***rapidly decelerating density improvements*** of only 30% and >10% for nodes out to 2025:[64,65]

| Node | Theoretical density (MTs/mm2) | Improvement | Apple SOC density (MTs/mm2) | Improvement | % of theoretical achieved |
|---|---|---|---|---|---|
| **N12** | 28.9 | - | 26.4 | - | 91% |
| **N10** | 52.5 | 82% | 49.05 | 86% | 93% |
| **N7** | 91.2 | 74% | 82.86 | 69% | 91% |
| **N7** | 91.2 | 74% | 89.97 | 83% | 99% |
| **N5**[66] | 171.3 | 88% | 134.09 | 62% | 78% |
| **N3E** (2023) | 222.7 | 30% | - | - | - |
| **N2** (2025) | 245.0 | 10% | - | - | - |

  - N3E (due in 2023) will only increase theoretical transistor density by 30%, in contrast to prior node increases of 88%, 74%, and 82%
  - N2 (due in 2025) will only increase theoretical transistor density by ">10%"
  - What's more, this likely overstates the actual improvements, as achieving theoretical densities is getting harder, as evidenced by Apple's A14 (iPhone 12 chip) only reaching 78% of N5's theoretical density, in contrast to prior generations' 99%, 91%, 93%, 91%

[64] TSMC Reveals 2nm Node: 30% More Performance by 2025 | Tom's Hardware
[65] The TRUTH of TSMC 5nm - by SkyJuice - Angstronomics
[66] Note that nodes like N4 are missing because they're considered part of the N5 family. N4 only increases density by ~6% of so. Here's a nice map of their relationships: TSMC Extends Its 5nm Family With A New Enhanced-Performance N4P Node – WikiChip Fuse

- Note too that forecasts are essentially best case scenarios—plans may be unexpectedly delayed by quarters or years (see Intel's 10 nm saga), but plans are never unexpectedly beaten

This lack of progress is not due to lack of effort; these companies are investing more in R&D than ever before. But despite record-setting R&D investments, chip progress is slowing. As each one-time innovation is developed, subsequent innovations are becoming harder and less fruitful. We expect transistor progress to continue but at a sub-Moore pace, as that is the broad consensus of leading semiconductor companies and experts.

Although we are confident that progress in chip *manufacturing* will slow (joining nearly every semiconductor expert), we remain more uncertain about progress in chip *design*.

Looking again at the jump from the A100 to the H100, the ~220% lift in FLOPS can be decomposed into three factors:

- a ~50% lift in transistor density
    - we *strongly* expect this factor to decelerate
    - this is a consensus view; see ASML's and TSMC's forecasts above
- a ~40% lift in clock speed
    - we *very very strongly*[67] expect this factor to decelerate
    - this is a consensus view; see the near-horizontal white curve on ASML's plot above as well as the -10% lift in the preceding generation, which illustrates this isn't a consistent trend
- A ~50% lift in operations per transistor per clock cycle
    - we expect this factor to slow from typical diminishing returns, but with decent uncertainty
    - we're unsure of the consensus view here

Projecting forward fifteen years or so, we see the most room for improvement in chip design that results in more operations per transistor per clock cycle.

Reasons for optimism:

- Today's H100 requires about 80,000 transistors per FP16 operation per cycle. In principle, it feels like there could be room for this seemingly large number to fall, though we are not chip design experts.
- We see the potential for many one-time gains that better match the chip design with the AI workloads, in ASIC-like ways. These high-fixed-cost designs might not be economical today, but could become economical if the AI chip industry scales massively (and if AI-powered tools continue to improve chip design).

Reasons for pessimism:

---

[67] Since the mid 2000s, computer chips have been topping out at 1–5 GHz, because of physics and economics. For more information, look up Dennard Scaling, leakage current, and the 60 mV/decade limit. Two decades of engineering ingenuity have been consistently stymied by this wall, and all signs point to this continuing for decades to come.

- Chip designs are likely fairly optimized, as NVIDIA has been laying out matrix multiplication circuits for decades. Improvements in the past generations may be mostly due to shifting transistor budgets between features, rather than consistent wins in digital circuit design, which may suggest poor prospects for sustained long-term scaling.
- Google's fourth-generation custom-designed machine-learning-specialized chip, the TPU v4, has nearly identical efficiencies to its 7N counterpart, the NVIDIA A100. Convergence of design efficiencies implies they're already near-optimal, or at least chip designers are out of unique clever ideas. The similarity between competitors may also suggest lower variance rates of improvement going forward.
- The limited number of AI chip designers means that a single company like NVIDIA or Google whiffing on a generation of chips could meaningfully slow the world's progress toward AGI. This could be worse relative to a counterfactual world in which a pack of competitors can still forge forward when one is left behind.
- Brian Chau argues that much of the one-time low-hanging fruit for parallelizing operations with CUDA have already been picked, and that we are already on the decelerating side of the S-curve.[68]

Extrapolating forward the rate of 50% every ~3 years, in 15 years chip design may improve efficiencies by 6.5x, not even a single order of magnitude. We expect progress to decelerate, but with wide uncertainty that includes some acceleration.

Combined with improvements from other factors, as well as generally falling manufacturing costs, **we roughly expect FLOPS/$ to improve by 10x–100x, a continuation of today's rate of progress (~38x)**. Low scenarios may result from the continued deceleration of Moore's law, while high scenarios may result from unprecedented investment and new design tools in a world approaching AGI.

**On cost of electricity:**

Lurking behind the challenge of transistor manufacturing costs is a less visible but even greater challenge: energy consumption.

Today, an A100 GPU's cost is dominated by manufacturing; the electricity needed for its operating lifetime is only roughly ~10% of its total cost.[69] However, if manufacturing costs are successfully reduced by 2+ orders of magnitude, as may be needed for cheap AGI, then electrical consumption will become the dominant cost. Electricity costs are much harder to reduce than manufacturing costs because they run up against the laws of physics.

---

68 https://www.fromthenew.world/p/diminishing-returns-in-machine-learning?fbclid=IwAR1PgugeesibMX-B_xG4ItKPxDTkcSV3qgQooFMzjnWID1rcfAa87p_AcEs

69 You can rent an A100 for about $1/hr, commercial electricity costs roughly $0.10/kWh, and an A100 (along with ancillary electronics and cooling costs) draws maybe 1 kW of power. This implies that electricity is roughly ~10% of the cost of operation.

An H100 SXM consumes 700 W of power at peak load, which ends up at maybe ~1.4 kW if you add ancillary electrical costs (~2x),[70] cooling (~2x),[71] and assume a steady state of half max power (~0.5x). At a generous commercial rate of $0.07 / kWh,[72] 1.4 kW costs ~$0.10 / hour. Supposing that a 1e21 FLOPS AGI needs ~1e6 H100s each supplying 1e15 FLOPS, that implies a running cost of ~$100,000 / hr in electricity, four orders of magnitude above the ~$10 / hr needed to undercut rich knowledge workers, and five orders of magnitude above the ~$1 / hr needed to undercut the world's median worker.

Plus, in reality, energy will not be the sole cost of AGI—we'll have chips, data centers, communications networks, robots, robot energy, repayment of investments, etc. all taking slices of the pie as well—so really we'll need more like five to six orders of magnitude of progress, not four to five.

We see the energy challenge as far more difficult than the manufacturing challenge. Although viewed as unlikely by us and the leaders of the semiconductor industry, chip costs can nonetheless plausibly plummet by known routes, with each providing a fraction of the necessary savings; e.g., huge economies of scale as we ramp global AI chip production, perhaps longer tool lifetimes that can be amortized over more chips, new transistor breakthroughs, and new design and manufacturing tricks, all potentially accelerated by huge improvements in AI technology and robotic automation.

However, when it comes to energy efficiency, we are up against the unyielding laws of physics. Even expensive superhuman AGI chip researchers may not be able to make the quick progress needed to make AGI transformatively cheap.

Let's work backward: supposing we succeed, which path to success seems most likely? The five orders of magnitude improvement must come from a combination of four factors:

- Cheaper energy (e.g., fusion)
- Less energy per transistor operation
- Fewer transistors per floating point operation
- Fewer FLOPS needed for AGI (e.g., perhaps 1/10,000 of a human brain)

We review each in turn.

### Cheaper energy

If the cost of energy can be brought down by five orders of magnitude, then 1 GW per AGI may not feel so expensive.

The only Hail Mary that feels remotely capable of dropping energy costs by orders of magnitude within years is rapid deployment of breakthrough nuclear fusion.

---

[70] ~2x is estimated from the DGX H100, which has a maximum power of 11.3 kW, which is twice the max power of its 8 constituent H100 GPUs (8*700 W = 5.6 kW)
[71] Calculating Total Power Requirements for Data Center
[72] Power in the Data Center and its Cost Across the U.S.

However, even if fusion is made to work (very unlikely), made to work cheaply (very unlikely), and made to work in a way that can quickly reach planet scale (extremely unlikely), there's still an issue: 100M AGIs each consuming ~1 GW implies energy consumption of 1e17 W, the same order of magnitude as all solar energy hitting the Earth.[73] Such heat would rapidly overheat the planet and make the planet inhospitable to life.[74,75] Now, perhaps overheating can be overcome by breakthroughs in cheap scalable planetary air conditioning that beams waste heat into space, but that adds a new series of technological breakthroughs and investments needed atop the already unlikely fusion breakthroughs needed, all by 2043.

Lastly, even if fusion energy were literally free (which it won't be), the costs of transmission and distribution and literal data center cabling would still place a non-negligible floor on the retail price paid for electricity far above the near-zero wholesale rate.

Therefore, the path of cheap energy feels nearly impossible (<1%).

### Less energy spent per transistor

The prospect of five orders of magnitude of energy efficiency improvement in transistors within 20 years is well below 1% likelihood, in our view. All field-effect transistors face a fundamental physical limit: the minimum subthreshold slope of 60 mV per decade at 300 K.[76] No new material or fancy geometry or trick of manufacturing can avoid this law of thermodynamics. Beating it requires:

- An entirely new transistor platform that works at cryogenic helium temperatures near 0 K (wildly unlikely, wildly expensive, and wildly limited by helium supply)
- OR extremely precise manufacturing that can tolerate smaller on/off margins (plausible, but not many orders of magnitude worth, and unfortunately it works against the orders of magnitude in cost reduction also needed)
- OR a transistor replacement that's 10,000x more energy efficient, not to mention wildly cheaper (new transistor replacements are barely plausible given the tight 20 year timeline to figure them out and then rapidly switch a $1T industry over, but even today's candidate technologies in research labs are trying their hardest to aim at 2x improvements, not 10,000x improvements[77])

Therefore, the path of greater energy efficiency feels nearly impossible (<1%).

### Fewer transistors per floating point operation

Via improved circuit design, we may be able to squeeze more juice out of the same fruit. This factor is promising. The last two generations of NVIDIA GPUs (V100 -> A100 -> H100) each roughly doubled

[73] Shining brightly | MIT News
[74] Southworth Planetarium
[75] Technically, it may be possible for us to live in a hot future if we (a) make breakthroughs in radiating or otherwise sending high entropy heat into space or (b) make breakthroughs in living in hot environments. But these seem unlikely enough over the next twenty years to discount to <1%.
[76] Subthreshold slope - Wikipedia
[77] WSe2/SnSe2 vdW heterojunction Tunnel FET with subthermionic characteristic and MOSFET co-integrated on same WSe2 flake | npj 2D Materials and Applications

the number of FLOPS per transistor, roughly half by running chips hotter and roughly half by choosing to lay out the transistors differently. If this pace of +50% every 3 years continues unabated, it will deliver a ~7x improvement within 15 years, or roughly one order of magnitude. It's hard for us to imagine this not hitting diminishing returns, but in the longer run there does seem to be plenty of room to encode AI operations in more complex hardware designs. Plus, energy costs and capital costs can be traded off against one another via the clockspeed. *If* progress in chip manufacturing is great enough that energy costs begin to dominate, chips may be designed to operate at lower clockspeeds and greater efficiencies, which could contribute a factor of 2 or more to energy efficiency. Still, supposing this pace can be accelerated to deliver ~100x improvements, that's still less than halfway to five orders of magnitude.

**On the computational intensity of AGI:**

The final possibility for human-cost AGIs is that it turns out they can be made to use orders of magnitude less compute than the human brain, as discussed prior. This is an entirely open question, and feels harder to refute than the other paths. Furthermore, it also nicely diminishes the need for Moore's law-like cost reductions.

Therefore, *conditional on* human-cost AGIs existing by 2043, we are nearly certain (99%+) it will be because we figured out how to make AGIs that require far less compute than the human brain and operate with far worse energy efficiency than the human brain, rather than relying on miracles of planet-boiling fusion generation or radical new transistors as energy efficient as human neurons.

If AGIs end up needing 1,000x less compute than the human brain to do equivalent work, then the semiconductor industry only needs to deliver a 100x improvement in manufacturing cost efficiency and energy efficiency by 2043. Or, alternatively, if AGIs end up needing 10,000x less compute, the semiconductor industry only needs to deliver about 10x improvement in manufacturing and energy efficiency.

However, we reiterate that it's unknown how much compute a human-level or superhuman AGI might need. We know that AGI is *possible* with a human brain's worth of compute, but how much will the first AGIs, as actually achieved by humans, require? There are good arguments both in favor of figures much higher than the amount of computation human brains require, and much lower.

There are some reasons to suspect that AGIs in their early years will be horribly inefficient compared to what's optimal. Nature has aggressively optimized brains for efficient use of computational capacity over time, and human-built systems seldom attain the efficiency of long-evolved systems with ease. For example, we have trouble replicating the feats of muscles, photosynthesis, selectively permeable membrane chemistry, and other core biological processes. Why should our first designed AGIs be more efficient than mature, evolved ones?
But perhaps they should be.

- We already have efficient explicit algorithms and purpose-specific hardware to perform tasks that absorb so much of the human brain, like the image cortices, and explicit mathematics. AGIs can use these efficient algorithms and hardware rather than e.g. using billions of biological neurons to perform arithmetic which can be done with mere thousands of transistors.

- We have the benefit of (at least potentially) not calling parts of our inference and training machinery which may not be needed for specific tasks; unlike a human call center employee, who must keep two image cortices and a smell center and athletic reflexes and a sex drive and so on running in real time while she is on the phone, an AGI call center employee need not even have these modules, let alone run them all the time. Or an AGI-driven robot may not need to keep its locomotion inference running while standing still, or run its language processing routines while doing physical labor alone.
- We are training and inferring our current AI algorithms not on the raw sensory inputs which all humans process, but on compressed, evolved representations like words, categories, and records. And even training on raw inputs may be branched with AGIs in a manner it can't be with humans. If this approach is ultimately sustained through our march to AGI, as many now think it will be, this may significantly reduce the amount of training and inference needed to achieve the kind of comprehension humans achieve.

We really don't know the answer to this question. The amount of compute used by an AGI could be 100k human brains worth, or 10k or 1k or 100 or 10 or 1 or 1/10 or 1/100 or 1/1,000 or 1/10,000. But only a subset of values—say, those 1/1,000 and below—unlock the potential for AGIs being as cheap as humans.

Even stretching our generosity, it feels hard for us to ascribe a probability above 20%. 20% could correspond to a flat probability distribution across the following 10 outcomes, i.e. a log normal distribution over the ratio between AGI and human computational requirements:

| **# of human brain equivalents** | **FLOPS** | **estimated likelihood** |
|---:|---:|---:|
| <=1/100,000 | <=1E+16 | 10% |
| 1/10,000–1/1,000 | 1E+17 | 10% |
| 1/1,000–1/100 | 1E+18 | 10% |
| 1/100–1/10 | 1E+19 | 10% |
| 1/10–1 | 1E+20 | 10% |
| 1–10 | 1E+21 | 10% |
| 10–100 | 1E+22 | 10% |
| 100–1,000 | 1E+23 | 10% |
| 1,000–10,000 | 1E+24 | 10% |
| >=100,000 | >=1E+25 | 10% |

Note that while above we introduced a pessimistic note, arguing that Cotra, Carlsmith, and Cerebras were being too minimal in their estimate of the computational complexity of the human brain, here we are **extremely optimistic** relative to these thinkers. The idea that AGI can be trained with 1/1000 the computation of a human brain's training is three orders of magnitude below the bottom end of Cotra's estimate, and we are giving it a 20% probability. This estimate could be regarded as optimistic, even irresponsibly so, but it's what we think, so we are proceeding with it. We are not reflexively using pessimistic estimates.

**Summary of computation cost of AGI:**

Summarizing, cheap AGI will require cheap chips and cheap energy.

Both an inside view of next-generation technology and an outside view of past trends suggest that 5 orders of magnitude improvement in transistor cost and energy efficiency in just 10–20 years is extremely unlikely, but 1–2 orders is fairly likely, or at least plausible. Although expert and industry consensus is that semiconductor scaling will continue to slow, there's room for design (including the deployment of new custom AI chips) to make additional steps alongside semiconductor scaling. We therefore **adopt a 98% probability** that chips will advance enough within 20 years for the total cost of computation, per floating-point operation, to reduce by at least approximately one order of magnitude, **and a 48% probability** that they will advance enough to reduce it by at least approximately two orders of magnitude.

| improvement from today's H100 SXM | FLOPS/$/hr | estimated likelihood |
|---:|---:|---:|
| <=1x | <=4E+14 | 2% |
| ~10x | 4E+15 | 50% |
| ~100x | 4E+16 | 40% |
| ~1,000x | 4E+17 | 6% |
| >=10,000x | >=4E+18 | 2% |

We adopt the estimate described above, that the probability of AGI hardware requirements, relative to human intelligence, is log uniform between +4 and -4 orders of magnitude, with two "orders" reserved for the upper tail and the lower tail. **This implies that the probability of -3 or less is about 30%, and -4 or less about 20%.**

**Integrating, this yields an estimate of the probability of a total reduction in AGI costs of 5 or more orders of magnitude, of about 16%.**

| | | computation cost efficiency | | | | |
|---|---|---|---|---|---|---|
| **computation needed to run one human-equivalent AGI** | | <=1x of 2023 | 10x of 2023 | 100x of 2023 | 1,000x of 2023 | >=10,000x of 2023 |
| human brain equivalents | likelihood | 2% | 50% | 40% | 6% | 2% |
| <=1/100,000 | 10% | 0.2% | 5.0% | 4.0% | 0.6% | 0.2% |
| 1/10,000–1/1,000 | 10% | 0.2% | 5.0% | 4.0% | 0.6% | 0.2% |
| 1/1,000–1/100 | 10% | 0.2% | 5.0% | 4.0% | 0.6% | 0.2% |
| 1/100–1/10 | 10% | 0.2% | 5.0% | 4.0% | 0.6% | 0.2% |
| 1/10–1 | 10% | 0.2% | 5.0% | 4.0% | 0.6% | 0.2% |
| 1–10 | 10% | 0.2% | 5.0% | 4.0% | 0.6% | 0.2% |
| 10–100 | 10% | 0.2% | 5.0% | 4.0% | 0.6% | 0.2% |
| 100–1,000 | 10% | 0.2% | 5.0% | 4.0% | 0.6% | 0.2% |
| 1,000–10,000 | 10% | 0.2% | 5.0% | 4.0% | 0.6% | 0.2% |
| >=100,000 | 10% | 0.2% | 5.0% | 4.0% | 0.6% | 0.2% |

| | |
|---|---|
| 16% | below human cost |
| 84% | above human cost |

*Source data & calculations: Transformative AGI by 2043: what will it take?*

**Sidenote: a new computing technology superior to field-effect transistors:**

Lastly, we consider the emergence of a new computing technology at global scale in this short timeframe extremely unlikely, and **assign a <1% probability**.

(In this section and others, we use ‘transistor’ as shorthand for ‘field-effect transistor.’)

One of the authors, Ted, did his PhD at UC Berkeley & Stanford where he researched future computing materials and the potential of many post-transistor computation technologies. As he wrote in his dissertation, and now, his expert opinion is that no alternative to the transistor is anywhere close to commercialization.

The continued dominance of the transistor is not a widely debated claim; it's the consensus view of the world's top computer hardware experts.

Here's how top experts expect computer hardware to look in 2047 (2047 is the 75th anniversary of Moore's law, and 4–5 years beyond this essay's forecast horizon of 2043):[78]

- "*Despite the variety, the fundamental operating principle - the field effect that switches transistors on and off - will likely remain the same*" - Suman Datta (IEEE Fellow, professor of ECE at Georgia Tech, and director of nanotech research center ASCENT)
- "*In this industry, it usually takes about 20 years from [demonstrating a concept] to introduction into manufacturing.... It is safe to assume that the transistor or switch architectures of 2047 have already been demonstrated on a lab scale*" - Sri Samavedam (SVP of CMOS technologies at Imec, the European chip R&D center)
    - "*[Tsu-Jae] King Liu agrees*" (IEEE Fellow, co-inventor of the FinFET, dean of the college of engineering at UC Berkeley, Intel board member, and tester of Ted's transistor knowledge in his PhD oral exams)
    - "*[Sayeef] Salahuddin, for one, doesn't think it's been invented yet.*" But "*Transistors will remain the most important computing element*" (IEEE Fellow, professor of EECS at UC Berkeley, co-inventor of negative capacitance FET)
- "*Now we're going to the RibbonFET which is going to last for probably another 20-plus years…so I expect we're going to be somewhere with stacked RibbonFETs*" - Ann B. Kelleher, general manager of technology development at Intel[79]
- "*The focus will be on reducing power and the need for advanced cooling solutions.*" - Richard Schultz (AMD senior fellow)
- "*Advances in quantum computing won't happen fast enough to challenge the transistor by 2047, experts in electron devices say.*"
- "*Transistors will still be very important computing elements for a majority of general-purpose compute applications.... One cannot ignore the efficiencies realized from decades of continuous optimization of transistors.*" - Sri Samavedam
- "*Twenty-five years is a long time, but in the world of semiconductor R&D, it's not that long*"
- "*Experts say that the heart of most devices, the transistor channel region, will still be silicon, or possibly [other semiconductors].*"

The quotations above aren't cherry-picked; they reflect widespread sentiment that the transistor is unlikely to be supplanted in the next couple decades.

The experts above aren't random professors or engineers or PhD holders; they are the world's foremost computer hardware researchers who make their living directing the research of ideas that can improve or supplant silicon transistors.

And despite having every incentive to overhype the future, these experts nonetheless agree clearly and consistently that it's very unlikely for a new technology to displace the silicon transistor in a mere two decades.

---

[78] The Transistor of 2047: Expert Predictions - IEEE Spectrum
[79] Intel's Take on the Next Wave of Moore's Law - IEEE Spectrum

Below, we elaborate on why we and the world's semiconductor experts feel so strongly that the core paradigm of semiconductor transistors will not be beaten at scale in the next 20 years.

### Why the silicon transistor is unlikely to be replaced in the next 20 years

The silicon transistor is arguably the most successful invention of the 1900s.

In 1960, a dollar could buy a single transistor;[80] in 2022, that same dollar (adjusted by the CPI-U to ~$10[81]) could buy ~25 million transistors.[82]

The silicon transistor's tremendous success means dethroning it will be tremendously difficult. Not only will a challenger technology need to surpass it, but the challenger will need to do so by margins large enough to overcome silicon's many incumbent advantages: e.g., huge economies of scale, decades of design optimization, decades of supply chain optimization, billions of sunk capital investments, high gross margins that can shrink in a price war, millions of skilled workers who only know how to make semiconductors, a vast market of chip designers who only know how to design silicon chips, a vast market of device makers who only know how to integrate silicon chips into devices, and a vast marketplace of software that only runs (and is optimized for) silicon chips, and so forth. These barriers are not impossible to overcome, but they are impossible to overcome quickly.

Replacing the silicon transistor early enough to enable transformative AGI by 2043 will require:

- New breakthroughs never achieved by past decades of research
- A global reorientation of global supply chains and manufacturing knowledge and capital investments
- Both of the above happening well within 20 years, such that AGI researchers still have enough remaining years to:
    - Migrate their AGI workloads over to the new technology
    - Invent a path to AGI using the new technology
    - Train AGI using the new technology
    - Scale AGI using the new technology

These are so difficult to do on a timescale of years that the odds seem nearly negligible to us (<1%). Even in a world with proto AGIs (or AI tools) that can improve the rate and cost effectiveness of computer hardware research, the proto AGIs will face the same exact difficulties as us humans.

### The 'tall ladder' of technological development

You can think of technology development like climbing a ladder. Silicon transistors started climbing the ladder in the 1960s, taking about a year or two to climb each rung. It reached each rung by "pushing off" the rung below, taking revenues and learnings from its current technology level and reinvesting them in the next level of technology. Today, silicon has climbed perhaps six stories up, a height that's dizzying to all but the most optimistic challengers. Any challenger hoping to quickly surpass silicon in the next 10 years faces two key hurdles. First, the challenger will need to climb much faster than silicon ever did, given silicon's long headstart. Second, all of the intermediate rungs on the ladder are now gone. Markets no longer exist for transistors that are expensive and crummy. Pushing a new industry through this stage must now be funded solely by investors rather than customers, and without the benefit of continuous market feedback. In other words, trying to take silicon head-on within 10 years means trying to leap up six stories in a single bound—a task that feels wildly difficult.

---

[80] How Much Did Early Transistors Cost? - IEEE Spectrum
[81] CPI Inflation Calculator
[82] As estimated from the NVIDIA H100, which sells for ~$30-40k and contains ~80B transistors. Smaller chips with higher yields and more mature processes will probably get you even cheaper transistors, but we use this example because it's the most relevant to AGI development.

*If* the silicon transistor *is* surpassed, we expect its challenger to approach obliquely rather than head-on. Perhaps it will be a disruptive technology that climbs a separate less important-seeming ladder, before jumping horizontally to compete on the main ladder (e.g., in the same way the ARM microarchitecture grew in mobile and embedded devices, before eventually competing directly against x86 in laptops). Or perhaps it will be a sister technology capable of quickly pulling itself up to the frontier by harnessing progress and investment from the original climb (e.g., in the same way early automobiles took advantage of roads and wheels that had been developed for horse-drawn carriages).

### What would development of a post-silicon technology feel like?

Let's make this a little more concrete and imagine what it might feel like for post-silicon technology to be successfully developed. Suppose the next Einstein works in the field of DNA computing and makes a dozen breakthroughs, simultaneously solving problems of cost, manufacturability, reliability, operating temperature, speed, input-output interfaces, and input-output bandwidth. She shops her technology to venture capitalists, gets some money, buys a factory to start making DNA chips, and in 2027, unveils to the world the first DNA-based computer chip.

In such a world, how long would it take to get this chip into Apple's next iPhone? Many barriers exist:

- The first chip will likely be crummy and expensive (essentially all first products are)
- It will take years to scale up manufacturing to something that can meet Apple's scale
- It will take years to build up trust with Apple to the point they'll bet their company on it (committing to a new chip doesn't mean betting a single product - due to high switching costs, it means betting their entire engine of product development)
- It will take years to figure how to write new compilers and libraries and software ports (and ported software will likely be far less efficient, as it was designed and optimized for a different technology)
- It will take years to understand how to properly design systems and software to take advantage of the DNA computer chip's tradeoffs
- It will require winning a price war against TSMC, who has a cost structure with low marginal costs (~59% gross margins in 2022Q2[83]), and whose suppliers also have room to drop prices if needed
- It will require selling into iPhone models that haven't yet been designed and contracted, which by itself pushes deployment out at least a couple years

---

[83] TSMC Reports Second Quarter EPS of NT$9.14

Just look at how difficult it's been for Intel to break into dedicated GPUs with ARC, or for Microsoft to make Windows run on ARM, or for AMD to make GPUFORT into something that can rival NVIDIA's CUDA. These are difficult projects taking perhaps thousands of workers and perhaps billions in investment - and yet these challenges of adapting hardware and software to new instruction sets and translation layers pale in comparison to the revolution in computer technology that would be needed to reorient the computer industry around a brand new hardware platform like a DNA computer chip.

The obstacles above are so severe that we find it difficult to imagine silicon transistors being widely replaced in the next 15 years, though we remain very open-minded to longer timescales.

*If* silicon transistors *are* replaced on such a short time scale, we expect it not to happen via upending the entire semiconductor industry (e.g., needing to switch over manufacturing & designs for telecom, automotive, PCs, etc.). Rather, we expect to see a breakthrough that leads to new chips narrowly specialized to machine learning workloads and given interfaces that allow them to operate in existing data centers. This pathway feels more likely because:

- Investment in new chip designs is limited to just a single line of chips
- Investment in new chip instruction sets is limited to just a single line of chips
- Extra efficiencies can come from manufacturing only a single design, and designing for only a single class of application

However, we still see this path as extremely unlikely, for the following reasons:

- The base rate of new computing technology breakthroughs is very, very low (0 technologies in the past 60 years have successfully dethroned silicon transistors)
- The base rate of new computing technology breakthroughs that can move from prototype to production to mass production in <10 years is very, very low (depending on how you count, you could plausibly point to a few key advancements in magnetic disk heads such as tunnel magnetoresistance heads, which took ~5 years between breakthrough and mass production; but swapping in Fe/MgO/Fe for Fr/Cr/Fe is a far simpler task than reinventing the transistor)
- Conditional on this technology not being good enough to succeed in most applications, it won't benefit from economies of scale needed to drive down costs (not just manufacturing economies of scale, but economies of scale in writing and improving and perfecting all of the software that will need to run atop this new hardware technology, and in developing the human capital needed to write this software)

### Unforeseen hardware improvements are extremely unlikely

Semiconductor improvements take time. They take time because the manufacturing process is complex and empirical and making a chip from start to finish takes months.[84] Advancing a prototype to mass production often takes 10+ years and legions of highly skilled researchers. Because of this, we consider it extremely unlikely (<1%) that an idea will arise in the next 10 years that is (a) unforeseen (b) improves the cost/energy efficiency of the transistor by >10x and (c) is manufactured in high volume. Any idea to improve chipmaking over the next 20 years is very likely already prototyped in R&D labs today.

[84] Chipmakers Are Ramping Up Production to Address Semiconductor Shortage. Here's Why that Takes Time

Examples transistor improvements from history (not cherry-picked)[85]:

| Transistor improvement | Year of first[86] prototypes | Year of high-volume manufacturing | Development time |
|---|---|---|---|
| Immersion lithography | 1989[87]<br>(but patented in 1984[88]) | 2007[89] | ~18 years |
| EUV lithography | Idea explored in 80s, and companies began investing significantly in early 90s<br><br>ASML's first prototype was done in 2006 | 2020[90] | ~25 years |
| Strained silicon | 1991/1992[91] | 2003[92] | ~12 years |
| Hafnium oxide gate dielectrics | 1967[93] | 2007[94] | ~40 years |
| Tri-gate transistor | 1987[95]<br>(related FinFET was prototyped in 1998[96]) | 2011 | ~24 years |
| Gate-all-around / nanosheet / ribbon / multi-bridge channel field effect transistor | 1988[97] | 2022 by Samsung; Intel in 2024; TSMC in 2025[98] | ~34 years |

[85] The author picked the first 6 major transistor improvements from the past 20 years that came to mind.
[86] Consider these upper bounds, as they come from the author's Google scholar searches, which were admittedly not exhaustive.
[87] https://doi.org/10.1016/0167-9317(89)90008-7
[88] A. Takanashi, T. Harada, M. Akeyama, Y. Kondo, T. Karosaki, S. Kuniyoshi, S. Hosaka, and Y. Kawamura, U. S. Patent No. 4,480,910 (1984)
[89] A highly scaled, high performance 45 nm bulk logic CMOS technology with 0.242 μm2 SRAM cell | IEEE Conference Publication | IEEE Xplore
[90] Our history | ASML - Supplying the semiconductor industry
[91] Electron mobilities and high-field drift velocities in strained silicon on silicon-germanium substrates | IEEE Journals & Magazine | IEEE Xplore
[92] High-k, strained Si leaving the lab - EE Times
[93] MOS transistors with anodically formed metal oxides as gate insulators | IEEE Journals & Magazine
[94] https://en.wikipedia.org/wiki/45_nm_process
[95] https://doi.org/10.1109/IEDM.1987.191536
[96] DARPA put out a call for 25 nm technologies in 1996, a Berkeley group won funding and started work in 1997, and they published results in 1998: FinFET History, Fundamentals and Future
[97] https://doi.org/10.1109/IEDM.1988.32796
[98] TSMC Says it Will Move to Nanosheet Transistors at 2nm | Extremetech

**No candidate technology is close**

If we expect that a successor technology to silicon transistors would probably exist today in order to have the potential to be mature and globally scaled by 2043, we can confine our evaluation to technologies that already exist and are known today.

The immense difficulty of surpassing silicon has not stopped researchers from carefully investigating alternative technologies, such as:

- Quantum computing (which has dozens of potential physical implementations; e.g., superconducting qubits, trapped ions / optical lattices, diamond vacancies, etc.)
- Optical computing (potentially high bandwidth and power efficient)
- Spintronics or magnetic computing (potentially very power efficient)
- DNA computing (potentially very space efficient)
- Graphene or carbon nanotube transistors (potentially higher mobility & transconductance)
- Memristors
- Tunnel field effect transistors (capable of beating the 60 mV/dec limit)
- Negative capacitance transistors (capable of beating the 60 mV/dec limit)
- High-electron-mobility transistors (e.g., GaAS/AlGaAs)

Analyzing the potential of each of these technologies in detail could fill up a shelf of textbooks, and would require a fair bit of trust on your part that we are representing them fairly.

In lieu of such a review, we'll highlight the single, least ambiguous evidence we see: the current absence of investment by the semiconductor industry and outside investors.

Look at quantum computing for instance:

- "Over the past three years, the rate of publicly announced quantum computing start-up founding has slowed" - McKinsey (2022)[99]
- D-Wave's stock has fallen 95% since going public (market cap of only ~$60M as of Apr 2023)
- Rigetti Computing's stock has fallen 95% since going public (market cap of only ~$60M as of Apr 2023)
- IonQ's stock has fallen 50% since going public (market cap of only ~$1.3B as of Apr 2023)

There appears to be near-zero confidence by investors that quantum computing will yield AGI-scale trillion-dollar returns in the next 20 years.
Other technologies like spintronics, DNA computing, memristor computing, etc. have even fewer signs of commercial traction than quantum computing.

—

We ascribe the following cascading conditional probabilities to the emergence of a new computing technology at global scale in this short timeframe as follows:

---

[99] Quantum computing funding remains strong, but talent gap raises concern

- 20% chance that a mass manufacturable device is invented with superior $/floating-point operation and J/floating-point operation than the semiconductor transistor by ~2035 (our personal favorite is the TFET, which can (a) beat the physical 60 mV/dec limit inherent to semiconductor transistors, and (b) shares the same manufacturing process)
- 5% conditional chance that it's >10x better, which is minimum conceivable margin for rapidly trying to supplant the enormous incumbent advantages of traditional chip manufacturing (this is a high bar, even conditional on the prior step being achieved)
- 30% conditional chance that it can be scaled up within 5–15 years, given the potentially massive investments in physical and human capital up and down the supply chain (note too that the distributed nature of the industry means that not all segments of the supply chain will simultaneously invest gung-ho before seeing initial markets proven out, which delays real-world scaling)

In total, this leads us to ballpark a **~0.3%** chance of a superior computing alternative being developed and scaled and used as the substrate for AGI research and deployment by 2043.

**Summary**

**In composite, then, we assign a probability of ~16% that cheap AGI can be trained.**

- ~16% that, conditional on developing AGI, we can design a version that gains five orders of magnitude in cost efficiency relative to today's computation costs and computational requirements of the human brain, from a combination of better transistors and better computational efficiency
- <1% that a new transistor is invented that is both five-ish orders of magnitude cheaper and five-ish orders of magnitude more energy efficient

# Element 4: We invent and scale cheap and capable robot bodies, or robot bodies are not necessary for AGI to be transformative

## What it is and why it's necessary:

Imagine a world where human-level AGI is available and costs $25/hour, but is exclusively in the cloud and incapable of instantiating robot bodies that can maneuver in unprepared environments and manipulate objects the way humans can. In this world, it's quite possible that the transformative impact of AI would be pretty limited. AGIs would replace some portion of human jobs devoted to driving heavy machinery, but other than that it would mostly just be the small minority of existing jobs that are entirely electronic in nature and pay more than $25/hour. Given that about 90% of human workers make less than $25/hour, the fraction of human work that is both more than $25/hour and purely electronic is even smaller (despite the fact that we, and perhaps most readers of this essay, will fall into this category). And we wouldn't expect to see AGI pals making everyone's life easier by turning phones and computers into verbal-interface Djinn-like entities, or other AGI applications that are not already organized as existing jobs. Personal assistants can already be hired to do these tasks for $25/hour or less, including the ability to do physical tasks, and most people can't afford to hire them or otherwise don't.

We think robot bodies will be needed for transformative AGI. If you disagree, then the question is moot, of course. The discussion below is predicated on this assumption, but please feel free to assign a probability of 100% and skip it if you do not agree.

So, arguably, in order for AGI to really transform human life, it will need robot bodies that are capable, cheap and scalable.

- Capable enough to do many/most/all blue collar jobs
- Cheap enough to be consistent with a total package wherein a robot body plus the back-end AGI driving it cost less than $25/hour together
- Scalable enough to make a difference globally. For example, if AGIs did jobs equating to 10% of human workers (500 million people working 2000 hours per year) and 80% of them required robots, we might need robots to perform 800 billion hours of work per year.

Now, right now there are lots of reasons to believe that capable robots will emerge by 2043. The Boston Dynamics Atlas is already remarkably capable, and they are by no means the only group seriously working on this problem with a real chance of success.

But cost and scalability are more serious concerns.

To know how much robots would have to cost, and how many we would have to make, for them to contribute to AGI, we need to know how long they would last. If capable robot bodies emerge, how long would they last on average? Probably at most low tens of thousands of hours of labor service, like other industrial machines. Long haul trucks last about 12,000 hours[100]. CNC mills last about 30,000 hours before all their components other than structural iron need to be replaced[101]. Tractors last between 4,000 and 10,000 hours[102]. Airliners like the Airbus 320 are often rated for no more than 60,000 hours, with average lifetimes shorter than this[103]. And, of course, robot bodies for performing versatile human labor tasks will not be nearly as mature in 2043 as trucks, CNC machines, tractors and airplanes are today; they may last less long.

[100] What is the Average Lifespan of a Long Haul Truck?

So, if a robot body lasts for (e.g.) 20,000 hours, and if we suppose nominally that the $25/hour cost of AGI labor could be composed of $10/hour for the body and $15/hour for the computation, then a robot body costing about $200,000 would be needed in order to achieve these kinds of goals. So, how much might such robots cost? Boston Dynamics' Spot robot sells for $75,000[104], and is about one third the size and number of degrees of freedom of their humanoid Atlas model[105], so this might imply that if Atlas were for sale it might cost about $200,000. However, Spot is a boutique product, with only hundreds sold, and the $75,000 price may not be sustainable[106]. In addition, Atlas currently has only 28 degrees of freedom, with very limited gripping and holding ability, while Tesla's Optimus group has settled on a hand design which raises the total for the whole robot to 50 degrees of freedom[107]. So, perhaps an Atlas style robot with complex hands, sold profitably, would cost more than the current Atlas would. But there is significant disagreement with this general genre of cost number; Tesla has said that it thinks its Optimus robot could sell for as little as $20,000, hands and all. So, cost figures that are compatible with transformative AGI are somewhere within the range of expert opinion.

How about scale? If robot bodies last 20,000 hours, to perform 800 billion hours of labor per year with them, we might need to manufacture about 40 million such robots per year to power an AGI future. Boston Dynamics robots are currently hand-assembled, and they have not spoken about large scale. Tesla has said that its Optimus robot would be produced at a scale of "probably ultimately millions of units[108]," but also that it would produce "thousands, maybe millions" of robots[109]. Both groups are pretty far from even discussing the kind of scale that would be needed to make AGI economically significant in blue collar occupations globally. And scale does take time; Tesla took 13 years from launching the Tesla Roadster, at approximately the scale Boston Dynamics is currently achieving with Spot, to manufacturing a million cars in one year.

## Our estimate and justification:

We think that given the progress in robotics over the last twenty years, and Tesla's fairly responsible price and scale estimates coming in not too far from the needed figures, it's actually fairly likely that robot bodies will be available for use by AGI intelligences at scale in 2043, if the global economy and technology in general keep advancing. In addition, many types of work may be able to use non-humanoid robot bodies, including wheeled models like the Boston Dynamics Grip, which will be simpler to manufacture and scale. We give the suitable availability of robot bodies in 2043 a 60% chance of happening.

---

[101] Rebuilding a CNC machine takes skill and money. Is it worth it? | Cutting Tool Engineering
[102] Understanding Used Tractors' Life Expectancy
[103] Airbus plans A320 life-extension - Asian Aviation
[104] Inside Boston Dynamics' plan to commercialization
[105] Atlas™ | Boston Dynamics
[106] Boston Dynamics sells to Hyundai Motor Group in $1.1 billion deal | Ars Technica
[107] Tesla Reveals Optimus, a Walking Humanoid Robot You Could Buy in 2027 - CNET
[108] Elon Musk said Tesla's AI robot Optimus will eventually 'cost less than a car' and could lead to a 'future with no poverty'
[109] Elon Musk Shows Off Working Optimus 'Tesla Bot' Prototype | PCMag

If you feel that robot bodies will not be necessary, you will of course assign a 100% probability.

# Element 5: We quickly scale up semiconductor manufacturing and electrical generation

## What it is and why it's necessary:

One key question for forecasting transformative AGI timelines is how long it will take to manufacture the mental substrates (e.g., patterned silicon wafers) to support (a) training AGIs to do any task at or above a human-level and (b) a workforce of trained AGIs whose output approaches humanity's (or something equivalently transformative).

**A transformative workforce of AGIs will need unprecedented compute**

Right now the world contains about 5 billion working humans, implying that there are about 10 trillion commercial human work hours per year. If you wanted to do even 10% of global commercial work using AI, this would be about 1 trillion hours per year, or, with 50% utilization, about 250 million devices.

Put another way, with global GDP on a trajectory to hit perhaps $200 trillion in 2042, to make 10% of global GDP at $25/hour and 50% utilization would be about 800 billion hours per year, or about 400 million devices working during working hours.

Of course, future AIs might not be configured into human-equivalent packages, so consider 'device' shorthand for the capital equipment share required to generate $25/hr of value. In reality, there might be a single gigantic chip doing the work of 1,000 humans. Or 1,000 cheap chips networked together to do the work of a single human. To simplify the language, we'll use 'device' as shorthand for human-equivalent share.

Above, we estimated that de novo training of AGI, might require about 1e27 to 1e32 floating-point operations, and running one human-equivalent AGI in real time might require about 1e16 to 1e18 FLOPS, in the optimistic scenario where we make AGIs train and infer with about 1/1000 the computation performed by the human brain, which, for reasons we discuss above, is the likeliest scenario for transformative AGI in 2043.

How many floating point operations per year might this be? Above, we discussed a world wherein AIs perform about one trillion hours of labor per year, equivalent to about 500 million human workers, and wherein the equivalent of one de-novo AGI training run (probably divided into branched partial runs from saved waypoints) could support on the order of one billion hours of AI labor. This might imply that we would require:

- About 4e31 to 4e33 floating-point operations per year for inference (1e16 to 1e18 FLOPS times one trillion hours of labor per year)
- About 3e30 to 3e35 floating-point operations per year for training (about 1000 de-novo training events equivalent, to support one trillion hours of inference)

(If you object to necessarily oversimplified terms like "AGI" or "human-equivalent" or others that seem to unnecessarily constrain the diversity of unknown futures, please know that we recognize the oversimplification as well, and use it reluctantly to smooth the articulation of what we hope is a more general argument. Feel to mentally substitute or amend these terms as you wish, and try to see past them. We recognize the fuzziness here.)

**In this world, what would a workforce of AGIs require?**

Scaling up AI technology may be bottlenecked by two key inputs—semiconductor fabs and power plants—both of which are capital intensive and take years of planning and construction.

Today, NVIDIA's recently released[110] H100 SXM can perform a *maximum* of 4e15 FLOPS at an energy cost of 700 W, assuming:

- Work can be performed by highly parallel tensor core core operations (unclear for unbuilt AGIs)
- Full utilization (e.g., no bottlenecking / idle time due to memory capacity or IO; for reference, PaLM's FLOPS utilization on Google's custom TPUs was 58%[111])
- FP8 precision (⅛ with FP64)
- Sparse tensors (½ without sparsity)
- Running hot to maximize compute per second, not compute per kWh[112]
- Full uptime

However, in reality:

- Its useful output is more like 1e15 FLOPS (or less), assuming no sparsity and 50% utilization
- Its total energy cost is more like 2 kW, given 1 kW of ancillary power[113] and +100% for data center cooling, and minus a chunk for 50% utilization

Conditional on AGI workers running at <$25/hr, which we estimated would take ~5 orders of magnitude improvement in cost-efficiency and energy efficiency, we will conditionally assume:

- Hardware has gotten 100x as cost-effective as today's H100s
- Hardware has gotten 100x as energy efficient as today's H100s

Let's call this hypothetical hardware an "X100," and assume it has useful output of about 1e17 FLOPS at the same 700 W energy consumption.

We translate the compute needs estimated earlier into chip wafers and GW-scale power plants as follows:

---

[110] NVIDIA: H100 Hopper Accelerator Now in Full Production, DGX Shipping In Q1'23
[111] Estimating 🌴PaLM's training cost
[112] DGX-A100 Face to Face DGX-2—Performance, Power and Thermal Behavior Evaluation; based on the Mandelbrot benchmark, you can save ~36% energy in exchange for a +38% runtime
[113] As an example, the max power of the DGX H100 system (11.3 kW) is about twice the max power of its 8 constituent GPUs (8*700 W = 5.6 kW)

<table>
<tr><th></th><th></th><th colspan="2">AGI training</th><th colspan="2">AGI inference for 10% of labor</th></tr>
<tr><th></th><th></th><th>Lower estimate:<br>3e30 operations</th><th>Upper estimate:<br>3e35 operations</th><th>Lower estimate:<br>1e16 FLOPS/AI</th><th>Upper estimate:<br>1e18 FLOPS/AI</th></tr>
<tr><td></td><td>FLOPS needed</td><td>1E+23</td><td>1E+28</td><td>3E+24</td><td>3E+26</td></tr>
<tr><td rowspan="3">2023 HW needs</td><td>NVIDIA H100 SXMs needed, assuming:<br>- Low-precision tensor operations<br>- 50% utilization efficiency</td><td>1E+08</td><td>5E+12</td><td>2E+09</td><td>2E+11</td></tr>
<tr><td>Wafers needed, assuming:<br>- 50 GPUs per wafer</td><td>2E+06</td><td>1E+11</td><td>3E+07</td><td>3E+09</td></tr>
<tr><td>GW power plants needed, assuming:<br>- 2 kW per H100</td><td>2E+02</td><td>1E+07</td><td>3E+03</td><td>3E+05</td></tr>
<tr><td rowspan="3">2038 HW needs</td><td>"X100"s needed, assuming:<br>-100x the compute of H100<br>-100x the energy efficiency of H100</td><td>1E+06</td><td>5E+10</td><td>2E+07</td><td>2E+09</td></tr>
<tr><td>Wafers needed, assuming:<br>- 50 GPUs per wafer</td><td>2E+04</td><td>1E+09</td><td>3E+05</td><td>3E+07</td></tr>
<tr><td>GW power plants needed, assuming:<br>- 2 kW per X100</td><td>2E+00</td><td>1E+05</td><td>3E+01</td><td>3E+03</td></tr>
</table>

*Source data: Transformative AGI by 2043: what will it take?*

*Note: Inference costs are conditional on prior step that assumes ~$25 / hr inference costs, and are **much** more optimistic than our unconditional forecast.*

Note that this calculation already bakes in quite a bit of optimism from our earlier conditional steps in the sense that they assume in the next 20 years:

- We develop algorithms that lead to AGI
- Our AGI algorithms are far more compute efficient than the human brain
- Those algorithms can be implemented in highly parallelizable low-precision sparse tensor multiplications
- Implementing human-level intelligence in high parallelizable low-precision tensor multiplications has no hit on compute efficiency
- Compute hardware becomes 100x more cost-efficient and 100x more energy efficient

Looking at training needs in 2038:

- We may need **20k to 1B wafers**
- We may need **2 to 10k GW-scale power plants**

Looking at inference needs for enough AGI “workers” to do 10% of global labor in 2038:

- We may need **300k to 30M wafers**
- We may need **30 to 3k GW-scale power plants**

The upper ends of these estimates are unlikely to be achieved by 2043, suggesting a less than 100% chance of success in this step toward transformative AGI by 2043. Below, we try to pin down how much less than 100% it might be.

**What it will take to scale chip manufacturing**

As of early 2023, NVIDIA produces maybe ~1M A100 *equivalents* a year (~$16B annualized data center revenue / ~$16k price) or very roughly ~20k wafers worth per year.

We have anticipated that between training and inference, we may need between 300k and 1B wafers of X100 chips in the late 2030s. If manufactured over a 5-year period, that would require a production rate of roughly 70k–200M wafers per year.

Supposing that we need numbers in the middle to high part of this range, how quickly can the world scale production?

**The time it takes to buy compute depends on how much you need**

For individuals like you or us, buying compute is easy. You can sign up for an account on Google Cloud, and within minutes run an A100 GPU for ~$1/hr.[114]

But a lead time of minutes is a privilege that vanishes as soon as your order exceeds inventories. The larger your order, the longer it will take complex semiconductor supply chains to fulfill. At large enough scales, demand for compute will require construction of extra fabrication plants, which take 3–5 years to build (5 years is already 25% of the time until 2043!). At the gargantuan scales that might be needed for transformative AGI, expanding the industry by multiples could take more than a decade.

The more you order, the longer it takes

| Quantity of GPUs | Lead time |
|---|---|
| ~10 | **Seconds** for the datacenter to allocate installed GPUs |
| ~1,000 | **Weeks to months** for Google to order & install new GPUs |
| ~100,000[115] | **Months to years** for NVIDIA to order more manufacturing capacity from TSMC |

[114] GPU pricing | Compute Engine: Virtual Machines (VMs) | Google Cloud
[115] NVIDIA data center revenue is ~$4B / quarter and an A100 GPU costs ~$10-20k, placing a rough ceiling of ~2-400,000 A100s manufactured per quarter (~100k/mo). Adding a quarter's worth of production would likely involve a serious negotiation with TSMC and wait times for fab capacity to be allocated and switched over.

| | |
|---|---|
| ~10,000,000[116] | **3-5 years** for TSMC to build new fabs (and for toolmakers to build tools for those fabs) |
| ~1,000,000,000 | **~Years** for the world to scale up education of engineers[117] and technicians and fab construction firms, neon extraction facilities, power plants, semiconductor fabs, tools inside the fabs, tools to build the tools, tools to build the tools to build the tools, etc. Much of this work can be parallelized, but chains of tools to build tools to build tools cannot be. |

*Illustrative numbers*

### Semiconductor shortages illustrate how capacity ramps take years

The clearest proof of these time scales is the observed historical time it takes the semiconductor industry to resolve shortages.

Consider the 2020–2023 chip shortage,[118] which hit automakers especially hard. When COVID19 struck in 2020, car sales plunged by 30%–90%,[119] inducing many automakers to cancel future chip orders.[120] Then when demand rebounded unexpectedly quickly, automakers found that there were not enough chips to buy, nor enough available manufacturing capacity to make them.

This shortage of chips has been severe. Every major automaker (including Toyota,[121] Volkswagen,[122] Daimler,[123] Ford,[124] Honda,[125] and GM[126]) has been forced to idle manufacturing plants. In 2021 alone, the chip shortage reduced auto production by ~11 million vehicles,[127] costing an estimated $210B in

[116] An ASML EUV lithography tool can roughly process ~150 wafers per minute. An advanced logic wafer may take ~15 EUV lithography steps. A large fab with 10 EUV tools will therefore produce ~100 logic wafers per hour. With perhaps 5,000 hours of production in a year, a large fab might therefore produce 500,000 logic wafers. Each wafer can hold 64 GPUs. Assuming a yield of 75%, that means a large fab might have the capacity to produce ~24M A100 GPUs a year. Therefore, orders on the order of ~10M chips will likely require a new fab, either for the chips themselves or the other chips they outbid and displace.
[117] ASML Faces a Much Bigger Problem Than a Chip Glut
[118] 2020–present global chip shortage - Wikipedia
[119] UK car sales plunged 97 percent in April, making Tesla number one | Ars Technica
[120] Semiconductor shortage that has hobbled manufacturing worldwide is getting worse
[121] Volkswagen and Toyota face production cuts due to chip shortage | Automotive industry | The Guardian
[122] Volkswagen and Toyota face production cuts due to chip shortage | Automotive industry | The Guardian
[123] Daimler AG to idle Mercedes-Benz plant in Hungary for a month | Nasdaq
[124] A silicon chip shortage is causing automakers to idle their factories | Ars Technica
[125] Honda to idle 5 North America plants for a week amid chip shortage - Nikkei Asia
[126] G.M. to Idle North American Plants Because of Chip Shortage - The New York Times
[127] What Happened With the Semiconductor Chip Shortage—and How and When the Auto Industry Will Emerge

auto revenue(!).[128] In response, the White House convened 3 summits[129] and signed the CHIPS and Science Act of 2022.[130]

The current semiconductor shortage is receding, but may echo into 2024, as building extra semiconductor manufacturing tools is taking years.[131] Even in late 2022, lead times of various tools were as high as 24-30 months,[132,133] and lead times of some chips were still over 100 weeks.[134] These lead times would be even longer if demand was higher.

In 2021, TSMC stated it would invest $100B in new fabs over the next three years, and Samsung said it would invest $116B over the next decade.[135] Note that the timescales of these investment decisions are already 15% and 50% of the time to 2043, respectively.

Although making chips for cars in 2022 is admittedly a different situation than making chips for AGI in the 2030's, it helps illustrate the timescales of semiconductor supply chains. Literally hundreds of billions of dollars of automotive revenue (tens of billions of profit[136]) were sitting as a prize to anyone who could make some extra chips for automakers. And this is a relatively "easy" task - automotive is only a small ~10% slice of the semiconductor market,[137] and doesn't even need the leading edge technology nodes, where production is even more bottlenecked. Yet even with this massive prize of billions of dollars, the industry couldn't match this demand in under 3 years. It's not for lack of skill or effort—these companies have hundreds of thousands of employees, a large fraction with advanced degrees—but simply because the tools and buildings needed to fabricate modern semiconductors are so fantastically complex that they take years of physical effort to construct and calibrate and test and ship.

---

128 The Semiconductor Shortage Explained: The Auto Industry's Big Challenge (2023) | Automoblog
129 Readout of Biden Administration Convening to Discuss and Address Semiconductor Supply Chain | The White House
130 FACT SHEET: CHIPS and Science Act Will Lower Costs, Create Jobs, Strengthen Supply Chains, and Counter China | The White House
131 Intel CEO now expects chip shortage to last into 2024
132 Equipment lead times stretching out to 18 months and beyond
133 https://www.thelec.net/news/articleView.html?idxno=3179
134 Market Update - A2 Global Electronics + Solutions
135 https://www.bloomberg.com/news/articles/2021-04-01/tsmc-to-invest-100-billion-over-three-years-to-grow-capacity?leadSource=uverify%20wall
136 Auto & Truck Manufacturers Industry Profitability
137 Semiconductor shortage: How the automotive industry can succeed | McKinsey

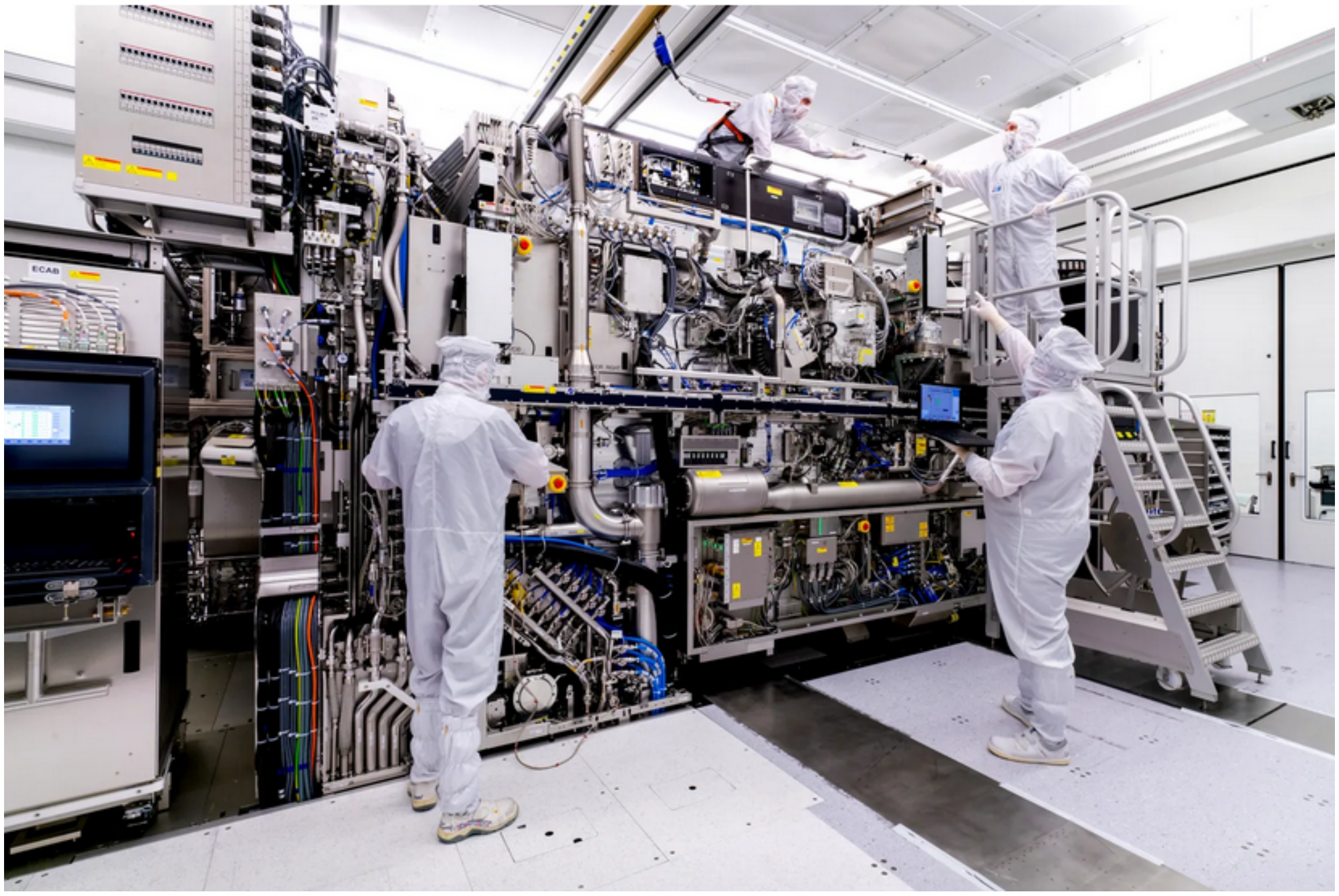

*The world's most complex semiconductor tool, an ASML Twinscan NXE. It costs ~$200M, is the size of a bus, and contains hundreds of thousands of components.*[138,139] *Just shipping it requires 40 freight containers, 20 trucks, and 3 cargo planes.*[140] *As of late 2022, only ~150 have ever been manufactured. ASML is backlogged for years, with €33 billion in orders it cannot immediately fulfill. Over 100 EUV tools have been ordered, and ASML is producing them at a rate of ~15 per quarter (40–55*[141] *in 2022 and ~60 in 2023).*[142] *The next generation version will be larger, more complex, more expensive (~$400M), and slower to produce (~20 in 2027–2028).*[143]

As hard as ramping capacity is for automotive and other semiconductor customers, it's a drop in the bucket compared to what may be needed to ramp capacity for a planet-scale workforce of AGIs.

---

[138] ASML is the only company making the $200 million machines needed to print every advanced microchip. Here's an inside look
[139] Chipmaking's Next Big Thing Guzzles as Much Power as Entire Countries
[140] The $150 Million Machine Keeping Moore's Law Alive | WIRED
[141] ASML expects to ship 55 tools in 2022 but only recognize revenue for 40, as 15 will be shipped before testing is completed on-site.
[142] TSMC and ASML: Demand for Chips Remains Strong, But Getting Fab Tools Is Hard
[143] ASML High-NA Development Update: Coming to Fabs in 2024 - 2025

**Top-down estimate:**

As of 2022, the world manufactures around **~2M wafers per year** of leading edge logic chips (5 nm[144] technology).[145] These chips have been in volume manufacturing at TSMC and Samsung since 2020, and reflect mature production lines running at high capacity (>90%).

NVIDIA's share of leading edge wafer production is perhaps ~0.5%[146] of the world's, suggesting that if humanity creates AGI or other high-ROI uses for NVIDIA GPUs, we *theoretically* have room to increase production ~100-fold by reallocating production away from iPhones and laptops and other chips.

NVIDIA data center revenue has grown ~9x over the past 5.5 years,[147] or ~50% per year. If growth continues at this blistering rate, it will reach $36T/yr in 2042, and require approximately 50M wafers per year, or 25x the world's current production at 5 nm. A 25x increase over 20 years means that wafer production needs to double every 4–5 years just for AGI chips alone, which will be on top of demand for increasingly fruitful applications of narrow AI and other compute-intensive non-AGI use cases. This scenario feels aggressive. But it would still leave us about twentyfold short of the highest end estimate of compute requirements above.

**Bottom-up estimate:**

Semiconductor fabrication depends on many inputs, but none is more critical than lithography tools.

An EUV tool processes ~160 wafers per hour. A logic wafer typically needs ~20 EUV steps[148], and assuming 84%[149] uptime, that's ~60k wafers per year. In 2022, ASML expects to ship 55 tools.[150] As of 2022, there are an estimated ~150 EUV systems in the world.[151] If half the world's EUV-powered supply chain was reoriented to build A100/H100 GPUs, in one year we could produce ~2M wafers. (At this rate, it would take 20 years to make just ~1 GPU per person on Earth.) This rate of ~2M wafers per year is in close alignment with the top-down figures, and gives us greater confidence in our estimates. Overall, we believe there's a ~50% chance that semiconductor production can be successfully ramped to meet the needs of training and inference. This estimate comes from the rough guesses in the table

---

[144] Confusingly, TSMC has processes for vanilla N5, performance-enhanced N5P, N4, N4P, N4X, and Nvidia-specific 4N, but they are all part of the N5 family.
[145] In April 2022, TSMC was rumored to be manufacturing 150,000 5 nm wafers per month (TechnoSports). In September 2022, Samsung was reportedly manufacturing 60,000 5 nm & 7 nm wafers per month, with the split unknown (Samsung's ups and downs (4): The manufacturing capability of Samsung's foundry biz). Assuming a 50/50 split of 5 nm / 7 nm for Samsung and constant production from TSMC over 2022, this works out to 2.16M 5 nm wafers per year.
[146] NVIDIA GPU wafers: 20k / yr. World 5 nm production: 2M / yr. This suggests a ratio of 1%, but because the bulk of NVIDIA's 2023 revenue will be from both N5 and N7 processes, we double the denominator to arrive at a ballpark figure of ~0.5%. This is a very rough estimate, but we think likely to be within an order of magnitude.
[147] This date is not cherry-picked to sandbag the probabilities. It's looking at 2018 forward, the era after NVIDIA entered the data center market with the V100.
[148] TechTaiwan: "To Lower 3nm Process Costs, TSMC Launches Plan to Reduce EUV Layers" : r/hardware
[149] EUV Challenges And Unknowns At 3nm and Below
[150] Increased EUV Shipments Help ASML Deliver Strong Growth in Q2 2022
[151] ASML is the only company making the $200 million machines needed to print every advanced microchip. Here's an inside look

below, in which we first apply a flat (log-normal) distribution over 7 possible outcomes of 5 orders of magnitude plus 2 tails, and then guess at the likelihood that each level of production will be achieved. We find the lower compute estimates highly likely but the upper estimates highly unlikely. Even in a world where AI chips have enormous ROIs, it will still take many years to build enough fabs to scale semiconductor production by 100x relative to today. (Also note that these quantities are far outside the industry's planning forecasts for the next decade, which are literally self-fulfilling.)

| # of wafers needed | wafer prod needed (5-yr lifetime) | implied 15-yr growth rate | odds of being needed | odds of being achieved |
|---:|---:|---:|---:|---:|
| <100k | <20k | <0% | 1/7 | 99% |
| 100k | 20k | 0% | 1/7 | 99% |
| 1M | 200k | 17% | 1/7 | 90% |
| 10M | 2M | 36% | 1/7 | 50% |
| 100M | 20M | 58% | 1/7 | 10% |
| 1B | 200M | 85% | 1/7 | 5% |
| >1B | >200M | >85% | 1/7 | 2% |
| | | | | **Mean: 51%** |

We therefore estimate a **51%** likelihood of success that enough semiconductor wafers will be manufactured for large AGI training runs and inference. Our uncertainty is wide, but we feel confident that it should not be close to 100%, given the immense construction requirements of the upper scenarios.

**What it will take to build power plants**

In addition to semiconductor fabs, power plants will also need to be constructed to run the GPUs.

Looking at training needs in 2038:

- We may need **20k to 1B wafers**
- We may need **2 to 10k GW-scale power plants**

Looking at inference needs for AGI to do 10% of global labor in 2038:

- We may need **300k to 30M wafers**
- We may need **30 to 3k GW-scale power plants**

If AGI requires building 1,000 GW power plants solely for AGI in ten to fifteen years, there is significant doubt whether such a thing will happen (and doubly so if those power plants need to be concentrated around giant training data centers or even just restricted to the USA due to increasingly stringent AI regulations).

To put this in perspective:

- The entire state of California generates ~30 GW[152] of electricity, and the entire US generates about 500 GW of electricity, and its infrastructure took many decades to build
- Nuclear power plants take about a decade to plan, approve, build, and connect
- Thermal power plants take ~4 years to construct[153]
- Renewable power plants take ~2 years to construct (but intermittent renewables by themselves are not suitable for datacenter workloads)
- Generation typically costs $1–$10 per W, and prices would undoubtedly rise if demand spikes by orders of magnitude, as factories pay overtime to build transformers and equipment around the clock; 100 GW plants could easily cost $1T, which is far beyond any private investment ever made

Optimistic scenarios are possible. Global solar installations are currently about 150GW of nameplate capacity (about 30GW levelized energy) per year, and growing at above 20% per year[154]. If this trend continues, by 2043 we might be producing over 1000 “plants” per year worth of solar, and it’s theoretically possible for this trend to accelerate. Wind, fission, and fusion may make significant contributions too. Cheaper batteries will only help. So, very high power numbers are possible on a global scale.

But on the other hand, much depends on the degree of concentration necessary. If, say, 10–100 GW power needs to be highly concentrated around a handful of gargantuan, high-bandwidth data centers for AGI training, it may be monumentally difficult to build that many GW in a metro or state-sized region. The necessary cooperation from utilities and regulators and citizenry would be immense and unprecedented. Furthermore, if the investment need is concentrated into a single company’s R&D efforts, it may be wildly difficult to achieve $100B or $1T in financing, supposing transformative AGI is a speculative venture that may or may not succeed. So even if it is technically possible to rush to build GWs of power plants, investors may prefer to finance more gradual deployments with time to see progress along the way. This is typical for growing companies; we find it hard to think of examples of startups scaling an input in wildly discontinuous fashions. The self-driving AI industry, for example, has gradually deployed self-driving technology, rather than spending a trillion to scale up fleets everywhere at once.

Nor will all new power be used for AGIs. Global electrical power will be undergoing a bit of a crunch in coming decades. Right now, global primary energy consumption is over 20 TW on a continuous basis[155], but only about a third of this amount is used to produce electricity[156]. Thus, as decarbonization is predicated on electrifying uses of energy that are currently not electric, thousands of “plants” of electricity will be needed just to sustain people’s current lifestyles and the global economy, with economic growth and AGI demands coming on top of this. Sparing a thousand “plants” for datacenters may not be possible in time, and sparing tens of thousands seems highly unlikely. Plus, in a world

---

152 2021 Total System Electric Generation
153 Average power generation construction time (capacity weighted), 2010-2018 – Charts – Data & Statistics - IEA
154 Installed solar energy capacity
155 Global direct primary energy consumption
156 Electricity production by source, World

where intermediate AIs have become fantastically good, future research efforts may need to compete for power (and GPUs) with those profitable but not-yet-transformative-AGI applications.

Lastly, the odds of success depend on the time available to build these power plants. If we started building them all today, the odds of constructing some concentrated patches of 10–100 GW might be quite high. But if it takes a dozen more years of software and hardware progress before the ROI of AGI is clear to investors and before more exact power requirements are understood, we may only have gangbusters power plant construction for single digit years before hitting the 2043 deadline of this forecast. The later AGI precursors are developed, the less likely we see high power requirements being met.

In the end, we are quite uncertain, both in terms of the likelihood of various power requirements and their likelihood of being achieved. For the former, we estimate a flat distribution over 5 orders of magnitude plausibly needed for AGI along with 2 tail outcomes. For the latter, we record our guesses in the table below, informed by the discussion above:

| # of GW plants needed | odds of being needed | odds of being achieved |
|---:|---:|---:|
| <1 | 1/7 | 99% |
| 1 | 1/7 | 99% |
| 10 | 1/7 | 95% |
| 100 | 1/7 | 67% |
| 1,000 | 1/7 | 33% |
| 10,000 | 1/7 | 5% |
| >10,000 | 1/7 | <1% |
| | | **Mean: 57%** |

We therefore estimate a **57%** likelihood of success that enough power generation can be acquired for large AGI training runs and inference. Our uncertainty is wide, but we feel confident that it should not be close to 100%, given the immense energy requirements of some scenarios.

### AGI seems unlikely to solve physical constraints

One key question is how amenable semiconductor fabrication, and/or energy infrastructure, is to AI-driven R&D. It's theoretically possible that expensive AGIs or narrow AI tools emerge in the 20s or 30s, and enable massive leaps in these areas. Some people will surely react to our analysis by saying that it doesn't matter, because the first AGIs will easily clear the way for the rest.

While this is theoretically possible, we don't feel that it's reliable, for a couple reasons, which have been fleshed out before. The time lag on these large, capital intensive projects is long, and they depend on scarce natural resources that also have long timelines. Because even expensive AGIs depend on the achievement of many of the milestones above, they will likely emerge late on the timeline toward 2043, when even less time is left to resolve the fundamental issues underlying these challenges.

And when AGIs emerge, they are likely, especially at first, to be human-esque, or superhuman, not omniscient. They won't be likely to solve logistical problems that thousands of smart humans have been working on for decades, and to do so immediately and completely.

However, we have constructed our estimates to be deliberately humble, accounting for at least the possibility of unexpectedly rapid growth, like wafer production growing at 85% per year for 15 years, or the growth rate of solar power installations doubling from its current heady level and the bulk of global power installation being available for AGI. These estimates may be considered to have some level of probability for AGI-driven manufacturing transformation baked into them.

## Our estimate and justification:

Combining this with the prior wafer analysis, we find that both silicon wafer fabrication and power plant construction make significant contributions to the risk, and together they lead us to a final estimate of **46%** likelihood of success.

| forecast of necessity | # of wafers needed | forecast of achieving wafers | # of GW power plants needed | forecast of achieving power | joint odds of achievement |
|---:|---:|---:|---:|---:|---:|
| 1/7 | <100k | 99% | <1 | 99% | 98% |
| 1/7 | 100k | 99% | 1 | 99% | 98% |
| 1/7 | 1M | 90% | 10 | 95% | 86% |
| 1/7 | 10M | 50% | 100 | 67% | 34% |
| 1/7 | 100M | 10% | 1,000 | 33% | 3% |
| 1/7 | 1B | 5% | 10,000 | 5% | <1% |
| 1/7 | >1B | 2% | >10,000 | <1% | <1% |
| | | | | | **Mean: 46%** |

(We recognize there may be scenarios in which chip compute efficiency and energy efficiency progress differently in the next decade, thereby decoupling the wafer and power needs. However, we expect wafer needs and power needs to be more correlated than not, and so, to simplify our already uncertain and lengthy analysis, we consider them jointly linked.)

# Element 6: Humans, and AGIs, don't purposely slow AI progress after seeing its trajectory

## What it is and why it's necessary:

Humans do not invent every technology we could invent, nor deploy every technology we do invent. There are lots of examples of technologies which certain civilizations never invented despite having the ability, like mesoamerican civilizations with the wheel, Chinese civilization with phonetic alphabets, or Europeans with ice in beverages. But in addition, many technologies are invented, but never deployed, or never deployed to their potential, due to safety concerns, cultural taboos, or random events. These include biological weapons since approximately the 1940s, nuclear power since the 1970s, human cloning since the 1990s, direct-to-consumer genetic testing in the USA from 2013 to 2017, and human germline editing since about 2017. And in addition to this, in an analogous case, resources that exist in various countries are often not exploited due to political decisions and instability; three of the top 10 oil-bearing countries (Venezuela, Canada, and Libya) are producing less than half the oil, relative to their reserves, of leading producers.

The transformative potential of AI, particularly AGI, is already a major news topic in the developed world on a chronic basis, focusing not only on existential risk, but on smaller matters like individual safety, employment security, social justice, and the ethical obligations humans would have toward AGI if and when we create it. These concerns have already led to organized efforts to guide the development of AGI, even at the cost of impeding it, by being discreet rather than forthright about certain kinds of information (including the contest's instruction not to publicly release potentially harmful essays submitted!), and abjure some applications of AI (like bans on facial recognition and armed robots). There are already societal campaigns for AI civil rights (like Blake Lemoine) and AI Ethics (like Timnit Gebru). These concerns have also translated into government discussion about regulation of AI R&D and industrial deployments of AI. And there is already significant geopolitical tension over elements of the AGI supply chain like semiconductor manufacturing and 5G devices. The issue is already quite salient.

As AGI becomes closer, and AI technologies display more and more capability, governments, companies, and society at large may decide to take significant measures along these lines, with the effect of halting or slowing progress:

- The judges believe there is a 5%–50% chance of existential risk from AGI (conditional on its development by 2070); if society as a whole comes to believe the same, they may take action to stop or slow AGI development.
- Humans could impede AGI development due to ethical concerns about AGI. The assumption that AGIs may be instantiated, manipulated and deleted at will, and freely used for labor at only the cost of operation, with the surplus fully accruing to their human owners, may not remain a consensus position as human morality changes.
- Humans could also impede AGI development for less enlightened reasons. The AGI supply chain, from neon to fab tools to machinery manufacturing to software, is global. If one country or bloc of countries, or entities headquartered there, are ahead, will rivals let them proceed, knowing how powerful AGI could prove to be? Or will competition-driven instability delay AGI development?

- In fact, as AGI is invented, even AGIs themselves may take action, or advise humans to take action, to slow or halt the future progress of AGI. This could be because AGI entities more intelligent than humans have the inherent insight to anticipate X-risk from further development, or for reasons of their own (including X-risk to a current generation of self-interested AGIs from the development of subsequent AGIs).

Even a relatively small delay, from any of these sources or others, could be decisive with the timeline until 2043 already so tight.

## Our estimates and justifications:

We think it's most likely humans and early AGIs will not stall this technology despite its risk, because of its potential salutary impact on the economy and human lifestyles. We guess that there's a 70% chance humans will not stall it, and a 90% chance early AGIs will not.

# Element 7: AGI progress is not derailed by war, pandemics, or economic depression

## What it is and why it's necessary:

As AGI technology is researched and scaled over the next 20 years, there is no guarantee that human civilization and the global economy will continue as normal. In the event of a world war, severe pandemic, or global economic depression, AGI progress may be delayed by years.

In particular, we believe that transformative AGI may be especially sensitive to potential derailments due to the tight 20-year timeline. Because transformative AGI is difficult, and in our view unlikely to occur by 2043, we believe that *conditional on transformative AGI being developed by 2043*, it is more likely than not to occur at the tail end of the period, say, in the final five years. Therefore, if a historical event delays progress by only a few years, we believe that it already has a decent chance of pushing AGI beyond 2043.

Lastly, although severe wars, pandemics, or depressions may feel very rare (because they are), it is worth emphasizing that a 'once a century' event is ~20%[157] likely to occur within 20 years. Any one of five 'once a century' events is ~63% likely to occur within 20 years. These probabilities are large enough that it would be a mistake to neglect 'once a century' events in any 20-year forecast of AGI progress.

## Our estimate: War

The odds of severe war are not insubstantial. War has the potential to delay transformative AGI by many years.

**One way that war can delay progress is by reducing humanity's ability to fabricate semiconductors.** Semiconductor supply chains are highly complex, highly global, highly entwined with Taiwan and the United States and South Korea[158] (e.g., TSMC, NVIDIA, Samsung, etc.). Wars can destroy long-term production capacity in many ways:

- War can destroy data centers in which AIs are being trained and researched
- War can destroy power plants or electric infrastructure or fossil fuel supply chains that threaten the ability to power AI data centers
- War can destroy semiconductor fabs producing chips for data centers
- War can destroy the manufacturing sites for key inputs to semiconductor fabs (e.g., lithography tools, optical components, or other critical components made by specialists at single locations on Earth)
- War can halt global shipping and trade needed to supply inputs to manufacturing
- War can destroy the demand for goods produced by the semiconductor industry, thereby destroying investment in continued production

---

[157] Technically 18.2%, but ~20% with one significant figure.
[158] Largest semiconductor companies by market cap

- War can destroy the human capital of the industry if workers are drafted into combat and if the education of the next generation is delayed
- War efforts can redirect compute and semiconductors toward the war effort, and away from AGI research
- War can decrease investor confidence by mere threat of all the above, thereby reducing available funding for AGI projects

Depending on the damage, rebuilding lost or deferred physical capital could take many years, and delay transformative AGI beyond 2043.

**War can also delay progress by reducing humanity's ability to improve AI software.** AI human capital is heavily concentrated in the San Francisco Bay Area and London, which house AI research labs and startups such as OpenAI, Anthropic, and Google DeepMind. Losing thousands of AI experts who live in these cities would plausibly set AI progress back by years. The loss of AI experts could happen via death, draft, or evacuation. (And if those losses were targeted, they could have a chilling effect on survivors wishing to continue the work.)

**Even cold wars or trade wars between great powers that simmer below full-scale nuclear war have the potential to drastically curtail world semiconductor production.** For example, consider the 'best case' scenario of China invading Taiwan—they invade quickly without bloodshed, and other nations eschew military escalation and instead sanction China by halting trade. Even this 'best case' scenario would drastically damage world semiconductor manufacturing! TSMC Chair Mark Liu has stated that if China invades Taiwan, TSMC fabs will not be operable: "Nobody can control TSMC by force. If you take a military force or invasion, you will render TSMC factory not operable…. Because this is such a sophisticated manufacturing facility, it depends on real-time connection with the outside world, with Europe, with Japan, with U.S., from materials to chemicals to spare parts to engineering software and diagnosis."[159] No doubt with enough time, China could internally rebuild the supply chains needed to run TSMC's fabs, but this would take years, and even then, might not be willing or able to trade the output to non-Chinese AGI efforts, as the rest of the world would at minimum apply harsh economic trade sanctions. This could potentially delay transformative AGI to 2043 or beyond.

**There is no guarantee that the (relative) peace between great powers will continue forever.** Russia's invasion of Ukraine has demonstrated plainly that the age of war is not behind us, especially for countries led by dictators who feel they're losing power. China too has spoken again and again in words that are quite plain about how it plans to reunify (invade) Taiwan. These two conflicts have the potential to cascade into global war, along with unknown conflicts in the future.

**Forecasters predict roughly a 1/3 chance of major war before 2043.** As we are not experts in forecasting war, we survey below quantifiable predictions of various war scenarios. We have not cherry-picked the results or purposefully omitted predictions unfavorable to our argument.

---

[159] Apple chipmaker TSMC warns Taiwan-China war would make everybody losers

**Forecasts of short-term nuclear war:**

| Prediction | Event | Source | Date of prediction |
|---|---|---|---|
| 6% | Russia will detonate a nuke in the next 2 months (Nov/Dec 2022) | Polymarket ($156K volume, $29k liquidity) | 2022-10-26 |
| 11% | Hostile nuclear detonation in the next 2 months | Hypermind Prediction Market ($1k prize) | 2022-10-27 |
| 15% | Russia or the US raise their nuclear alert levels in the next 2 months | Insight Prediction Market ($6k volume) | 2022-10-27 |
| 9% | Hostile nuclear detonation in Europe in the next 6 months | Swift Centre | 2022-10-13 |
| 2% | A NATO country directly intervenes with ground troops in the next 6 months | Swift Centre | 2022-10-13 |
| 16% | Russia detonates a nuclear weapon in Ukraine in the next 12 months | Samotsvety | 2022-10-04 |
| 5% | Russia detonates a nuclear weapon in Ukraine in next 12 months | Good Judgment | 2022-11-18 |
| 55% | Russia and Ukraine do not announce peace in next 24 months | Good Judgment | 2022-11-18 |
| 1.6% | Nuclear war escalates outside of Ukraine in the next 12 months | Samotsvety | 2022-10-04 |
| 17% | Imminent global nuclear war | Max Tegmark | 2022-10-07 |
| 5% | Russia will detonate a nuke in the next 1.5 months (May/June 2023) | Polymarket ($44K volume, $12k liquidity) | 2023-05-15 |
| 5% | A nuclear weapon will be used to kill in the next 6.5 months (2023) | Hypermind Prediction Market | 2023-05-15 |

In the fall of 2022, we gathered forecasts of imminent nuclear war and found they ranged from 1.6%–17%. In May of 2023, the risk still appears elevated: Polymarket frighteningly predicts a 5% chance in the next ~45 days, whereas Hypermind forecasters predict 5% in the next 6.5 months. Although we'd normally place our faith in markets with skin in the game (e.g., Polymarket), we also recognize that prices can be heavily distorted at the tails, where $1 of a bad actor's money needs to be matched by a profit seeker's $19. Furthermore, in a world with near-risk-free nominal interest rates of ~5%/yr, probabilities might be off (in either direction) by as much as ~5%/yr.

Conditional on escalation of the war with NATO powers, there is still a fair chance that the war is limited enough to not substantially delay AGI timelines. If we assume a 2% chance of escalation and a 25% chance of the escalation being severe enough to delay AGI timelines, that implies a 0.5% chance that transformative AGI is delayed (or a 99.5% chance that AGI is not delayed).

**Forecasts of world war:**

| Prediction | Event | Source | Date of prediction |
|---|---|---|---|
| 31% | A non-test nuclear detonation before 2035 | Metaculus | 2023-05-14 |
| 45% | Great power war by 2043 | Metaculus | 2022-10-26 |
| 23% | World War 3 by 2050 | Metaculus | 2023-05-14 |
| 2% | Nuclear war kills 10%–100% of humans by 2050 | Median forecast of ~30 superforecasters and subject matter experts in Tetlock's Hybrid Forecasting-Persuasion Tournament[160] | 2022-10-30 |
| 49% | China-US conflict kills 100+ by 2050 | Metaculus | 2023-05-14 |

Longer-term forecasts of world war are harder to come by. Metaculus predicts something in the range of 20%–45%, but given the lack of monetary incentive, we don't attach much credibility. We attach more credibility to the superforecasting group in Tetlock's Hybrid Forecasting-Persuasion Tournament, whose 2% we consider a lower bound, as there may be nuclear wars large enough to delay AGI that still fail to kill 10% of humans.

We split the difference between these estimates, and forecast a 10% chance of great power nuclear war with a conditional 70% chance that the war is severe enough to delay transformative AGI beyond 2043. This implies a 7% chance that transformative AGI is delayed by nuclear war (or a 93% chance that AGI is not delayed).

We also note that this is inclusive of the short-term scenario above.

**Forecasts of Chinese invasion of Taiwan**

| Prediction | Event | Source | Date of prediction |
|---|---|---|---|
| 14% | China-Taiwan conflict with 100+ deaths, within 5 years | Six superforecasters (median) | 2021-07-06 |
| 75% | Conditional on invasion, that US tries to sink Chinese ships | Six superforecasters (median) | 2021-07-06 |
| 17% | Claims of military or police weapon use between China and Taiwan in 2023 | Good Judgment | 2022-11-18 |
| 6% | Armed conflict between Taiwan and China by 2023 | Insight Prediction Market (~$8k) | 2022-11-03 |

[160] (in particular, the median forecast of superforecasters and subject matter experts on Ted's team)

| | | | |
|---|---|---|---|
| 7% Yes / 71% No (median confidence: 7 of 10) | China uses military force against Taiwan in next year | Snap poll of international relations experts (who correctly predicted Russia-Ukraine in Jan by 3:1 margin) | 2022-03-14 |
| 5% | China invades Taiwan by 2025 | Metaculus | 2023-05-14 |
| 29% | China invades Taiwan by 2030 | Metaculus | 2023-05-14 |
| 40% | China invades Taiwan by 2035 | Metaculus | 2023-05-14 |
| 11% | China blockades Taiwan by 2025 | Metaculus | 2023-05-14 |
| 35% | China blockades Taiwan by 2030 | Metaculus | 2023-05-14 |
| 45% | China blockades Taiwan by 2035 | Metaculus | 2023-05-14 |
| 23% | China-Taiwan conflict with 100+ deaths by 2026 | Metaculus | 2023-05-14 |
| 68% | China-Taiwan conflict with 100+ deaths by 2050 | Metaculus | 2023-05-14 |

Forecasts of conflict between China and Taiwan are worryingly high.

We ascribe high credibility to the polls of international relations experts, who correctly predicted the Russia-Ukraine conflict by a 3:1 margin, months before it erupted. However, extracting forecasts from their polls is unclear. Interpreting the median respondent's 7/10 confidence level as a 70% chance of no invasion would imply a 30% chance of invasion per year, which sounds irresponsibly high. Over the past three years, small fractions of respondents—6%, 11%, and 7% (geometric average of 8%)—have gone as far as to predict military conflict in the next year (with a range of confidences). If those fractions of respondents are interpreted as rough likelihoods, it perhaps implies a 5-year risk of 34%, a 10-year risk of 57%, and a 15-year risk of 71%.

We also ascribe high credibility to the superforecasting group, whose **5-year risk of 14%** would extrapolate out (again, assuming a constant risk of ruin) to a **10-year risk of 26%** and a **15-year risk of 36%**.

These figures are in decent agreement with Metaculus, where 264 forecasters have made ~2,000 forecasts that average out to a **40% risk by 2035**.

In sum, ***all*** quantified forecasts we located predict substantial risks of conflict between China and Taiwan.

Our own independent research corroborates these forecasts.

**We see many factors indicating China may invade Taiwan:**

- China has said—plainly and repeatedly and consistently—that unifying with Taiwan is its "greatest national interest"[161]

[161] China's defense minister says resolving 'Taiwan question' is a national priority

- An online survey from 2020–2021 showed a majority of Chinese people (55% vs ~33%) support “launching a unification war to take back Taiwan entirely”, though only after trying other options; only ~22% were fine with “unification not necessarily being the endgame”[162]
- Taiwan has little interest in peacefully unifying with China
- Xi has ordered China’s military to develop the capability to take Taiwan by 2027[163]
- China has been investing in military amphibious lift capabilities[164]
- China’s navy has substantially increased its encroachment on Taiwan’s waters[165]
- Xi is unifying power,[166] which increases the variance of outcomes
- Xi is about to age into his 70s and 80s,[167] which increases the variance of outcomes
- Anecdotally, many of our Chinese-American friends, who are more knowledgeable than we, have expressed high certainty that China will attempt to invade Taiwan in the next decade (we are skeptical of the certainty, but hearing this consistent message from diverse sources has adjusted our personal forecasts up)

**We also see a few factors against China invading Taiwan:**

- The base rate is low; China and Taiwan have had 44 straight years of unofficial ceasefire, since the Second Taiwan Strait Crisis in 1979[168]
- Taiwanese stocks are still trading at decent valuations (TSMC at ~13 p/e as of May 2023), and we trust the million and billion dollar prizes implicit in the stock market more than polls and small prediction markets
    - Then again, this data point only precludes invasion occurring both soon and with high likelihood; a p/e of ~13 could theoretically be consistent with a 100% chance of invasion in 13 years or a 50% chance of invasion discounting off of a ‘safe’ p/e of 26; both scenarios might severely delay AGI timelines
- Russia’s attempted invasion of Ukraine has reminded the world of the sharp economic costs of modern war and the many advantages accruing to a determined defender with outside support, which may dissuade China
    - On the other hand, it has also shown the reluctance of the US to enter a costly war, which may embolden China

**Given the factors above, the 15-year chance of 36% implied by superforecasters feels reasonable to us.** We also emphasize again that an invasion of Taiwan is likely to delay transformative AGI timelines by years, if it occurs before enough computing hardware has been built for transformative AGI. Even if direct war between China and the USA is avoided in an invasion, it is likely that TSMC’s semiconductor output will be drastically reduced because of scenarios such as:

---

162 Just over half of mainland Chinese people back full-scale war to take control of Taiwan, poll finds | South China Morning Post

163 https://www.thedefensepost.com/2022/09/21/china-seize-taiwan-us-intel/; https://edition.cnn.com/2022/09/19/asia/biden-us-troops-defend-taiwan-intl-hnk/index.html

164 Japan urges Europe to speak out against China’s military expansion

165 https://www.theguardian.com/world/2022/sep/07/china-using-cognitive-warfare-to-intimidate-taiwan-says-president-tsai

166 Analysis: Xi's new generals offer cohesion over possible Taiwan plans | Reuters

167 Xi Jinping - Wikipedia

168 Chinese Civil War - Wikipedia

- Taiwan government sabotaging the fabs
- TSMC employees sabotaging the fabs
- Taiwan loyalists sabotaging the fabs
- USA or anti-China forces sabotaging the fabs
- China sabotaging the fabs if the USA or other forces attempt to free Taiwan
- Accidental destruction of fabs during fighting
- Fab shutdowns due to lack of skilled labor (if TSMC employees refuse to work, as Chinese workers unfamiliar with the precise processes will be unlikely to operate the fab at economically viable yields)
- The fabs shutting down due to lack of inputs (e.g., imports of neon gas, tool parts & support)
- The fabs shutting down due to remote disabling of software needed by semiconductor tools
- Nothing shutting down after a successful and peaceful invasion, but investment in new fabs halting due to the climate of uncertainty
- Nothing shutting down after a successful and peaceful invasion, but investment in new fabs halting due to China extracting profits to help buttress losses from Western economic sanctions
- TSMC output continuing, but being redirected to Chinese markets and away from western AGI efforts
- No war at all, but the risk of a war making investment in infrastructure in Taiwan questionably wise and resulting in a slowdown of semiconductor activity compared to a little-risk-of-war scenario

In short, we expect the typical outcome of an invasion is that TSMC's output going to AGI will be drastically reduced for many years. This will slow transformative AGI timelines by years, as TSMC is the #1 producer of advanced semiconductor chips and makes 100% of advanced AI chips (NVIDIA's V/A/H100, Google's TPU v1/v2/v3/v4, Cerebras's WSE/WSE-2). TSMC's major fabs in Taiwan will shut down, or at least stop exporting chips to companies building AGI. In the medium term, production will shift to Samsung and Intel, as well as TSMC fabs outside of Taiwan (e.g., Arizona). However, moving chip production to new fabs will take years, as (a) there will be no free capacity in other fabs and (b) it's time consuming to repurpose a design to a new manufacturing process and for the fab to ramp up production with decent yields. We also expect prices to skyrocket, as AGI research dollars try to outbid Apple cell phone chips and other highly profitable in-demand chips.

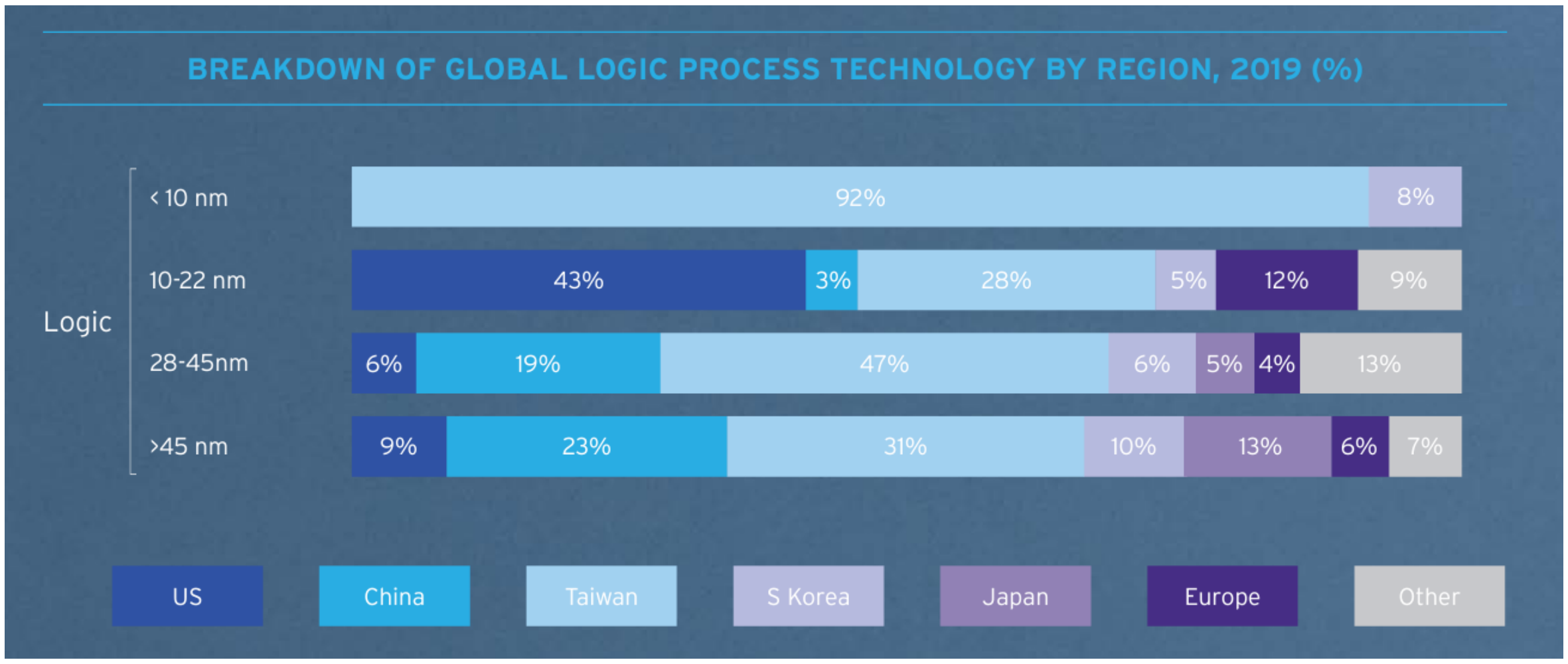

*As of 2019, 92% of leading edge semiconductor manufacturing was in Taiwan.*[169]

**Lastly, we see the development of AGI as potentially VERY destabilizing.** Imagine how China or Russia or any other power will react if it were to become clear that the United States or a US company was close to achieving AGI, or had already achieved AGI and was in the process of making it cheaper, or had already achieved cheap AGI and was in the process of scaling it to a global workforce. These are near-existential threats with only a brief window in which it may be possible to react. If a US company were on track to put US workers out of their jobs, it's plausible that the US government would force redistribution of wealth, or even nationalize the AGI technology. But if you were the leader of Russia, how confident would you be that your geopolitical foe would altruistically redistribute its AGI wealth with Russian citizens about to face mass unemployment due to cheaper AGI workers? Another country developing AGI is a direct threat to your exports and economy at large. An attack on companies or nations building AGI—whether nuclear, conventional, cyber, or economic—could easily delay transformative AGI by years.

**Summarizing:**
- Semiconductor supply chains are so complex and global that a war involving the USA or Taiwan would likely delay transformative AGI by years.
- Expert forecasters predict substantial likelihoods of great power wars: 2%–10% in the next year, and as high as 20%–45% by 2043.
- Expert forecasters across many groups all expect a high chance of conflict between China and Taiwan, with extrapolated 15-year estimates ranging from 36%–71%.
- Progress in AGI will be destabilizing; therefore, conditional on successful progress in AGI, we expect the odds of war will rise.

**Conditional on being on a trajectory to transformative AGI, we forecast a <u>40% chance of severe war erupting by 2042</u>.** Not being experts in war, we lean heavily on other forecasters, who appear to predict a 36%–71% chance China attacks Taiwan as well as a 20%–45% chance of global war (though these scenarios may overlap substantially). Given the relative peace of our ~35-year lifetimes, we lean toward lower forecasts (perhaps ~30% and ~10%). However, *conditional* on a world where

[169] 2021-SIA-State-of-the-Industry-Report.pdf

transformative AGI is being successfully developed, we expect the odds of war to rise substantially due to (a) AI progress increasing the incentive for war, as described above, and (b) AI progress increasing capabilities for war and potentially destabilizing current balances of power. These factors adjust our estimate substantially upward. Our subjective judgment call is to end up at a 40% chance of severe war, with most of it coming from the uncertainty around Taiwan. If you disagree, feel free to substitute your own estimate.

**Conditional on (1) being on a trajectory to transformative AGI and (2) severe war erupting, we forecast an ~75% conditional probability that the war pushes transformative AGI beyond 2042**. We expect invasions of Taiwan or global nuclear wars to impact AGI timelines by many years, though there may be cases where the fighting is limited, or erupts late enough that transformative AGI is already close to completion and ends up not being delayed by enough to push it beyond 2042.

**Therefore, we estimate a 70% chance that war does not delay AGI progress beyond 2042.** Forecasting war is hard to root in any sort of justified quantitative analysis, but we find it hard to have confidence above 80% given the uncertainty here and the many plausible scenarios by which war could ignite (e.g., Russian invasion of Ukraine escalating into world war, Chinese invasion of Taiwan escalating into world war, future trade wars, future AI tech cold wars, and many conflicts yet to be imagined). We consider AI progress to be a destabilizing factor that increases the conditional probability of war. We consider a Chinese invasion of Taiwan to be the most likely scenario in which transformative AGI is delayed past 2042.

## Our estimate: Pandemics

The odds of severe pandemics are not insubstantial.

**COVID19 was a "mild" pandemic; severe pandemics may delay transformative AGI by years.** As COVID19 has made clear, even "mild" pandemics can kill millions and tangle semiconductor supply chains for years. Severe pandemics have the potential delay transformative AGI timelines for many years, by:

- Throwing the world into economic recession or depression
- Reducing investment in semiconductor fab construction
- Tangling semiconductor supply chains due to lockdowns, border closures, staff shortages
- Killing key AGI researchers
- Reducing AGI research productivity

**There are good reasons to think the odds of pandemics in the 2000s will be much higher than the 1900s.**

**The risk of natural pandemics is now higher than at any point in history.** As human populations grow and increasingly encroach upon animal habitats, the emergence rate of infectious disease is rising. From the 1600s to the 1900s, the frequency of extreme epidemics tripled.[170] From 1940 to 2000, the emergence rate of small-scale infectious diseases quadrupled.[171] Extrapolating these trends, a non-cherry-picked paper estimates that the yearly probability of extreme epidemics will triple again in the coming decades, implying a ~1% annual risk of a Spanish flu-like pandemic.[172] (The Spanish Flu is estimated to have killed about ~1% of humans, roughly ~10x COVID19's toll so far.) **A ~1% annual risk of a Spanish flu-like pandemic implies an 18% chance by 2043.** This 18% chance is in-line with informed estimates from other non-cherry-picked sources.

[170] Intensity and frequency of extreme novel epidemics | PNAS

**Forecasts of natural pandemics:**

| Prediction | Event | Source | Date of prediction |
|---|---|---|---|
| 66.8% | COVID-like pandemic within 20 years (~2% annual risk) | Intensity and frequency of extreme novel epidemics (2021) | 2021 |
| 60% | Natural pandemic kills >1M people by 2100 | Global Catastrophic Risks Survey | 2008 |
| 18.2% | Spanish flu-like pandemic in 20 years (~1% annual risk) | Intensity and frequency of extreme novel epidemics (2021) | 2021 |
| 5% | Natural pandemic kills >1B people by 2100 | Global Catastrophic Risks Survey | 2008 |
| 3% | Natural pandemic that kills 1%–100% of humans, by 2050 | Median forecast of ~30 superforecasters and subject matter experts in Hybrid Forecasting-Persuasion Tournament | 2022 |

The majority of these forecasts offer numbers at or above the 1% annually number, and the two which do not are forecasting extremely severe pandemics worse than the 1918 pandemic. Nor are they particularly cacaphonistic; one of the five has already been borne out.

**The risk of engineered pandemics is also rising rapidly.** The risk of accidental engineered pandemics has likely been rising as bio-labs proliferate across the world. The risk of intentional engineered pandemics is also rising as genetic engineering technology and information becomes cheaper, better, and more ubiquitous. In particular, if we are on trajectories to develop transformative AGI, it seems likely that AI technology will be an accelerant for technology enabling engineered pandemics. Computational design of toxic and pathogenic agents is already a threat, one which will likely escalate as biotechnological AI advances.

**Forecasts of engineered pandemics:**

| Prediction | Event | Source | Date of prediction |
|---|---|---|---|
| 30% | Engineered pandemic kills >1M people by 2100 | Global Catastrophic Risks Survey | 2008 |
| 10% | Engineered pandemic kills >1B people by 2100 | Global Catastrophic Risks Survey | 2008 |
| 10% | Bird flu pandemic caused by accidental release before 2043 (~0.5% annual risk) | Lynn C. Klotz, Biological and Chemical Security Program, Center for Arms | 2020 |

[171] Global trends in emerging infectious diseases
[172] Intensity and frequency of extreme novel epidemics | PNAS

| | | | |
|---|---|---|---|
| | | Control and Non-proliferation | |
| 3% | An engineered pandemic that kills 1%–100% of humans by 2050 | Median forecast of ~30 superforecasters and subject matter experts in Hybrid Forecasting-Persuasion Tournament | 2022 |
| 3% | Engineered pandemic that kills *all* humans or human potential, within 100 years | The Precipice, by Toby Ord | 2020 |

The experts we found seem aligned on single digit odds of severe engineered pandemic in the next couple decades.

The Global Catastrophic Risks Survey predicted a 10% chance of >1B people killed within 92 years. If we make the simplifying assumption of constant risk (doubtless wrong, but hopefully not too wrong), that implies a ~0.1% annual risk, which multiplies out to a risk of ~2.3% by 2042.

The Hybrid Forecasting-Persuasion group's median predicts a 3% chance by 2050, perhaps implying a ~2% risk by 2042.

And Toby Ord's book The Precipice predicts a 3% chance of *extinction* in 100 years; making the same simplifying assumption of constant risk, that implies a ~0.6% risk in 20 years. And if we further suppose that an AGI-derailing pandemic is 10x more likely than a literal *extinction* pandemic, which requires an extraordinary pathogen capable of reaching or permanently surrounding every island and uncontacted tribe, the odds might look more like ~6%.

Overall, we perceive decent agreement from expert forecasters that the odds of an engineered pandemic killing tens of millions to billions are low but non-zero, perhaps in the range of 2%–6%. On the one hand, some of these severe pandemics still might not derail transformative AGI, if transformative AGI is invented prior to the pandemic, which might suggest a lower range like 1%–3%. On the other hand, we expect that proliferation of improved biotechnology and AI will increase the likelihood of future engineered pandemics, which might suggest a higher range like 4%–12%. And of course, if these forecasters are considering scenarios where transformative AGI releases an engineered pandemic, that, by definition, will not qualify as a delaying pandemic. Ultimately we split the difference and estimate a 5% chance that a severe engineered pandemic delays transformative AGI beyond 2042.

**Summarizing:**

- A severe pandemic could delay transformative AGI by years
- There's perhaps an ~18% chance within 20 years of a Spanish flu-like pandemic that kills >1% of all humans
    - Progress toward transformative AGI will likely accelerate the potential to develop countermeasures, lowering the odds of delay
- There's perhaps a ~3% chance within 20 years of an engineered pandemic that kills ~1% of all humans

- Progress toward transformative AGI will likely accelerate the potential of bad actors to create engineered pandemics, increasing the odds of delay

**We estimate a 95% chance that natural pandemics do not delay AGI progress beyond 2043 and a 95% chance that engineered pandemics do not delay AGI progress beyond 2043, for a total probability of 90%.** Again, this is hard to root in hard quantitative analysis, so you may reasonably disagree. We personally find it hard to have confidence approaching 100%.

These low probabilities of pandemic reflect the many ways things can go right. Maybe there's no pandemic at all. Maybe the pandemic is mild. Maybe it's severe, but we rush vaccines and other countermeasures. Maybe AI tools help us invent new countermeasures. Maybe it's severe, but the delay to transformative AGI is minimal enough that it still leaves time to develop transformative AGI by 2043. Maybe the pandemic occurs after transformative AGI is developed. Overall, we are quite optimistic that AGI progress is unlikely to be delayed beyond 2042 by a pandemic. However, given the arguments above and our own uncertainty, we find it difficult to stray far from the expert consensus estimates, and therefore difficult to assert more than 90% confidence.

## Our estimate: Economic Depression

In the last century, the developed world suffered only one recession widely regarded as a depression, and even during this depression, technological progress broadly continued happening; lots of things were invented in the 1930s. On the other hand, utility patent applications in the USA dropped by over 40%, and didn't reattain their pre-depression high until the 1960s[173]. So you could consider the Great Depression a technology-slowing depression, and measured this way, the only one in US history. If you chose to view it as one such event in about 200 years, there may be a 10% chance of such an event in the next 20 years.

From another point of view, even milder recessions may have slowed individual developments significantly. E.g. the recession of 1969, a mild and short one, may have played a decisive role in slowing the development of interplanetary space travel by decades, due to its effect on congressional budgeting of follow-ons to the Apollo program.

Conditional on being in a world on track toward transformative AGI, we estimate a ~0.5%/yr chance of depression, implying a ~10% chance in the next 20 years. Of course, some of these depressions may not derail transformative AGI, and some might occur after transformative AGI is achieved. Assuming half of depressions do not delay transformative AGI via their chilling effect on research and/or investment, we forecast a healthy 95% chance that transformative AGI is not derailed by depression.

**In line with published expert opinions and some historical experience, we estimate a 70% chance of no disruption due to world war, 90% chance of no disruption due to a pandemic (natural or engineered), and 95% chance of no disruption due to economic depression.**

---

[173] https://www.uspto.gov/web/offices/ac/ido/oeip/taf/h_counts.htm

*Note: All potential derailments are only derailments if they occur well prior to the development of transformative AGI. We purposely neglected detailed discussions of timelines as we expect that* ***if*** *transformative AGI is developed in the next twenty years, it will most probably be developed at the tail end of this period. If you believe transformative AGI is likely to be developed within 5 or 10 years, you may want to substantially reduce our forecasts of derailment.*

# Discussion

Above, we have presented our estimates for the probabilities of all the key events in our cascade, and they result in a very low probability estimate, at a mere 0.4%. This is much lower than your current estimation, and lower than our intuition suggested before we started trying to formalize an answer to this problem. Despite this, it's the best estimate we, personally, are able to come up with right now.

Part of the reason we're unwilling to yield to our intuition that the chance must be higher is that the individual estimates were undertaken with such care not to bias things against the AGI hypothesis. When considering our individual probabilities, some of them seem quite optimistic to us (although not optimistic enough that we feel confident in lowering them).

## Try it yourself

We encourage you, rather than making a binary choice between accepting our estimate of 0.4%, on the one hand, or rejecting this essay entirely, on the other, to use this framework and chart your own course. You can input your own estimates here.

In addition, there are other ways you can use these probabilities to express beliefs different from ours:

- If you believe that one or more of these conditions is unnecessary for transformative AGI, then you can remove it from consideration by setting its probability to 100%. You may want to do this if you believe that robot bodies aren't necessary for AGI to be transformative, or if you believe that active learning methods could be done at accelerated speed so that AL-free methods are unnecessary, for example.
- If you think that a subset of these conditions are not independent of each other (conditionally), but highly correlated, so that they are essentially coterminous, then you can choose which one element of that subset you think is least likely, and set the others all to 100%. For example, you might think that progress in manufacturing robot bodies and silicon computation will be so correlated that if one happens, they both do.
- If you think there are other predicate events that we missed, and you have probability estimates for them, then you can insert them.

We hope you'll spend some time playing with it.

## It's hard to support high probabilities

You'll find that it's surprisingly punishing, and rather hard to get to even 10%. For example, doing either of the following will not get you there:

- Completely removing all software factors and all sociopolitical factors
- Assuming that robot bodies and semiconductor manufacturing scaling are 100% guaranteed, and assigning 80% probability to the 5 orders of magnitude improvement in computational efficiency

You could get to barely above 10%, barely into the pre-contest range of credence of the judges, but not much above, by taking an extreme set of assumptions like:

- Doubling the probability of all software and hardware factors (capped at 100%) and removing the possibility of human and AI decisions to slow development

Reaching a probability substantially in excess of 10% is even harder. We think most of the range of credence values of the judges, pre-contest, are essentially unsupportable within this framework, and should be revised downward substantially.

# Is our framework a trick?

The reason it's so hard to get high probabilities from our framework is that it assumes the probability of transformative AGI can be expressed by the product of many uncertain conjunctive factors. Once this structure is accepted, small probabilities are nearly guaranteed, as multiplying a series of numbers less than 1 will yield a much smaller number.

So it's worth considering deeply whether our framework is valid.

There are two main sleights of hand to beware of when people model probabilities as a product of conditional probabilities:

- They can fail to model parallel disjunctive paths by which an outcome is achieved
  - For more, see: Gwern's Technology Forecasting: The Garden of Forking Paths.[174]
- They can fail to full grapple with the idea of conditional probabilities, and accidentally or purposefully smuggle in unconditional probabilities
  - This is potentially why Nate Silver was so wrong in 2015 when he suggested Trump had only a 2% chance of winning the Republican nomination; he modeled the event as 'six stages of doom', each with a 50% chance of occurring; however, conditional on Trump besting the first 5 stages, was it really right to model the sixth as a coinflip?[175] Probably not.[176]

We believe we have not committed either error.

Regarding failing to model parallel disjunctive paths:

- We have chosen generic steps that don't make rigid assumptions about the particular algorithms, requirements, or timelines of AGI technology
- The one opinionated claim we do make is that transformative AGI by 2043 will almost certainly be run on semiconductors powered by electricity, and we spend many pages justifying this belief

Regarding failing to really grapple with conditional probabilities:

---

174 Technology Forecasting: The Garden of Forking Paths · Gwern.net
175 Donald Trump's Six Stages Of Doom | FiveThirtyEight
176 How I Acted Like A Pundit And Screwed Up On Donald Trump | FiveThirtyEight

- Our conditional probabilities are, in some cases, quite different from our unconditional probabilities. In particular, we assume that a world on track to transformative AGI will…
    - Construct semiconductor fabs and power plants at a far faster pace than today (our unconditional probability is *substantially* lower)
    - Have invented very cheap and efficient chips by today's standards (our unconditional probability is *substantially* lower)
    - Have lower risks of disruption by natural pandemic
    - Have *higher* risks of disruption by regulation
    - Have *higher* risks of disruption by engineered pandemic
    - Have *higher* risks of disruption by war

To help convince ourselves, and you, that our framework is valid, we can use the approach to estimate the probability of transformative AGI emerging in the longer term. What might we say about the probability of a cascade through all seven of our key events by 2100, for example?

The probabilities of all the hardware and software R&D factors would go up substantially. With extra decades to work, we would have a much higher probability of inventing core AGI algorithms and suitable robot bodies, and the probably of two, three, or four orders of magnitude in transistor improvements would go up. The possibility of a non-silicon computing substrate could become substantial, not farfetched.

Training by non-sequential reinforcement learning ceases to be an issue. If AGI was invented in 2060 or even 2070, it could easily be trained for nearly every relevant task by 2100.

Hardware logistics factors could essentially be done away with. With nearly eighty years until 2100, the timeline is long enough that the ability to scale manufacturing of semiconductors, robots and power plants can be taken for granted; imagine trying to argue in 1880 that scaling car manufacturing would be too hard to put one in every home, or in 1920 that the logistics of airliner production would not be solved in the year 2000, or in 1940 that transistors, once invented, would be too hard to scale to put a computer in every household.

Most sociopolitical factors could be done away with, at least as regards delay. There is enough time before 2100 for AGI development to be arrested for decades by a war, pandemic or depression without preventing transformative AGI from emerging prior to 2100. Only if these things result in a durable inability to produce, or lack of interest in, transformative AGI will they matter on this timescale.

Below we present what we think is a reasonable guess of how our framework predicts the probability of transformative AGI by 2100, and reach a number of about 41%.

| Event | Forecast by 2043 conditional on prior steps | Forecast by 2100 conditional on prior steps |
|---|---|---|
| We invent algorithms for transformative AGI | 60% | 85% |
| We invent a way for AGIs to learn faster than humans | 40% | N/A |
| AGI inference costs drop below $25/hr (per human equivalent) | 16% | 75% |
| We invent and scale cheap, quality robots | 60% | 90% |
| We massively scale production of chips and power | 46% | 99% |
| We avoid derailment by human regulation | 70% | 90% |
| We avoid derailment by AI-caused delay | 90% | 95% |
| We avoid derailment from wars (e.g., China invades Taiwan) | 70% | 90% |
| We avoid derailment from pandemics | 90% | 95% |
| We avoid derailment from severe depressions | 95% | 98% |
| Joint odds | **0.4%** | **41%** |

We hope this will indicate that the framework is indeed capable of producing a high number, and is not a rhetorical mugging. The reason we produce a low number is that the timeline before 2043 is tight, not that we have adopted an inherently pessimistic conceptual framework.

Please try it yourself.

## Why AI likely won't clear the barriers to transformative AGI

If you mostly accept our framework, then the only way to end up with a high probability of transformative AGI by 2043 is to believe that most of the factors are close to 100%.

Our own factor estimates rely heavily on projections of current rates of technological progress and manufacturing.

However, it's worth considering the likelihood that early, pre-transformative AI or AGI accelerates technological progress and manufacturing to unprecedented rates, and ushers in transformative AGI much sooner than expected by our backwards-looking base rate analysis. This scenario of auto-catalyzing growth is sometimes called hard or fast takeoff.

While we acknowledge the possibility, we consider it unlikely for a number of reasons.

The main evidence that persuades us is the lack of historical precedent.

New general purpose technologies, even those powerful enough to spark industrial revolutions, have always started out crummy and expensive. It takes time to invent the technology. Time to improve the technology. Time for its supply chains to reconfigure to efficiently produce and distribute the technology. Time for other supply chains to incorporate it as an input. Time for humans to reorient businesses and organizations around it. Time for end users to figure out how to use it. Time for investors to gain confidence and scale investments in it. Time for the educational system to train workers more proficient in it. And technological value is always diminishing in the sense that the more efficient you make a segment of the value chain, the less valuable further improvements become, as other segments become the high-value bottlenecks.

Looking at history:

- Steam engines made it cheaper to mine coal to power steam engines, and to build new steam engines, but still UK per capita economic growth never exceeded 1%/yr in the 1700s.

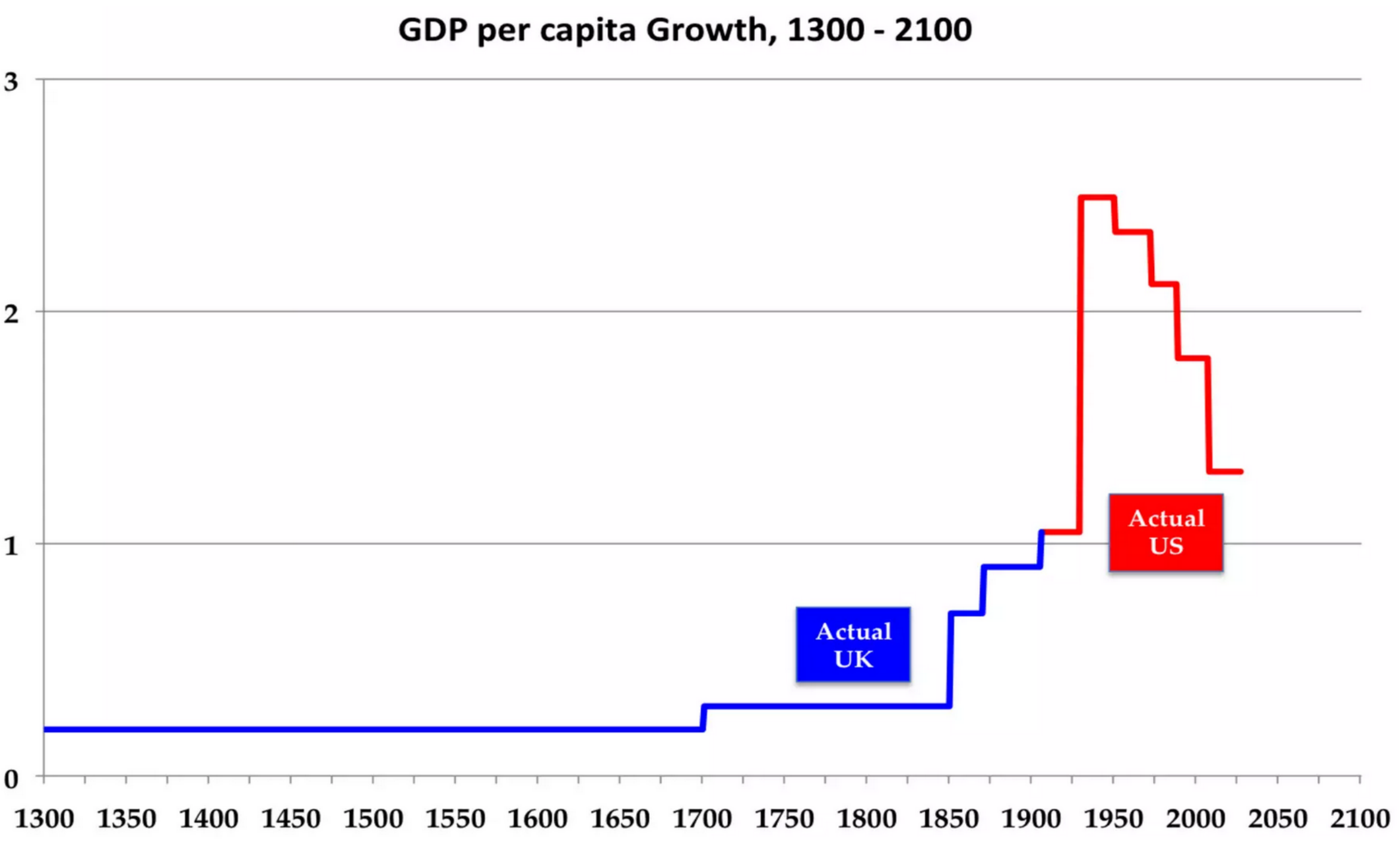

*Source: Robert Gordon*[177]

[177] Is U. S. Economic Growth Over? Lessons from the Long 20th Century, Robert Gordon

- Electricity made it cheaper to manufacture inputs to electrical generation and distribution, yet prices and quantities of electricity moved gradually over decades. (Tim Harford has some excellent writing [here](#) on why it took so long for electrification to affect manufacturing.[178] He highlights how factories first used electricity to simply replace steam-powered mechanical drives, before learning how to take advantage of the efficiencies that came from redesigning manufacturing processes around it.)
- Computer chips allowed us to write programs to help design future computer chips, but chip design efficiencies never skyrocketed discontinuously.

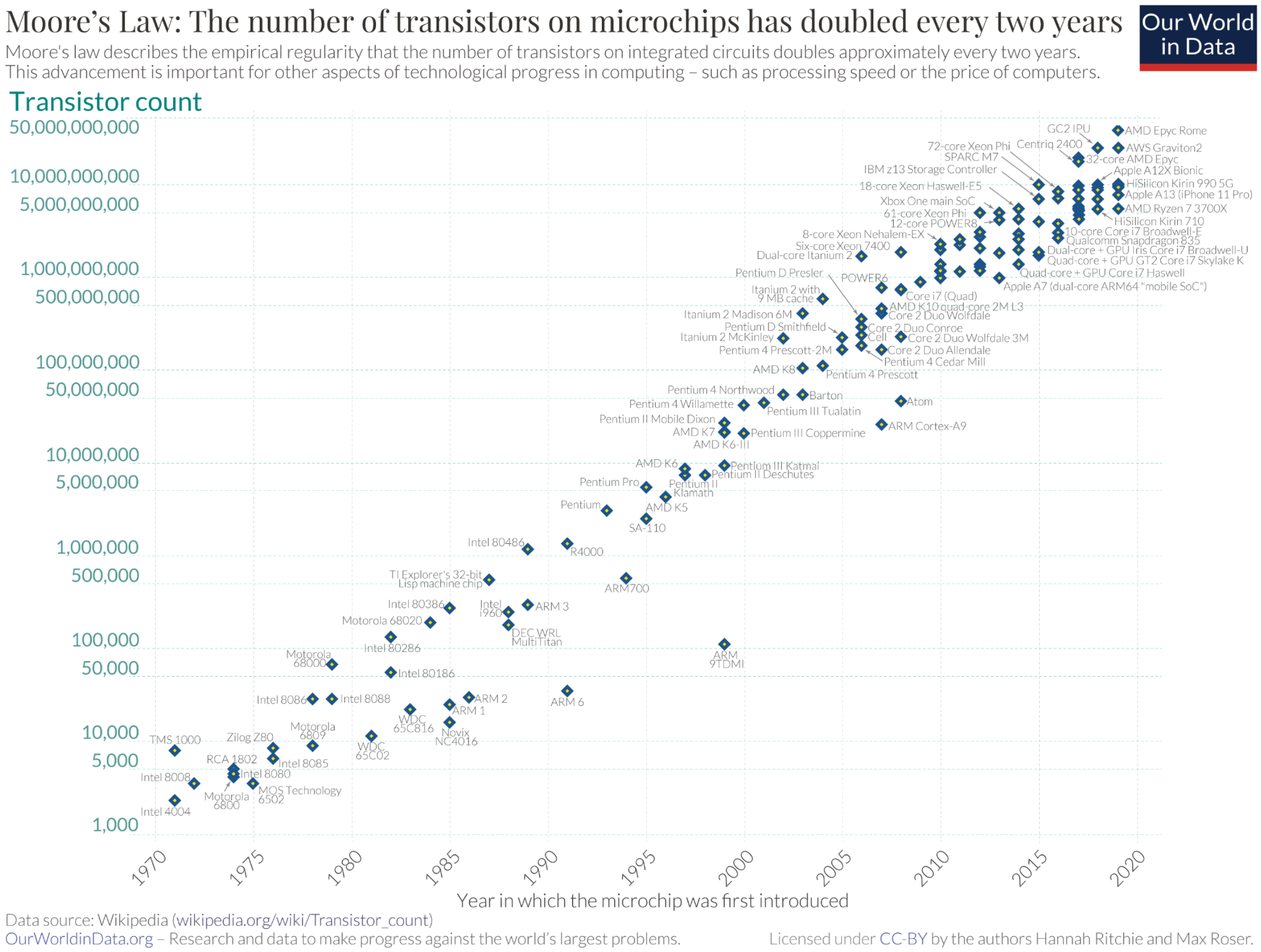

*Moore's Law shows no discontinuity around the invention of [Verilog](#) or other computer-powered chip design tools. Source: Our World in Data*[179]

[178] [Why didn't electricity immediately change manufacturing? - BBC News](#)
[179] [What is Moore's Law? - Our World in Data](#)

- Printed books spread knowledge that made it easier to write and print more books, but economic growth did not discontinuously explode, nor, apparently, did book publishing itself:

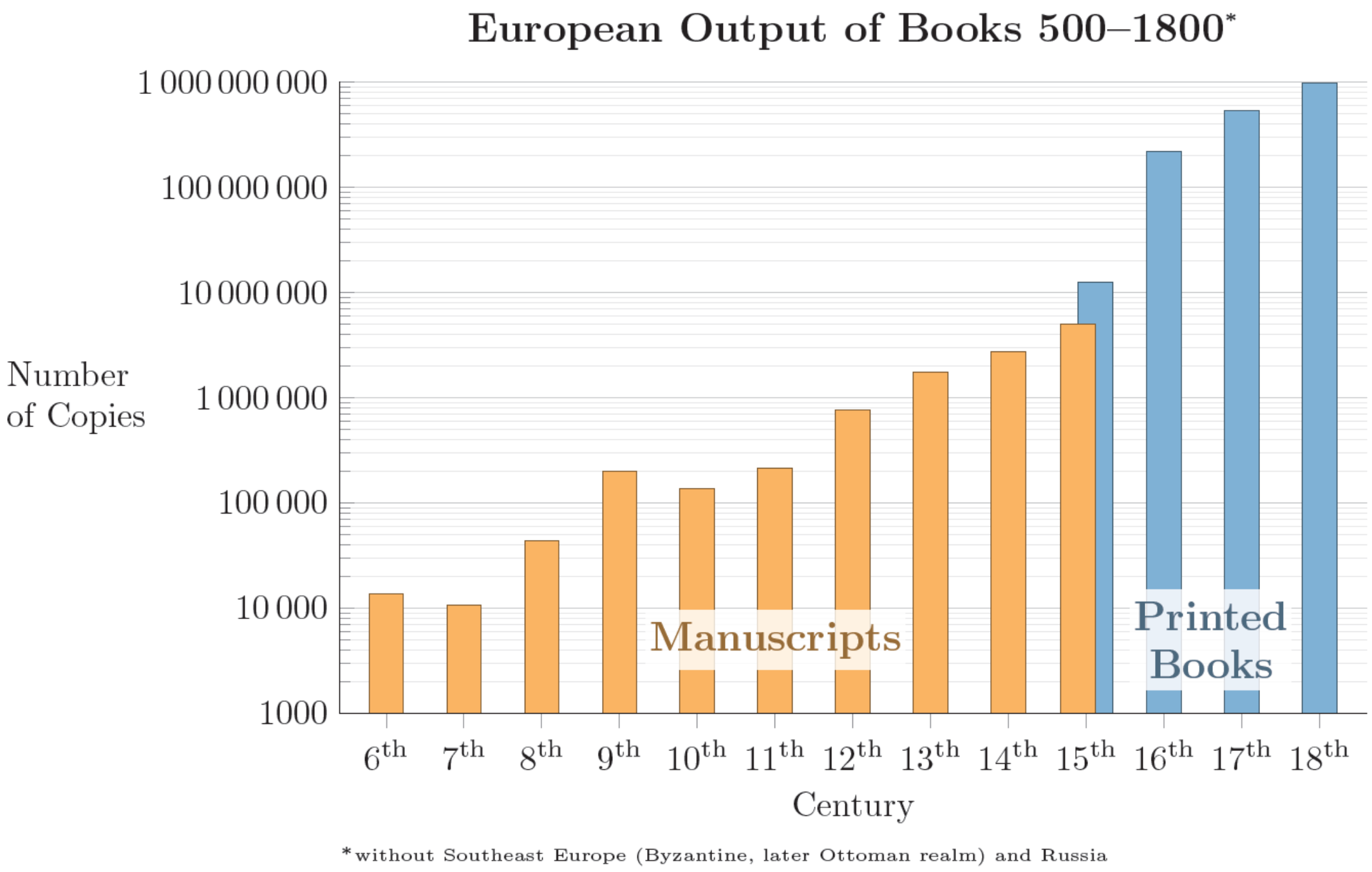

*Source: Wikipedia*[180]

- Automobiles allowed us to ship automobile parts around more cheaply, but economic growth did not explode.
- Better agricultural yields supported larger populations, which had greater ability to figure out how to improve agricultural yields, but economic growth did not explode.

With each of these technologies, there was positive feedback, but the feedback was subcritical and it took decades or more for their impacts to be fully realized.

We expect AGI takeoff to be similarly gradual, for a constellation of reasons:

- We expect early AGIs to be extremely expensive and frustratingly unreliable, in ways that impede fast takeoff.
- We expect companies to take years to figure out how to take full advantage of improving AI technology. E.g., what is it best at? What is it worst at? What should the process be to assess its growing capabilities? To what extent should the AI be supervised? Should it amplify human effort or replace human effort? How should it change the workforce they hire? How should they retrain their current workforce to use these tools? How should they retrain or fine-tune the AI to be most effective at their use cases? What types of data will be the most valuable for training? Answering these questions well will take many years. And large companies have tremendous inertia. The business case for videoconferencing and remote work in 2023 is largely the same as 2019, but it took a global pandemic for companies to reassess business as usual. It may similarly take companies years to fully integrate AI, even when the fruits are ripe, and theoretically ready to be picked.

[180] Copyright - Wikipedia

- Technological development has *already* been accelerated by superhuman computer technologies. E.g., chip design and manufacturing rely tremendously on computers for calculations and optimizations and automated design. The lack of takeoff caused by computer and narrow AI tools gives us confidence that improvements in narrow AI will not soon lead to takeoff. It could well be that a human-level AGI still chooses to use human-written, battle-tested design tools that accomplish tasks more efficiently than directly applying its brute-force intelligence to the task.
- Even smart-human-level, smart-human-cost AGI will not necessarily accelerate technological progress. There are many millions of smart humans on Earth and only a tiny fraction of them work in AGI R&D or semiconductor R&D. The limiting factor in 2023 is willingness to spend, not the supply of human brains and bodies. (Of course, superhuman AGI or below-human-cost AGI *could* wildly accelerate R&D, but at that point transformative AGI would have already arrived. Here we are explicitly considering the narrow question of how *pre*-transformative AGI might accelerate transformative AGI.)
- Also, as we highlighted previously, active training loops can be slow. It may be that AGI self-accelerates in easily simulated domains (e.g., chess, Go, debate) but does not self-accelerate as much at skills or tasks where real-world interaction costs bottleneck the rate at which crucial training data can be acquired. It's possible that self-acceleration is not a binary yes or no, but an uneven heterogeneous landscape over different types of tasks and skills. Remember, transformative AGI requires that *nearly all* valuable human work can be done cheaply by AGI. Even half would not count.
- Lastly, all technologies are limited by physics. Even an AGI that figures out how to build amazing digital metaverses or cancer-curing drugs will still face the same physical constraints when it comes to physical, capital-intensive industries like semiconductor fabrication, or electricity production. Even if pre-transformative AGIs drive the price of less-physically constrained goods like entertainment and medicine to zero, physical technologies and investments will still take time, no matter how smart the entity directing them is.

In sum, despite thinking it's eminently plausible for cheap, superhuman AGI to self-accelerate, we are far more pessimistic about the prospects of expensive or near-human AGI self-accelerating. We noted that R&D is not bottlenecked by brains, and we noted that many self-accelerating technologies from history took decades for their impact to be fully felt.

Every successful technology starts out crummy and uneven and expensive, with users and buyers who at first don't understand how to maximize its benefits. We expect early AGI to be no different.

# Conclusion

If you’ve read the entire essay up to this point, we’ve taken a very long journey together. And it’s led us into a lot of intellectual nooks and crannies that are pretty far from what we usually think about when discussing AGI, including the philosophy of probability, the history of the development of many different technologies, the likely durability of robot bodies, the complexity of the synaptic proteome, nerve density in the spinal cord, simulating neurons, nematode worms, silicon lithography tooling, chess theory, self driving cars, and exotic subjects like engineered pandemics and the game theory of the Taiwan crisis.

But we hope that the main point has come through clearly: Transformative AGI is a very high bar we’re currently very far from clearing. It will require huge transformations in complicated, capital-intensive sectors of society that are slow to change.

When we think clearly about the many things that would all have to happen for this transformation to happen on a short timeline like 20 years, and make candid efforts to assign probabilities to them, the cascade of probabilities through all the different necessary events makes it hard to justify a conclusion that the probability of transformative AGI by 2043 is even 3%, let alone 10%. Although transformative AGI could in theory happen before 2043, and some of our factors could be wrong, and some of our assumptions could be wrong, the largeness of the margin makes the broader argument robust against this kind of error.

We started this project in broad agreement with you, and with a suspicion that when we reckoned the cascade, the value would be below 10%, but we were surprised to see a small fractional percentage. That specific quantitative result depends critically on a lot of rather new information, including the critical result of Beniaguev et al. about the computational complexity of neurons, which had been suspected by computational neurologists for a while, but not experimentally confirmed until just last year. But the broad pattern result, $p<3\%$, is a pretty clear consequence of our shared way of thinking about this problem, once systematized in this way.

If you’re convinced of this, it’s mostly good news, we think. Nothing about this result implies that the happy potential of AGI to radically transform human life for the better, and add new forms of life to the universe, will remain unfulfilled forever, or for the lifetimes of those alive today. Nor does it imply that 2043 will not be an age of wonders, or that work in AGI, whether in development or alignment research, is fruitless or vain. It couldn’t be more critical.

But it does mean that we have a fair degree of certainty that the fundamental problems around alignment and X-risk are not going to press down on us within the next twenty years; we have time for this critical work to proceed.

It also means that to the extent any form of AI alignment research depends on identifying and halting dangerous AGI work, rather than rendering it fundamentally impossible or disincentivized, the capital-intensive, highly visible events we contemplate here would be a good way to think about the task of identifying dangerous AGI work.

We hope and suspect that when we revisit this essay in 2043, we will not look like fools. But we would be happy to look like fools in service to the overall work of AI alignment research, of which this inquiry is a small part: making sure that humanity as a whole will have the chance to look back on its past at all.

# End matter

## Appendix: Why reading this essay should extremize your confidence

One funny fact of Bayesian belief updating is that you should expect to get more confident after reading our essay, *even before you read its content*. There are two related reasons why:

- Forecasts should get more confident with more information
- Forecasts should get more confident with more diversity of information

### Forecasts should get more confident with more information

In a theoretical world in which we're unbiased Bayesian updaters, new information tends to increase confidence (e.g., bring beliefs closer to 0% or 100%).

To illustrate the point, consider the most extreme case of information gathering: a fortune-telling oracle with perfect accuracy who you trust fully. One day you ask the oracle: "will transformative AGI occur within 20 years?" Even before hearing her answer, you can already know something about how your belief should change. If, walking in, you believed the odds were ~20%, then you should expect to walk away from the oracle with a 0% belief ~80% of the time and a 100% belief ~20% of the time.

Although it's a bit more complicated to write down, the same logic applies to imperfect sources of information as well. Even though new information shouldn't change the *expected value* of your belief, you should expect new information to move you away from your starting point and toward the poles of 0% and 100%.

Therefore, if this essay contains information or perspectives that are new to you, then you should expect that in most cases your confidence will be pulled closer to 0% (counterbalanced in expected value by rarer cases where you make larger jumps toward 100%).

### Forecasts should get more confident with more diversity of information

Philip Tetlock writes that one of the four performance drivers in forecasting tournaments are algorithms to (a) overweight better forecasters and (b) *extremize* their forecasts.[181]

He illustrates the idea with a high-stakes example: US President Obama deciding whether to attack terrorist Osama Bin Laden.[182] Suppose the President goes around his room of 10 advisors and asks each for their forecast of success. Each advisor says, "Mr. President, I think there's a 70% probability of success."

Given this distribution of forecasts, what should the President conclude?

It turns out the answer depends on the diversity of the advisors.

---

[181] Edge Master Class 2015: A Short Course in Superforecasting, Class II
[182] Philip Tetlock on Superforecasting - Econlib

If the 10 advisors are all clones of one another, all working from the same pool of known information, then it makes no difference whether the President hears 70% from one advisor or from 10. The extra advisors add nothing new and the President's best guess should be 70%.

However, if  the 10 advisors are each drawing on different pools of information—e.g., one advisor has satellite data and one advisor is a codebreaker and one advisor has human intelligence signals—then hearing that each is 70% confident should result in an aggregate forecast much higher than 70%. The positive evidence from satellites and codebreaking and human intelligence are not duplicative—they're additive. If the President learns that every perspective and information source is saying that success is likely, then by combining those perspectives he can be far more confident than any single source alone.

When Tetlock applied this idea to groups in the IARPA forecasting tournament, he found that large levels of extremizing improved accuracy. E.g., if a group of diverse forecasters thought something was 30% likely, an accurate extremizing algorithm would adjust the aggregate estimate down to 15% or 10%. This is not a small effect!

This notion of extremizing forecasts that aggregate diverse information has substantial implications for how you should expect your probabilities to move after reading our essay and others. ***Even in a world where you receive 10 essays all agreeing completely that the odds of transformative AGI in 20 years are ~20%, you should still be inclined to <u>reduce your estimate</u>***, assuming those 10 essays draw from different sources and different perspectives, which they will undoubtedly will. (One key assumption here is that you consider others' views as valid as your own; if you heavily weight your own beliefs, then others may not move you much.) Using the ratios estimated by Tetlock in his IARPA forecasting tournaments, this extremizing alone might even mean dropping an estimate from ~20% down to 10% or 7%. And of course, if these essays, like ours, argue for much lower probabilities, then it's very conceivable to end up with a final estimate of 5% or lower, even without any seismic paradigm shifts to your inside views.

## Appendix: Why 0.4% may be less confident than it seems

From one point of view, 0.4% may *feel* like a very confident forecast. Given our guiding principle of avoiding high confidence, wouldn't a conservative, more open-minded forecast start closer to the 20%–80% range, leaving a healthy margin for error?

The neutral 'starting point' for a forecast is subjective, and depends on how you frame the problem.

Consider being asked to forecast the odds that a *weighted* 20-sided die lands on 20. Because the die is weighted in some fashion unknown to you, it's hard to know what to guess.

From one point of view, there are two outcomes—either you roll a 20 or you don't—so you might say a neutral starting guess is 50%.

But from another point of view, there are 20 possible outcomes—one for each face of the weighted die—so you might say a neutral starting guess is just 5%.

Which point of view is correct? In the case of a 20-sided die, we think it feels more 'right' for a neutral weighting to split probability among the 20 faces, not the two outcomes of rolling a 20 or not. So we'd guess 5%, acknowledging there's a good chance we're way off.

In the case of the 20-sided die, we suspect you'd probably agree that 5% is a better starting point than 50%. However, in scenarios more ambiguous than a 20-sided die, the question of the 'right' starting guess remains quite subjective.

Personally, we think transformative AGI is more like a 20-sided die than a coinflip. Although one possible partitioning of outcomes is transformative AGI vs not (a fifty-fifty proposition), you can imagine far more detailed partitions, such as:

- Non-transformative AGI scenarios (20)
  - Little progress is made toward AGI, and AGI is not invented
  - Large progress is made toward AGI, but AGI is still not invented
  - Something that looks like AGI is invented but it's too brittle to reliably replace human workers
  - AGI is invented, but AGI inference is too expensive to replace most human work
  - Cheap AGI is invented, but happens so close to 2043 that there's not enough time for a trillion-dollar effort to scale up semiconductor production to a level where AGIs can replace most human work
  - Cheap AGI is invented, and there's theoretically enough time to scale up semiconductor production, but practical difficulties of complex semiconductor supply chains result in shortages, and there's not enough compute to replace most human work
  - Cheap AGI is invented, but China invades Taiwan and global semiconductor production collapses, thereby delaying the scaling of transformative AGI beyond 2042
  - Cheap AGI is invented and scaled, but not robot bodies are never made cheap and reliable, meaning that AGI's *direct* influence is limited to the world of information, and many humans end up continuing to work in normal-ish jobs that involve movement of matter

- AGI is invented, but foreign powers are worried enough about the prospect that they launch a physical or cyber war against the creators of AGI, slowing its development and deployment past 2042, and dissuading other would-be creators of AGI
  - AGI is invented, but its creators fear misalignment so much that they purposely halt development, delaying transformative impact beyond 2042
  - Cheap pre-AGI is invented, and the algorithms are immediately stolen and distributed across the world. The ineffectiveness of computer security dissuades profit-seeking parties from investing hundreds of billions in giant training runs, resulting in transformative AGI pushed back beyond 2042
  - Humanity is on track to develop transformative AGI but before it does, a severe pandemic far worse than COVID delays progress by killing off key researchers, snarling semiconductor supply chains, reducing productivity, and causing a global depression
  - Humanity is on track to develop transformative AGI but before it does, World War III erupts, which kills off key researchers, snarls semiconductor supply chains, and reprioritizes resources to the war effort and to post-war rebuilding
  - Humanity is on track to develop transformative AGI but concerns about alignment or mass unemployment or power imbalances between nations and corporations causes governments to stifle commercial application, as they've done successfully with nuclear proliferation and human cloning; this slows transformative AGI to beyond 2042
  - Humanity develops aligned superintelligent AGI, and it warns us to shut it down and cease future development, because the risk of unaligned AGI is too high. We temporarily shut down the major efforts, delaying the deployment of transformative AGI
  - Cheap AGI is invented and scaled, but it turns not to have many great ideas beyond the ones we have ourselves, and while it can replace expensive knowledge workers, most humans in <$25/hr jobs lead roughly similar lives after its development
  - Narrow AI accelerates progress in biotechnology, making it easier and easier to manipulate viruses, until one bad actor purposefully releases an engineered pandemic far far worse than COVID19, which kills millions, shuts down world trade, and leads to a decade-long depression
  - Narrow AI contributes to the development of increasingly entertaining and increasingly cheap information goods that reduce human productivity and fertility, which slow AGI development, slow the growth of capital available to invest in AGI development, and slow the expected demand for AGI-produced goods and services, all of which together combine to delay development of transformative AGI beyond 2042
  - Proto-AGI is invented, but the pace of trajectory is so rapid that no one wants to invest $1T in training a full AGI, because if they wait a year it might cost half as much or be twice as smart; this compounds for a few years, and big AGI training runs start too late to seed transformative AGI by 2043
  - AGI is invented, but it begins making scientific progress in computer hardware so quickly that everyone agrees it's foolish to spend trillions of dollars manufacturing expensive silicon wafers and gigawatts of new power plants, when we can instead spend 5-10 extra years scaling up cheap, power efficient DNA computer chips; unfortunately the delay caused by switching to a superior computing substrate pushes transformative AGI out beyond 2042
- Transformative AGI scenarios (2)
  - Humans invent and scale cheap AGI and robot bodies, leading to transformation of human life and work around the planet
  - Humans invent AI technology that amplifies research productivity, leading to AGI, which in turn figures out how to make itself more efficient, build robot bodies, and scale both to a planet-scale population

This subjective partitioning of outcomes, with 2 AGI scenarios and 20 non-AGI scenarios, might suggest a ‘neutral’ starting guess of 2/22 = 9% likelihood of transformative AGI. Further detailed analysis might then adjust this up or down.

Of course, had we chosen to list out 2 non-transformative AGI scenarios alongside 20 transformative AGI scenarios, we might equally argue that 91% is a decent ‘neutral’ starting point.

The point of this digression is not to say that any particular forecast is a good neutral starting point, but to challenge any latent assumptions that 0.4% is necessarily more confident or more extreme than ~50% or ~20%. In a hypothetical world where AGI is just one side of a 33-sided die, 3% would be a ‘conservative’ answer and 20% would be an ‘opinionated’ answer deserving extra scrutiny.

From our point of view, considering all the diverse and detailed ways that the future might unfurl over the next 20 years, reserving an entire 20/100ths of them to the outcome of transformative AGI is itself a fairly extreme claim. Out of 100 possible rolls of the future, it feels rashly confident to predict that 20 of them will *all* contain transformative AGI scenarios. There are so many other non-transformative AGI scenarios to allocate probability to as well.

This core idea is what’s embodied in our 7-factor analysis. Namely,

- There are many more paths that fail to achieve transformative AGI than paths that achieve transformative AGI
- Coupled with our principle of humility that we shouldn’t be too confident in any one of these paths, which implies the odds of transformative AGI are likely low

# Appendix: What narrow AI tells us about AGI

To what extent can humans forecast the impacts of superintelligent AGI?

**From one point of view, trying to understand superintelligence seems utterly intractable.** Just as a dog or chimpanzee has little hope of comprehending the motivations and powers of humans, why should humans have any hope of comprehending the motivations and powers of superintelligence?

**But from another point of view, forecasting the impacts of superintelligence may yet be possible.** The laws of reality that constrain us will similarly constrain any superintelligence. Even if a superintelligence achieves a more refined understanding of physics than us humans, it very likely won't overturn laws already known.[183] Thus, any inventions optimized against those physical laws, even if superior to our own, may end up looking familiar rather than alien.

**No matter how intelligent an AGI is, it will still be bound by physics.** No matter how smart you are, you still must obey the law of conservation of energy.[184] Just like us, an AGI wishing to affect the world will require an energy industry full of equipment to extract energy from natural sources. Just like us, its energy will have to come from somewhere, whether it's the sun (solar, wind, biofuels, fossil fuels, hydro), the Earth (geothermal, nuclear), or the Moon (tidal, nuclear).[185] Just like us, any heat engines will be limited by Carnot efficiency. Just like us, energy will need to be transported from where it is collected to where it is consumed, likely by electromagnetic energy carried by bound electrons (e.g., chemical fuels) or unbound electrons (e.g., electricity). If there are economies of scale, as there likely will be, that transportation will take place across networks with fractal network topologies, similar to our electric grids, roads, and pipelines. The physics of energy production are so constrained and so well understood that no matter what a superintelligence might build (even fusion electricity, or superconducting power lines, or wireless power), we suspect it will be something that humans had at least considered, even if our attempts were not as successful.

One way to preview superintelligent AGI is to consider the superintelligent narrow AIs humanity has attempted to develop, such as chess AI or self-driving AI.

## Lessons from chess AI: superintelligence is not omnipotence

In 2017, DeepMind revealed AlphaZero. In less than 24 hours of (highly parallelized) training, it was able crush Stockfish, the reigning AI world chess champion. AlphaZero was trained entirely de novo, with no learning from human games and no human tuning of chess-specific parameters.[186]

**AlphaZero is *superhuman* at chess.** AlphaZero is so good at chess that it could defeat all of us combined with ease. Though the experiment has never been done, were we to assemble all the world's chess grandmasters and give them the collective task of coming up with a single move a day to play

---

[183] The Relativity of Wrong by Isaac Asimov
[184] At least on human-relevant scales.
[185] Or space, if you want to, say, mine He3 for fusion fuels.
[186] Though there was tuning of non-chess-specific hyperparameters

against AlphaZero, we'd bet our life savings that AlphaZero would win 100 games before the humans won 1.[187]

From this point of view, AlphaZero is godlike.

- Its margin of strength over us is so great that even if the entire world teamed up, **it could defeat all of us combined** with ease
- It plays moves so subtle and counterintuitive that they are **beyond the comprehension of the world's smartest humans** (or at least beyond the tautological comprehension of 'I guess it wins because the computer says it wins').[188]

But on the other hand, pay attention to all the things that *didn't* happen:

- **AlphaZero's play mostly aligned with human theory[189]—it didn't discover any secret winning shortcuts or counterintuitive openings**.
    - AlphaZero rediscovered openings commonly played by humans for hundreds of years:

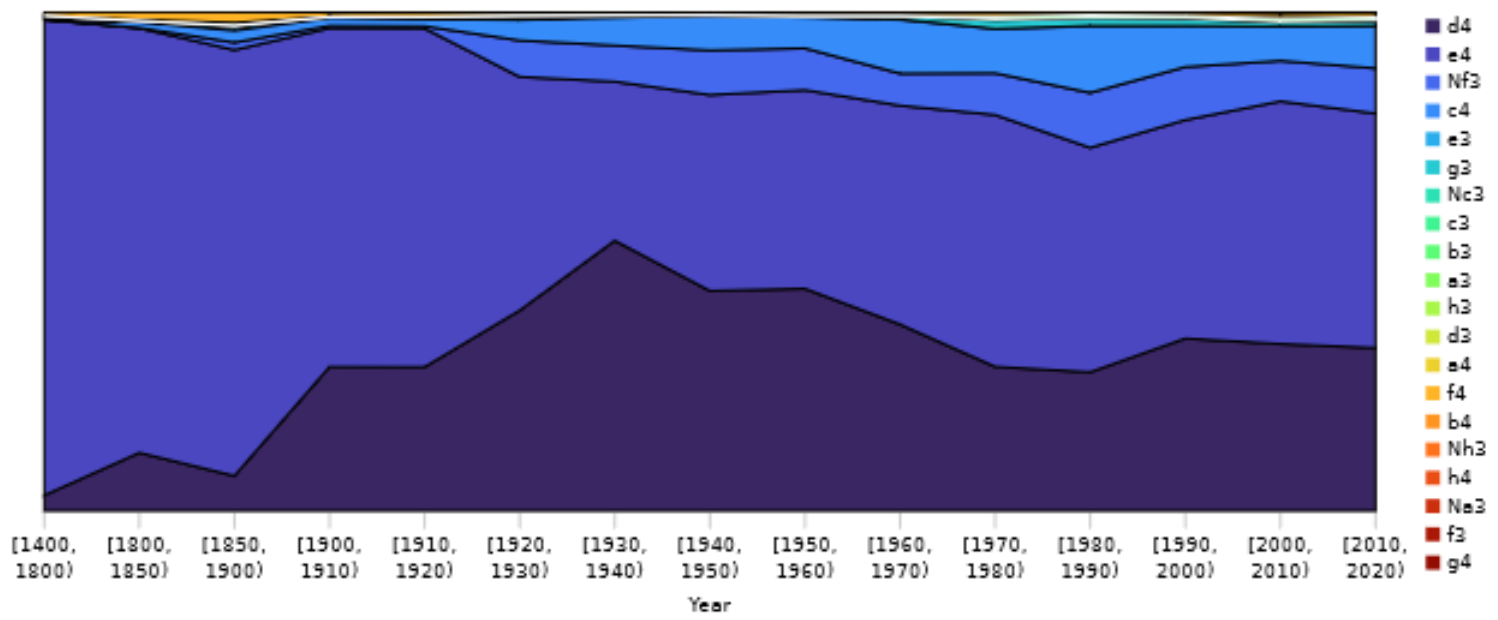

**(a)** The evolution of the first move preference for White over the course of human history, spanning back to the earliest recorded games of modern chess in the Chessbase database. The early popularity of 1. e4 gives way to a more balanced exploration of different opening systems and an increasing adoption of more flexible systems in modern times.

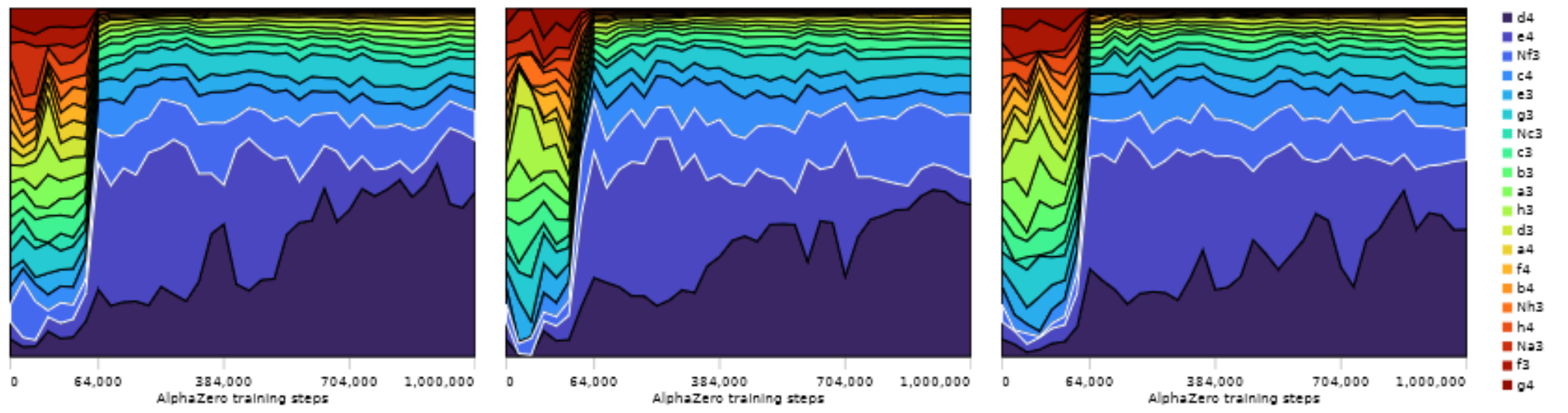

**(b)** The AlphaZero policy head's preferences of opening move, as a function of training steps. Here AlphaZero was trained three times from three different random seeds. AlphaZero's opening evolution starts by weighing all moves equally, no matter how bad, and then narrows down options. It stands in contrast with the progression of human knowledge, which gradually expanded from 1. e4.

        - **Figure 5.** A comparison between AlphaZero's and human first-move preferences over training steps and time.
    - When DeepMind looked inside AlphaZero's neural network they found "many human concepts" in which it appeared the neural network computed quantities akin to what

---

[187] Assuming humans don't implicitly use computer assistance by, for example, using 90 games as a budget to play AlphaZero against itself for 45 moves. (This assumes AlphaZero would play deterministically. See: Is AlphaZero deterministic? - Chess Stack Exchange)

[188] Here's a great example, though it relies on tablebases rather than AlphaZero: Just one of 17,823,400,766 positions | ChessBase

[189] [2111.09259] Acquisition of Chess Knowledge in AlphaZero

humans typically compute, such as material imbalance (alongside many more incomprehensible quantities, to be fair).

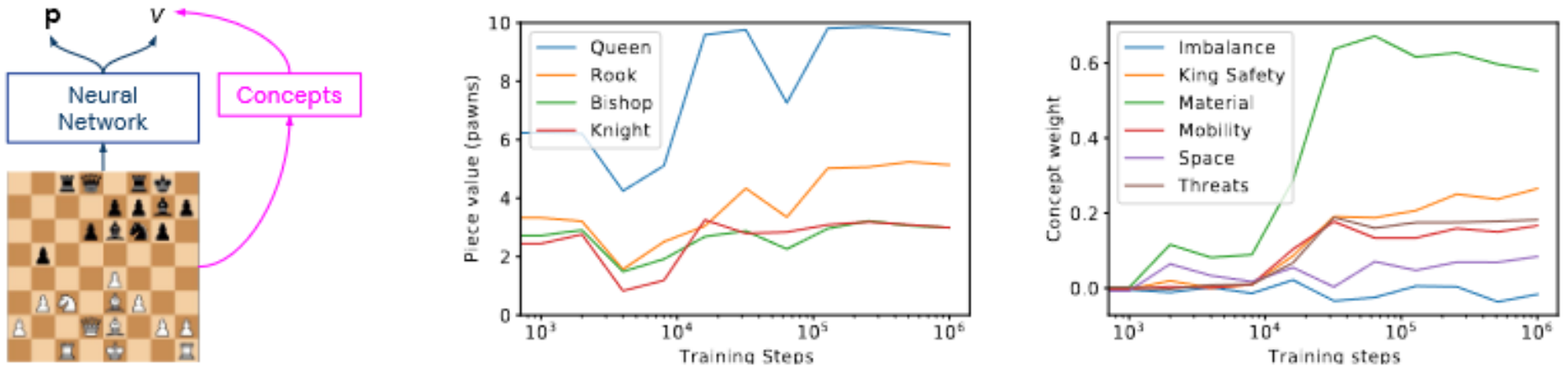

**(a)** Value regression methodology: we train a generalized linear model on concepts to predict AlphaZero's value head for each neural network checkpoint.

**(b)** Piece value weights converge to values close to those predicted by conventional theory.

**(c)** Material predicts value early in training, with more subtle concepts such as mobility and king safety emerging later.

**Figure 4.** Value regression from human-defined concepts over time.

- **AlphaZero was comprehensible**—for the most part, AlphaZero's moves are comprehensible to experts. Its incomprehensible moves are rare, and even then, many of them become comprehensible after the expert plays out a few variations. (By comprehensible, we don't mean in the sense that an expert can say why move A was preferred to move B, but in the sense that an expert can articulate pros and cons that explain why move A is a top candidate.)
- **AlphaZero was not invincible**—in 1200 games against Stockfish, it lost 5/600 with white and 19/600 with black.[190]
- **AlphaZero was only superhuman in symmetric scenarios**—although AlphaZero can reliably crush grandmasters in a fair fight, what about unfair fights? AlphaZero's chess strength is estimated to be ~3500 Elo. A pawn is worth ~200 Elo points as estimated by Larry Kaufman,[191] which implies that a chess grandmaster rated ~2500 Elo should reliably crush AlphaZero if AlphaZero starts without its queen (as that should reduce AlphaZero's effective strength to ~1500 Elo). Even superintelligence beyond all human ability is not enough to overcome asymmetric disadvantage, such as a missing queen. *Superintelligence is not omnipotence.*
- **AlphaZero hasn't revolutionized human chess**—although players have credited AlphaZero with new inspiration,[192] for the most part human chess is played at a similar strength and style. AlphaZero didn't teach us any secret shortcuts to winning, or any special attacks that cannot be defended, or any special defenses that cannot be pierced. Arguably the biggest learnings were to push the h pawn a little more frequently and to be a little less afraid of sacrificing pawns to restrict an opponent's mobility. Even chess computers at large haven't dramatically transformed chess, although they are now indispensable study tools, especially for opening preparation. Historical rates of human chess progress as measured by average centipawn loss (ACPL) in world championship (WC) games show no transformative step function occurring in the 2000s or 2010s.[193] Progress may have accelerated a little if you squint,[194] but it wasn't a radical departure from pre-AI trends.

[190] [1712.01815] Mastering Chess and Shogi by Self-Play with a General Reinforcement Learning Algorithm
[191] The Evaluation of Material Imbalances (by IM Larry Kaufman) - Chess.com
[192] https://twitter.com/olimpiuurcan/status/1139437778683322369
[193] Exact, Exacting: Who is the Most Accurate World Champion? | Blog • lichess.org
[194] And ignore confounders, like the effect of being Magnus Carlsen.

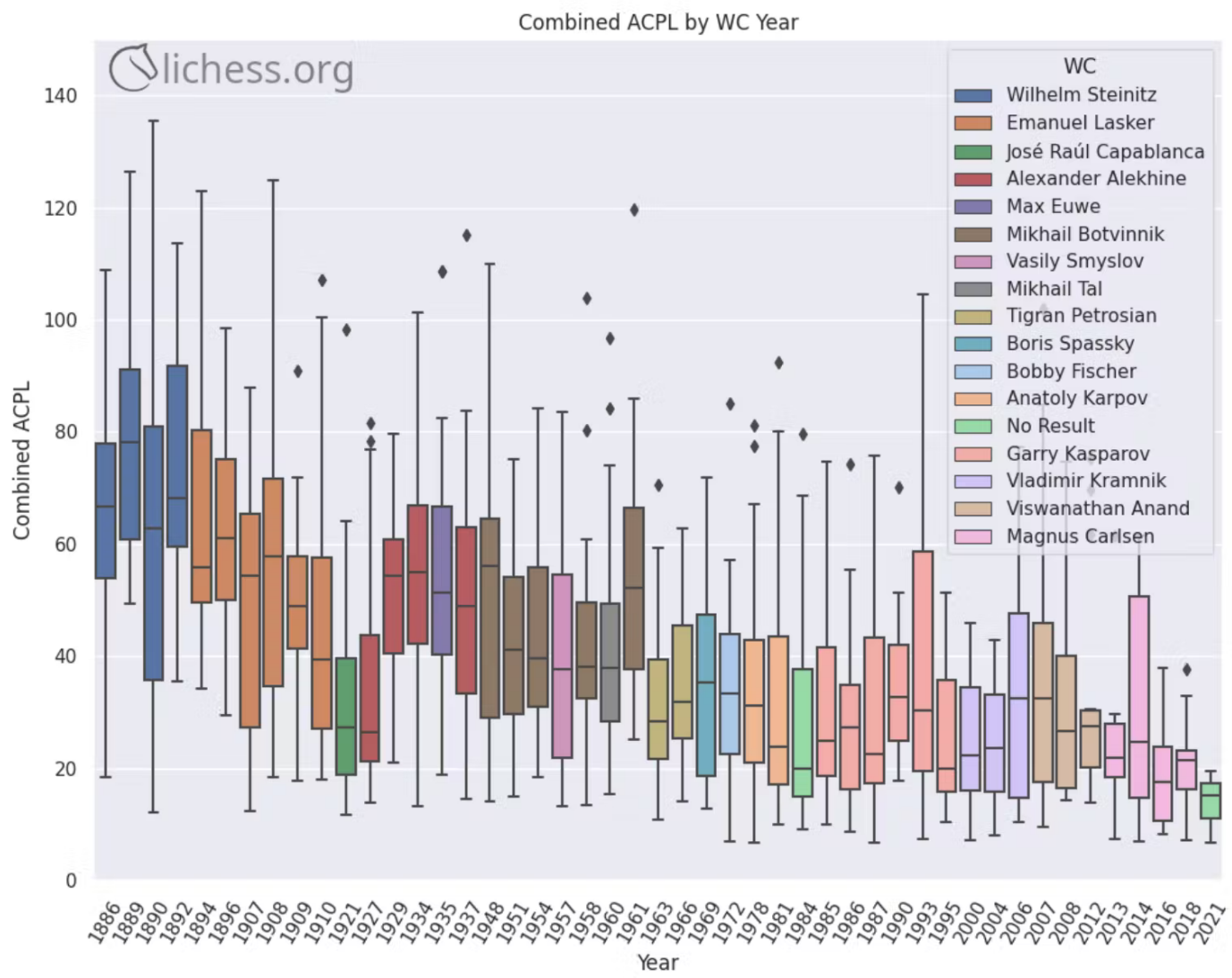

Ultimately, superhuman AIs didn't crush humans at chess by discovering counterintuitive opening secrets or shortcuts that dumb humans missed. In fact, AIs rediscovered many of the same openings that humans play. Rather, AIs do what humans do—control the center, capture pieces, safeguard their king, attack weaknesses, gain space, restrict their opponent's options—but the AIs do this more effectively and more consistently.

In the 5th century BCE, Greek philosophers thought the Earth was a sphere. Although they were eventually improved upon by Newton, who modeled the Earth as an ellipsoid, they still had arrived at roughly the right concept. And in chess we see the same thing: although a superhuman AI 'understands' chess better than human experts, its understanding still mostly reflects the same general concepts.

Refocusing back on AGI, suppose we invent superhuman AGI for $1T in 2100 and we ask it for advice to optimize humanity's paperclip production. Whatever advice it comes up with, I don't think it will be unrecognizable magic. Paperclip production is a fairly well understood problem: you bring in some metal ore, you rearrange its atoms into paperclips, and you send them out. A superintelligent AGI asked to optimize paperclip production will probably reinvent much of the same advice we've already derived: invest in good equipment, take advantage of economies of scale, select cheap yet effective metals, put factories near sources of energy and materials, ship out paperclips in a hierarchical network with caches along the way to buffer unanticipated demand, etc. If it gives better advice than a management consultant, we expect it to do so not by incomprehensible omnipotent magic that inverts the laws of physics, but by doing what we're doing already—just smarter, faster, and more efficient.

More practically, if we invent AGI and ask it to invent fusion power plants, here's how we think it will go: we don't expect it to think for a moment and then produce a perfect design, as if by magic. Rather, we expect it to focus on known plasma confinement approaches, run some imperfect simulations, and then strategize over what experiments will allow it to acquire the knowledge needed to iteratively improve its designs. Relative to humans, its simulations may be better and its experimentation may be more efficient, but we expect its intelligence will operate within the same constraints that we do.

Intelligence is not omnipotence.

# Appendix: What about progress in GPT?

To what extent should the tremendous success of the GPT paradigm, recently culminating in GPT-4, be evidence for AGI by 2043?

GPT-4 is *already* superhuman in some aspects:

- It outperforms most humans on most standardized tests
- It reads and writes faster and at lower cost than any human
- It speaks dozens if not hundreds of languages

And GPT-4's intelligence is *already* general in some aspects:

- It wasn't designed to solve a particular narrow problem
- It can learn, not only via training, but also via in-context learning, allowing it to solve problems never before seen in its training data

Given the wild pace of progress since GPT-1, and the existence of a system that feels to some to already be on the verge of AGI,[195] why are we so confident that 20 more years of progress and optimization won't result in something like transformative AGI?

To us, GPT-4 is far from transformative AGI for a variety of reasons.

The GPT paradigm, despite its many successes, still cannot do many things:

- Cannot learn continuously over time
- Has little self-knowledge
- Has little ability to plan and reflect and search
- Alongside its superhuman strengths, it has many embarrassingly subhuman weaknesses[196]
    - (these weaknesses will *of course* improve, but it will take time and possibly new breakthroughs, as current methods have clearly been unable to solve them)

Now, none of these obstacles are impossible, but solving them with more data and more compute will be expensive, both in terms of time and money. Already, a single call to GPT-4 can cost up to $2. An autoGPT-like system that consists of, say, a thousand calls per second to reflect on memories and beam search over parallel thoughts and synthesize conclusions could easily end up costing in excess of $100,000/hr to run, at today's costs. Future systems will no doubt improve; but the key questions are by how much and how quickly.

One issue is that the gains from scaling up GPT to GPT-2 to GPT-3 to GPT-4 are potentially diminishing. Wired reports:

> *"I think we're at the end of the era where it's going to be these, like, giant, giant models," [OpenAI CEO Sam Altman] told an audience at an event held at MIT late last week. "We'll make them better in other ways." [...] Altman said there are also physical limits to how many data*

[195] [2303.12712] Sparks of Artificial General Intelligence: Early experiments with GPT-4

[196] https://github.com/openai/evals is a fun open-source repo by OpenAI that shows all sorts of tasks that GPT-3.5 and GPT-4 regularly fail at.

> *centers the company can build and how quickly it can build them. [...] At MIT last week, Altman confirmed that his company is not currently developing GPT-5. "An earlier version of the letter claimed OpenAI is training GPT-5 right now," he said. "We are not, and won't for some time."*

## Exponential progress in AI is likely to slow

Progress in AI can be decomposed into three factors:

AI performance = (AI performance / operation) * (operations / $) * ($)

Performance per operation is **software efficiency**, and can be further divided into:
- algorithms (e.g., deep neural networks, transformers)
- data (e.g., giant corpuses of text, images, chemical structures)

Operations per dollar is **hardware efficiency**, and can be further divided into:
- chip design (operations per transistor)
- chip manufacturing (transistors per dollar)

The final factor, **willingness to spend**, has arguably been as important to AI progress as the first two.

In fact, **much of the progress over the past decade has been due to willingness to spend**, just as much as software and hardware efficiencies. A computer became the world's best Go player not only because DeepMind invented a new algorithm, but also because they convinced a Google executive that winning a Go match was worth many millions of dollars.

The number of floating-point operations used in large AI models has grown *much* faster than the cost per floating-point operation has dropped. In the 10-year period from AlexNet to PaLM, compute spent on the largest models rose by a factor of 5 million!

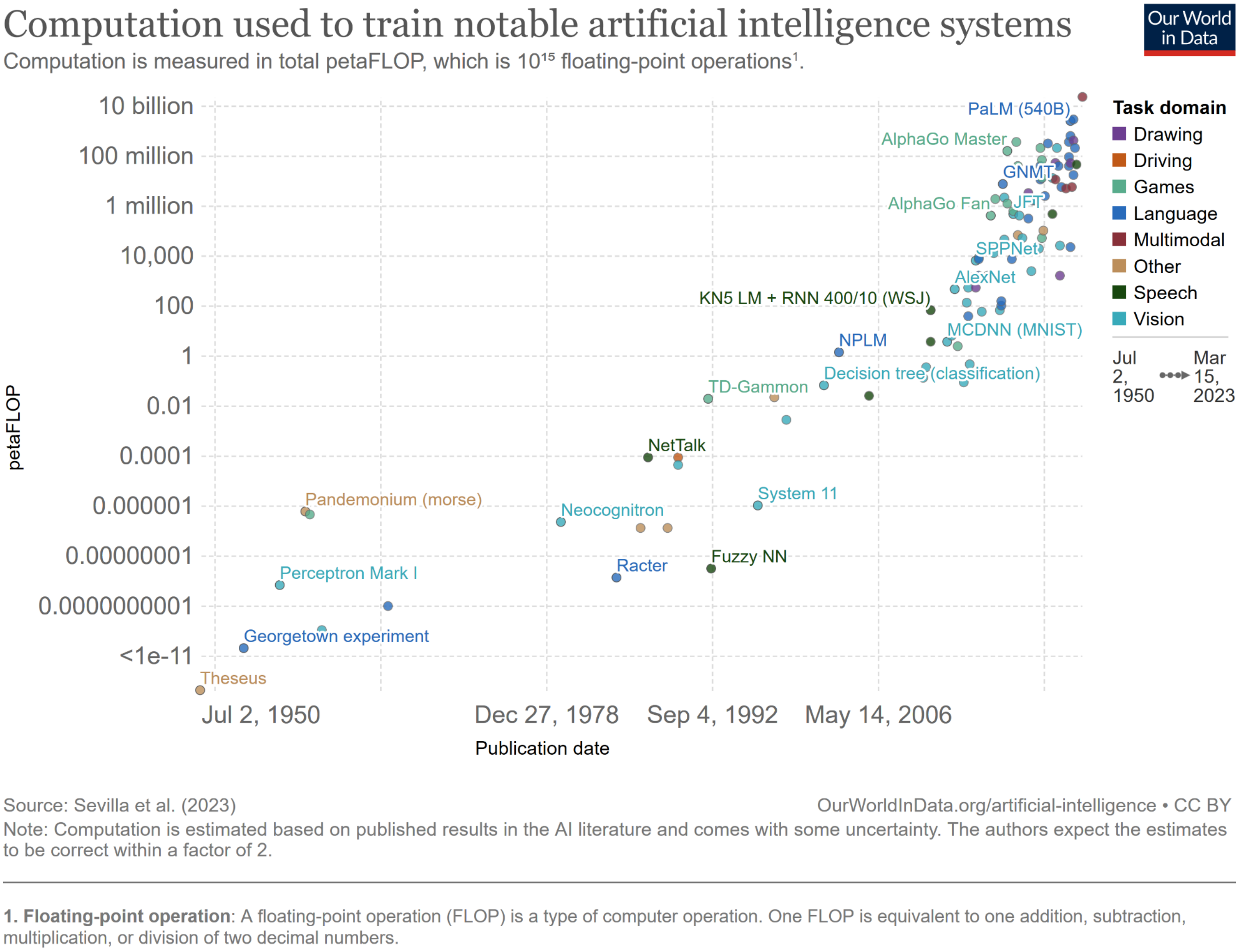

*Source: [Our World in Data]*

Although compute has been getting exponentially cheaper (~40% per year from 2016 to 2023, as benchmarked by the DGX-1 and DGX H100), it's not even close to the pace of growing spend (~370% per year from 2012 to 2022, as benchmarked by AlexNet and PaLM). AlexNet was trained by a few academic researchers using two GPUs for a few days, at a ballpark cost $3–10.[197] PaLM was trained by an industrial lab using 6,144 chips for more than a thousand hours, at an estimated cost of $9M–$23M.[198]

Now of course it shouldn't be a surprise that society spends more on a new technology as it gets better. A path toward transformative AGI will no doubt require unprecedented investment, and rising investment costs are as much a sign of progress as a sign of impediment.

---

[197] $3–4 is my estimate from two 580 3 GB GPUs costing $500 each, attributed to 5 days of cost from a 3-year amortization period. If they ran at 150 W each, and electricity was $0.10/kWh, that's another $3.60. So maybe $3–$10 total, ish.

[198] Estimating 🌴PaLM's training cost

Ross Gruetzemacher has humorously pointed out that DeepMind's operating costs, extrapolated beyond 2011–2018 suggests that it will soon exceed the costs of the Manhattan Project, Apollo Program, and Google's income, combined.

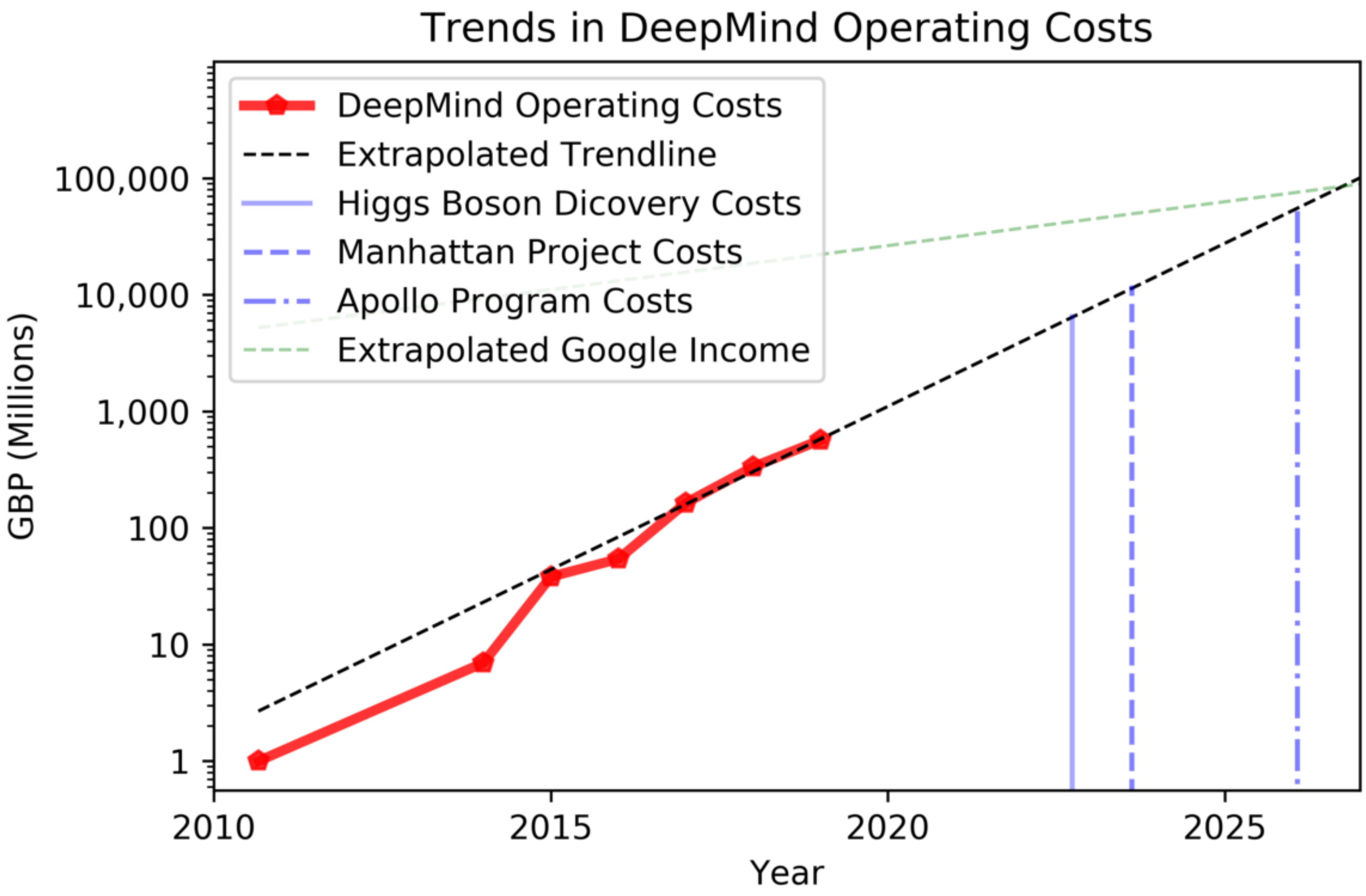

*Source: http://www.rossgritz.com/uncategorized/updated-deepmind-operating-costs/*

Now, of course, DeepMind's costs did not follow this extrapolation.

The reason why is instructive.

When models are small, scaling them is really easy. Scaling a $1 model by 1,000x only costs $1,000. Just go to Amazon, buy or rent a few GPUs, and you're good to go. In general, when costs are below $1M, or below $1B, it's "easy" to scale by writing bigger checks. This is partly why compute was able to rise by a factor of ~5 million over a decade.

But when models become large, scaling goes from easy to hard. Every factor of 10 is 10x more expensive than the one prior. Going from spending $1M to $1B to $10B to $100B is not something that can be done at retail prices with retail inventories at retail timescales. It will require new data centers and new power plants and new fabs. It will take *years* of planning and coordination. It will take stomach-cramping contracts to be signed by multiple parties, none of whom have any guarantee the investment will succeed.

Now, society makes large, complex investments all the time. And in a world with unprecedented benefit from AI systems, we will no doubt see unprecedented investment in the AI hardware supply chain.

But we really want to emphasize that this is going to be qualitatively unlike the scaling that took place from 2010–2020. It's going to be much harder and *much slower* than the dynamics behind AlexNet to PaLM. And it's only going to be affordable for a few industrial labs backed by deep-pocketed investors with billions to invest (e.g., Google DeepMind, OpenAI, Anthropic, etc.).

PaLM cost perhaps ~1 million times as much as AlexNet. Something that costs ~1 million times as much as PaLM would be ~$1T, larger than the entire semiconductor market. No one is going to spend $1T on a model by 2030. Spending growth *will* slow.

Ultimately we do not expect the GPT paradigm to scale in the future as it has in the past, and we do not expect the GPT paradigm to take us to AGI without new paradigmatic breakthroughs. For this reason, we do not see GPT-4's success, impressive as it is, as strong evidence for transformative AGI by 2043.

# Appendix: Investors forecast a low probability of transformative AGI by 2043

AGI is a technology that invents more technologies.

Transformative AGI, if invented, will be the most valuable technology of all time.

Today, world gross product is something like ~$100T/yr. In a world where transformative AGI reduces the scarcity of brains and bodies, production could easily skyrocket by 10x or more.

If such a world were on the horizon, we would expect to see:

- High interest rates
  - Demand for money might grow as the world invests in production of AGI inputs (e.g., semiconductors, electricity) as well as AGI outputs (everything)
  - Put another way, deferring consumption in lieu of high-ROI investment allows for greater consumption later
- Lowish asset prices
  - If transformative AGI can replace any company, those companies have less value (particularly if their value comes from things quickly obsoleted, like a well-trained workforce)
  - Companies highly substitutable by AGI should have lower asset prices (e.g., those without regulatory protections)
  - Companies with longer payoff horizons (e.g., growth tech companies) should have lower asset prices
- High asset prices for AGI owners or complements
  - Firms that (a) sell AGI inputs and (b) have market power may reap huge windfalls from the rise of AGI (at least until AGI-powered competitors compete them away)
  - Natural resources whose supply is not expanded by AGI may see a rise in value, as complementary processing steps become less expensive and more productive

Similarly, if negatively transformative AGI were on the horizon, we would *also* expect to see:

- High interest rates
  - Supply of savings would fall as there's nothing to save for if we all soon die or become disempowered
  - Demand for money would grow as people take loans they expect to never pay back
- Low asset prices
  - Productive assets would fall in value, as their future production would be difficult to reap
- High asset prices for consumption goods
  - Enjoyable consumption goods might see a spike in value as people vie to spend their wealth before they all die or become disempowered by negatively transformative AGI

Summarizing these three scenarios:

| Scenario | Interest rates | Asset prices |
|---|---|---|
| No transformative AGI | low | normal |
| Transformative AGI | high | Substitutes of AGI: low<br>AGI assets: high<br>Complements to AGI: high |
| Deadly AGI | high | Productive assets: low<br>Consumption assets: high |

Therefore, to see whether participants in financial markets believe AGI is coming, we can look at interest rates and asset prices.

So what do we see?

## Interest rates are low

Not only are interest rates low, interest rates are ***historically low***.

1-year real interest rates have been mostly negative the past decade, and as of May 10, 2023 sit at 1.2%, a value utterly at odds with AGI being developed in the next 20 years.

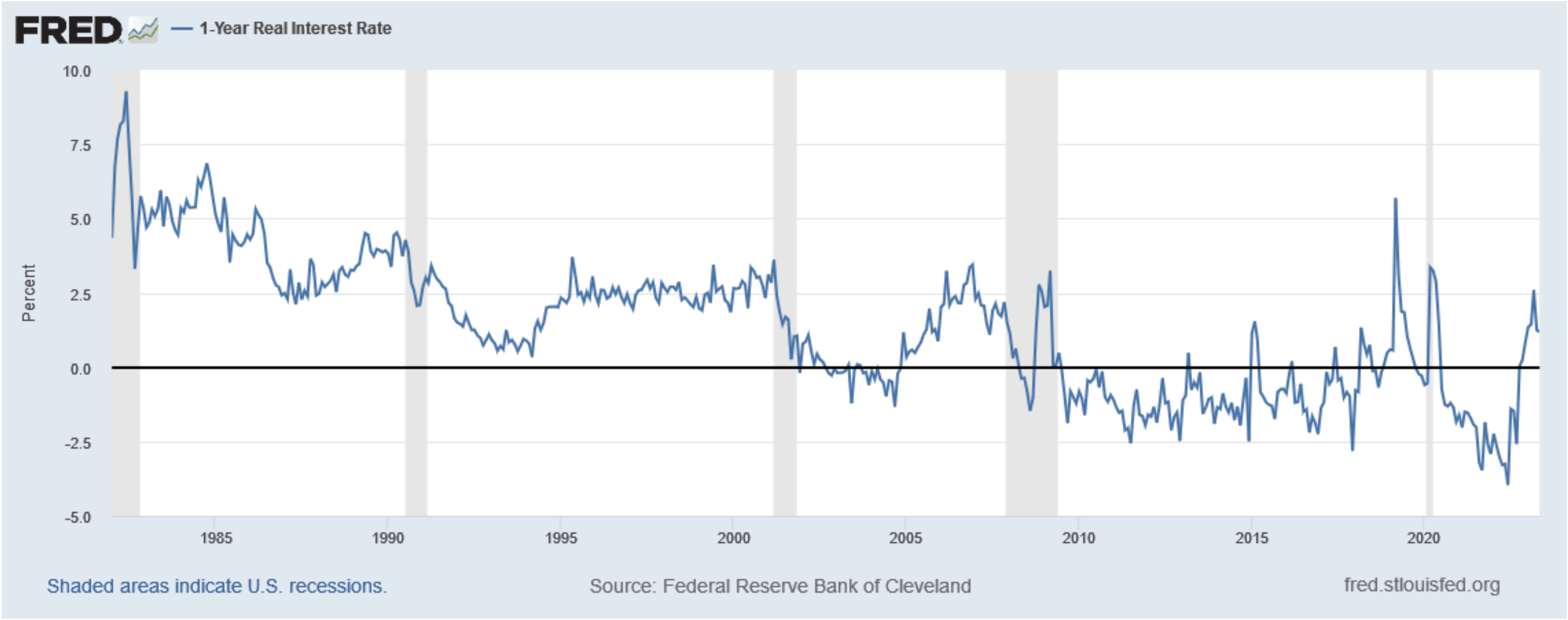

*Source: FRED*[199]

[199] 1-Year Real Interest Rate (REAINTRATREARAT1YE) | FRED | St. Louis Fed

The picture doesn't change much on longer time scales. The 30-year real interest rate sits at a slightly less measly 1.6%, as of May 10, 2023.

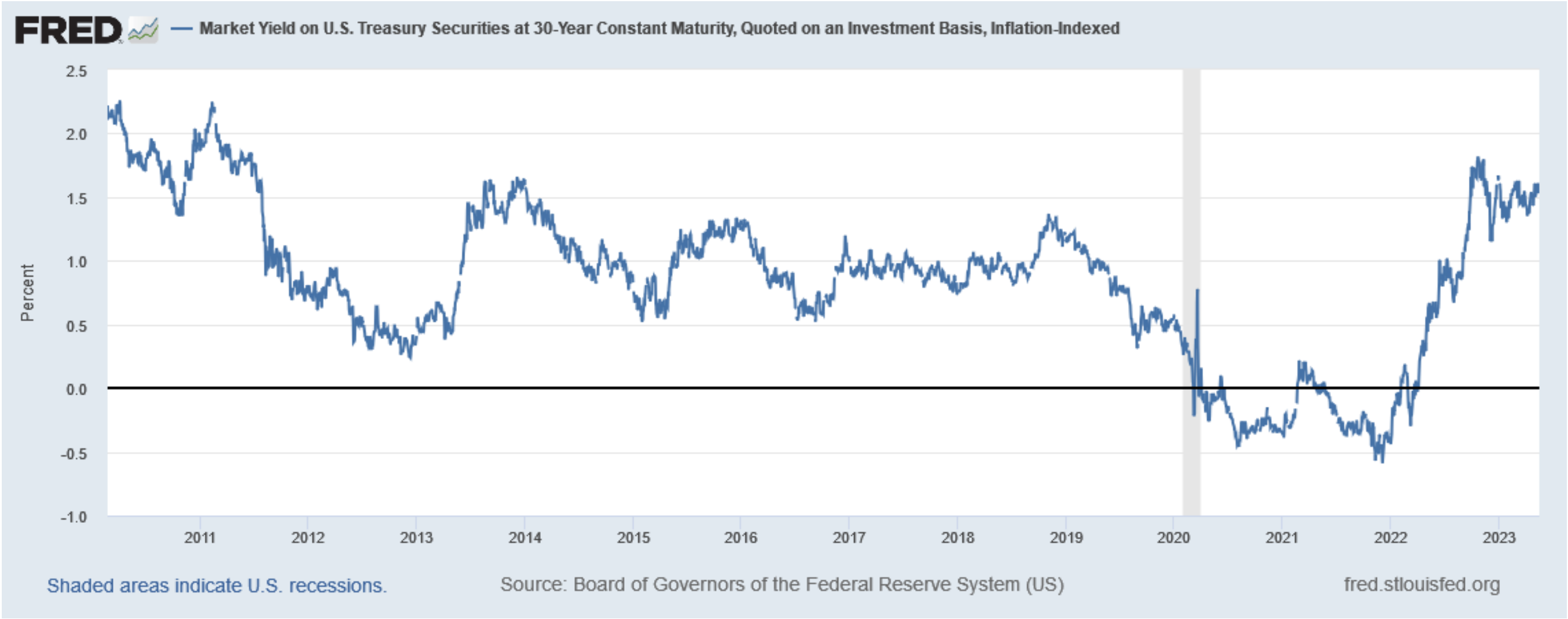

*Source: FRED*[200]

## Asset prices are normal

Asset prices are trickier to interpret, because some factors push them up (e.g., is it a complement to AGI? is it likely to win and hold market power?) whereas others push them down (e.g., is it a substitute to AGI? Will AGI destroy value? Will AGI outcompete the company quickly? Will windfall profits be nationalized? Will the specific company be a loser?).

Still, in a world with transformative AGI on the horizon, and all the above forces in play, we'd expect to see:

- Higher variation in asset prices vs historical patterns
- In the event of positive AGI: Higher spreads between AGI complements and substitutes
- In the event of negative AGI: Higher spreads between productive assets and consumption assets

But as far as we can tell, asset prices look pretty normal.

**Potential complements to AGI**

Computer hardware and software companies have valuations and P/E ratios that feel normal by historical standards. The high P/E ratios of NVIDIA and AMD could be interpreted as weak evidence for AGI. In contrast, the low valuations of Alphabet and Anthropic could be interpreted as weak evidence against AGI.

---

[200] Market Yield on U.S. Treasury Securities at 30-Year Constant Maturity, Quoted on an Investment Basis, Inflation-Indexed (DFII30) | FRED

| Asset | Market cap | P/E |
|---|---|---|
| TSMC (NYSE: TSM) | $442B | 14 |
| Intel (NASDAQ: INTC) | $120B | 38 |
| Samsung (KRX: 005930) | $330B | 10 |
| ASML (NASDAQ: ASML) | $266B | 35 |
| NVIDIA (NASDAQ: NVDA) | $746B | 173 |
| AMD (NASDAQ: AMD) | $167B | 430 |
| Alphabet (NASDAQ: GOOGL) | $1,540B | 27 |
| Microsoft (NASDAQ: MSFT) | $2,330B | 34 |
| Anthropic (private) | $4.1B | - |

*Prices as of May 17, 2023*

In addition, natural resources haven't skyrocketed or crashed in value, in anticipation of AGI:

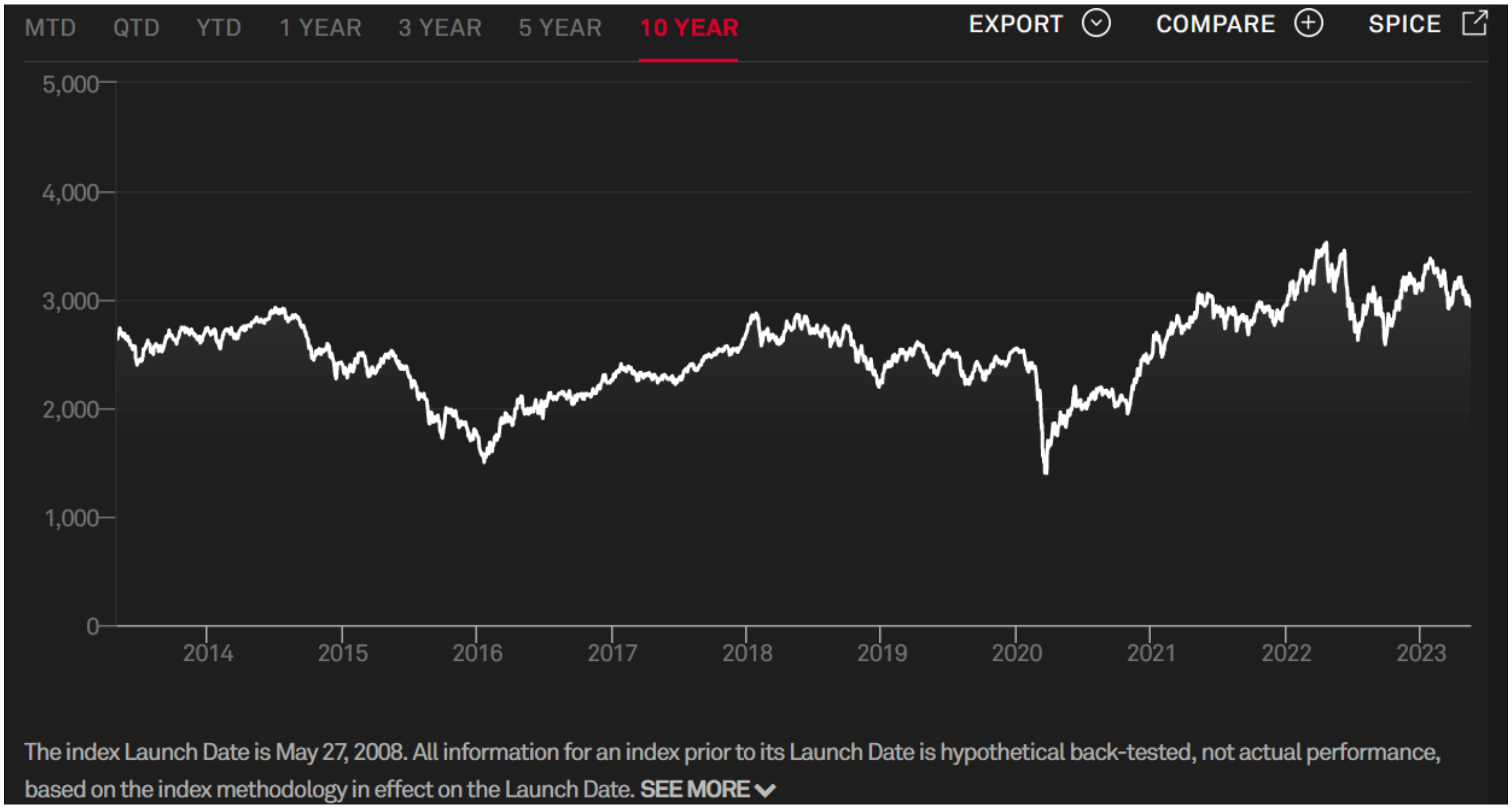

*S&P Global Natural Resources Index*[201]

[201] S&P Global Natural Resources Index | S&P Dow Jones Indices

## Public market summary

Obviously, interest rates and asset prices won't tell us whether transformative AGI will occur soon. All they tell us is that market participants, in aggregate, do not believe transformative AGI will occur soon. Given the trillions of dollars worth of potential value that could be earned by correct predictions of AGI, we take this as *weak* evidence that transformative AGI is unlikely to occur by 2043. We do not take it as *strong* evidence, given how poorly financial markets predicted the impacts of computers, internet, cell phones, etc.

## Informed private market investors *also* forecast low likelihoods of transformative AGI by 2043

In the prior section, we argued that public market investors, who are perhaps poorly informed, forecast low odds of AGI by 2043.

In this section, we argue that *informed* intracompany investors *also* appear to forecast low odds of AGI by 2043.

Consider Alphabet, the parent company of Google DeepMind. Alphabet is *not* constrained by financial capital: as of March 2023, Alphabet had $115B of cash on hand.[202] Alphabet is also *not* constrained by human capital: as of March 2023, Alphabet had 191,000 employees.[203] If Alphabet wanted to, it could *easily* hire more AI researchers and allocate more software engineers to AI research infrastructure. But it chooses not to. Why? This deliberate choice presumably implies that Alphabet management doesn't believe there is enough ROI from hiring more smart people to accelerate progress on AGI research. To us, this suggests that Alphabet management is bearish on the likelihood of transformative AGI by 2043. We also see this as a stronger signal than public markets, as Alphabet management is likely *much* more informed about AGI progress than the average stock trader.

Another implicit signal of low likelihood of transformative AGI is the comparatively low valuation of AGI companies. Consider Anthropic, which reportedly raised money in early 2023 at a valuation of ~$4B. $4B is ~0.004% of world product each year. If transformative AGI is soon able to do most human work, we expect it to be worth multiples of today's world product. Supposing transformative AGI is worth $100T/yr, and Anthropic has a 10% chance of capturing 10% of the value for an expected duration of 10 years, that would imply a valuation of $10T (before time discounting). $4B is less than 1/1,000th of $10T, which seems to imply that Anthropic and its investors do not collectively believe that Anthropic has a chance of inventing transformative AGI by 2043 and capturing its value.

We take these signals from Alphabet's management, and Anthropic and its investors, to be decent evidence that many informed experts do not expect transformative AGI by 2043, which is additional albeit imperfect evidence that transformative AGI will not occur by 2043.

---

[202] Alphabet (Google) (GOOG) - Cash on Hand
[203] Alphabet Number of Employees 2001-2023 - Stock Analysis

## Acknowledgements

We thank the many readers who gave us thoughtful feedback, including: Alexey Guzey, Andrew Kondrich, Boaz Allyn-Feuer, Hillary Sanders, Jacob Hilton, John Croxton, Jon Stokes, Kjirste Morrell, Linchuan Zhang, Patrick Waters, and Raimondas Lencevicius.